\documentclass{llncs}
\usepackage{amsmath}
\usepackage{amssymb}
\usepackage{bm} 
\usepackage{amsfonts}
\usepackage{booktabs}
\usepackage{tabularx}
\usepackage[compatibility=false]{caption}
\usepackage{subcaption}
\usepackage{graphicx}
\usepackage{pgfplots}
\usepackage[all]{nowidow}
\usepackage[utf8]{inputenc}
\usepackage{tikz}
\usetikzlibrary{er,positioning,bayesnet}
\usepackage{multicol}
\usepackage{algpseudocode,algorithm,algorithmicx}
\usepackage[inline]{enumitem}
\usepackage{csquotes}
\definecolor{blue}{HTML}{1F77B4}
\definecolor{orange}{HTML}{FF7F0E}
\definecolor{green}{HTML}{2CA02C}

\pgfplotsset{compat=1.14}

\begin{document}

\title{Deep Manifold Part 2: Neural Network Mathematics}

\author{Max Y. Ma \thanks{corresponding author: paper@deepmanifold.ai}  and Gen-Hua Shi}

\institute{deepManifold}

\maketitle

\begin{abstract}
This work develops the global equations of neural networks through stacked
piecewise manifolds, fixed--point theory, and boundary--conditioned iteration.
Once fixed coordinates and operators are removed, a neural network appears as a learnable numerical computation shaped by manifold complexity, high--order
nonlinearity, and boundary conditions. Real--world data impose strong data complexity, near-infinite scope, scale,
and minibatch fragmentation, while training dynamics
produce learning complexity through shifting node covers, curvature
accumulation, and the rise and decay of plasticity. These forces constrain
learnability and explain why capability emerges only when fixed--point regions
stabilize. Neural networks do not begin with fixed points; they construct them through residual--driven iteration. This perspective clarifies the limits of monolithic models under geometric and data--induced plasticity and motivates architectures and federated systems that distribute manifold complexity across many elastic models, forming a coherent world--modeling framework grounded in geometry, algebra, fixed points, and real--data complexity.

\end{abstract}

\section{Introduction}

Building on \emph{Deep Manifold Part 1: Anatomy of the Neural Network Manifold}  \cite{deepmanifoldpart1}, this second part continues the journey with a wider lens. In Part 1 our focus rested on the nonlinearity within neural networks themselves.  Here, we turn to data high-order nonlinearity, data scope, Neural Network mathematics, and learnable numerical computation.

\begin{enumerate}
    \item \emph{Neural Network Geometry}:
    Connected and stacked piecewise-smooth manifolds collectively form the geometric structure of the representation space. Node covers act as the local units of these piecewise-smooth manifolds, and their orientations change at every iteration. piecewise-smooth manifolds are differentiable and integrable.

    \item \emph{Neural Network Algebra}:
    The coordinate system changes continuously with iteration. Counting is the most primitive algebraic unit, manifested through the iterated-integral structure of forward propagation. Meanwhile, the activation values themselves do not carry no inherent attributes: propertyless.

    \item \emph{Neural Network Equation}:
    The fixed-point residual is the most primitive equation of a neural network. The residual itself serves as a boundary condition, and its Lagrangian form provides the iterative method toward the moving fixed point, thereby granting neural networks dynamic iterative computability.

    \item \emph{Neural Network Stochastic}:
    Stochasticity is expressed through inequalities and sum-based group statistics. This statistical structure allows neural networks to learn a world whose nature is fundamentally random and to form stochastic fixed points naturally.

    \item \emph{Neural Network Fixed Point}:
    Iterations traverse billions of piecewise-smooth manifolds, producing countless fixed points and convergence paths in foundation models. Because training data itself are highly nonlinear, curvature (second-order derivatives) together with moderate perturbations is required to discern the correct convergence direction of fixed points.

    \item \emph{Neural Network Boundary Condition}:
    Boundary conditions are the sole source of iterative direction and determine the convergence path during training. When no static fixed points exist in a foundation model, symmetric, weak, and discrete boundary conditions become essential to guide convergence in highly nonlinear systems.
\end{enumerate}

What emerges is that the mathematics of neural networks is primitive yet
structurally rigorous. Once the classical assumptions of fixed coordinates, fixed
manifolds, and fixed operators are relaxed, the network reveals itself as a
numerical computation shaped by \emph{complexity}: residual complexity,
boundary condition complexity, and geometric complexity of piecewise stacked
 manifolds. A neural network is evaluated through iterated integrals on
evolving submanifolds, with dynamic coordinates and propertyless activations
providing the algebraic substrate. In this sense, a neural network is a
\emph{learnable numerical computation framework} operating under structural
complexity.

This viewpoint clarifies the distinction between \emph{trainability} and
\emph{learnability}: trainability reflects the unbounded degrees of freedom
generated by moving coordinates; learnability is constrained by \emph{manifold
complexity}—curvature, discontinuity, and boundary–induced deformation.
High–order nonlinearity produces neural plasticity through interconnected
toroidal structures whose accumulated curvature defines effective capacity. During training, these structures remain elastic; later, the geometric complexity increases and resistive behavior appears. Capability arises not from emergent intelligence, but from the stabilization of fixed–point regions allowed by this geometry.

Real–world data impose an additional governing force: \emph{data complexity}
arising from high–order nonlinearity, near–infinite scope, stochastic variability,
and minibatch fragmentation. Each iteration reveals only a small local slice of
a vast manifold, causing the curvature to accumulate unevenly. As training progresses,
the node covers shift, and the representational geometry stiffens; plasticity initially rises
and then decays. These interacting forces define the \emph{learning complexity}
of neural networks.

Despite these constraints, neural networks offer a structural advantage: forward
and inverse problems unify under the same boundary–conditioned iteration.
Additional observations enter directly as boundary conditions, allowing the
network to approximate mappings without closed–form formulations. The model
functions as a residual–driven solver on a learned manifold, with training
serving as numerical correction under evolving complexity.

Because large models accumulate geometric complexity and \emph{data–induced
plasticity} and lose elasticity, architectures with higher geometric flexibility and
distributed systems become natural extensions. Many small elastic models,
each learning a local region of the global manifold, maintain numerical
stability where monolithic models do not. Their alignment forms a coherent
manifold system without centralized aggregation.

Manifold federation therefore provides the mathematical bridge between the
real world and an AI world model—built from piecewise geometry, fixed–point
structure, and \emph{complexity–conditioned} iteration. Neural networks do not
begin with fixed points; they construct them. Their mathematics is primitive,
yet it extends numerical computation into a highly nonlinear
high-dimensional domain governed by stacking complexity.

\section{Neural Network Mathematics}
\begin{quote}
    \textit{Mathematics is very bureaucratic; without an overall perspective, one could turn forever in the same unyielding room}
\end{quote}
\subsection{Neural Network Geometry}
\begin{quote}
    \textit{Geometry bestows the eye that beholds all things from above,
a very ladder to the freedom}
\end{quote}
\subsubsection{Stretched, Stacked Piecewise Manifolds}
\label{sec:stretched_stacked_piecewise_manifolds}

The geometry of neural networks provides deep insight into their behavior, training dynamics, and generalization capabilities, which are essential for designing better models, improving optimization, and explaining why neural networks work so effectively despite their complexity.
Our findings reveal that neural networks exhibit a geometry of stretched, stacked piecewise manifolds, as illustrated below.

\begin{figure}[H]
    \centering
    \includegraphics[width=0.75\linewidth]{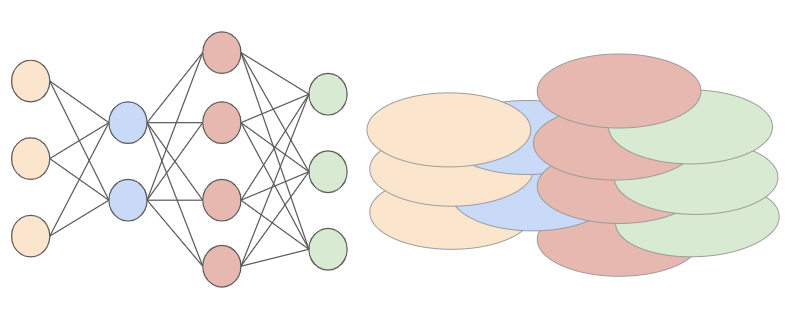}
    \caption{Neural Network Piecewise Manifold}
    \label{fig:nn_piecewise_manifold}
\end{figure}

Such geometry mirrors that of the Numerical Manifold Method (NMM)  \cite{shi1991manifold}, as illustrated below. 
\vspace{1em}

\begin{figure}[H]
    \centering
    \includegraphics[width=0.75\linewidth]{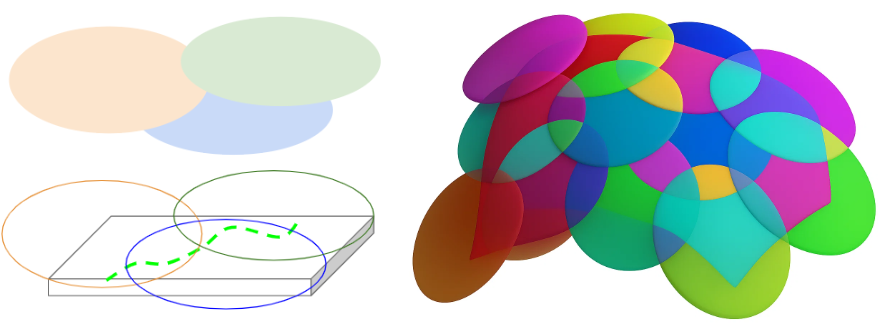}
    \caption{Numerical Manifold Method}
    \label{fig:numerical_manifold}
\end{figure}

The Numerical Manifold Method has been successfully applied to solid dynamics, fluid dynamics, aerodynamics, thermodynamics, electrodynamics, and acoustics, as well as complex coupling problems such as oil/gas–rock interactions. It has also been used in underground structural-stability analysis under nuclear or conventional blasts and in nuclear-waste repository design. 

One of the most relevant advances to \emph{Deep Manifold} is the Numerical Manifold Method with Independent Covers  \cite{independent_covers}, see Figure~{\ref{fig:numerical_manifold}. In particular, their work demonstrates accurate solutions for strongly convective one-dimensional Burgers systems and the two-dimensional incompressible Navier–Stokes equations, achieving high-precision shock and convection behavior while avoiding nonphysical oscillations. These results show that independent covers can stably capture high-order nonlinear flow structures on fixed meshes. In a way, neural networks are built on stacked, independent covers; that is, what we call \emph{stacked piecewise manifolds}.

\vspace{1em}
\begin{figure}[H]
    \centering
    \includegraphics[width=1\linewidth]{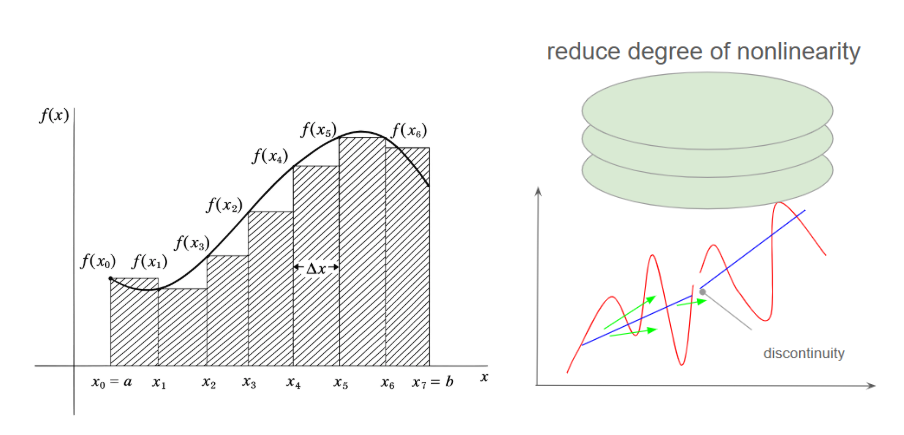}
    \caption{Calculus and Stacked Piecewise Manifold}
    \label{fig:calculus_stacked_piecewise_manifold}
\end{figure}

The Numerical Manifold Method is based on a simple idea: similar to integration. In addition to vertical partitioning, it also slices horizontally, breaking complex behavior into layers. This reduces the degree of nonlinearity, with each piecewise layer capturing a different aspect of the underlying behavior, even across discontinuities. 

\subsubsection{Piecewise Manifold Formalization}
\label{sec:piecewise_manifold_formalization}
Based on Section~\ref{sec:stretched_stacked_piecewise_manifolds} above and following Figure~\ref{fig:calculus_stacked_piecewise_manifold}, we formalize the representational geometry of each neural network layer as comprising a stacked collection of smooth piecewise manifolds. Specifically, for layer \(k = 0,1,\ldots,L\), we define
\begin{equation}
\mathcal{M}_k = \bigcup_{i=1}^{N_k} \mathcal{M}_{k,i}
\label{eq:piecewise-manifold}
\end{equation}
where each \(\mathcal{M}_{k,i}\) is a smooth piecewise manifold---a "pancake" region in the sense of the numerical manifold method, as shown in Figures~\ref{fig:nn_piecewise_manifold} and~\ref{fig:numerical_manifold}.

Given a model input \(x\), the initial embedding is
\begin{equation}
h_0 = E(x) \in \mathcal{M}_{0,i_0}
\label{eq:initial-embedding}
\end{equation}
with \(E\) a learned embedding operator that places the input in one of the sub-manifolds. Each neural network layer then operates in two synchronized dimensions:
\begin{enumerate}
    \item \emph{Representation update}:
    \begin{equation}
    h_{k+1} = f_k(h_k),
    \label{eq:representation-update}
    \end{equation}
    where \(f_k\) is the learned transformation of the activation.
    \item \emph{Manifold evolution}:
    \begin{equation}
    \mathcal{M}_{k+1,j} = \Phi_{k,i_k \to j}\big(\mathcal{M}_{k,i_k}\big)
    \label{eq:manifold-evolution}
    \end{equation}
    where \(\Phi_{k,i_k \to j}\) is a \emph{learned geometric operator} that transforms the manifold piece \(\mathcal{M}_{k,i_k}\) to a corresponding region in \(\mathcal{M}_{k+1}\).
\end{enumerate}

The operator \(\Phi_k\) is defined implicitly through the layer's learned architecture, such as multi-head attention, linear projections, nonlinear activations, residual pathways, and normalization in the Transformer. Although not explicitly parameterized as a closed-form function (e.g., \(y=ax+b\)), \(\Phi_k\) bends, stretches, and re-aligns local regions of the manifold dynamically during training. In this sense, \(\Phi_k\) plays an analogous role in coordinate transformations in differential geometry, although it is learned entirely from data rather than analytically derived.

Collectively, the neural network may be viewed as a stacked operator over activations and manifolds:
\begin{equation}
(M(L), h_L)
=
\big((f_{L-1}, \Phi_{L-1}) \circ \cdots \circ (f_0, \Phi_0) \big)
\big( M(0), h_0 \big)
\label{eq:stacked-manifold-operator}
\end{equation}
where \(M(0) = \{\mathcal{M}_{0,i}\}_{i=1}^{N_0}\) and \(M(L) = \{\mathcal{M}_{L,j}\}_{j=1}^{N_L}\). This formulation emphasizes that neural networks are not merely composed of successive functional layers but dynamically reshape the underlying representational geometry throughout depth, mirroring the \emph{Numerical Manifold Method's} decomposition both vertically and horizontally.

\subsubsection{Dynamic Node Cover}
\label{dynamic_node_cover}
Neural nodes act as evolving local covers of data transit manifolds, providing a geometric lens through which network learning can be understood. Each node defines a local approximation, piecewise in nature, that participates in covering the manifold of data transitions. As training progresses, these covers shift dynamically, adapting to better align with the underlying structure. This perspective reframes neural learning as the process of constructing and refining a piecewise manifold cover, where local patches collectively approximate high-order nonlinear data geometry.
\vspace{1em}
\begin{figure}[H]
    \centering
    \includegraphics[width=1\linewidth]{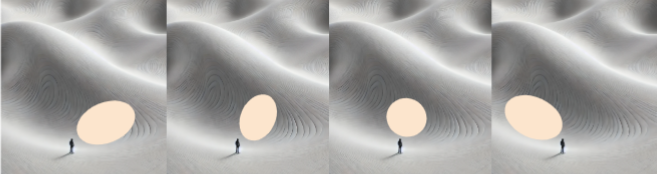}
    \caption{Node Cover}
    \label{fig:node_cover}
\end{figure}

Geometrically, neural networks build dynamic covers of data manifolds; algebraically, this manifests itself as a continual coordinate change induced by weight updates. Neural nodes act as evolving local covers of data transit manifolds. Their behavior can be summarized as follows.
\begin{enumerate}
    \item Each node defines a local approximation in a piecewise way.
    \item The node covers shift dynamically during training, adapting to the geometry of the manifold.
    \item Learning is the construction and refinement of a piecewise manifold cover, where many small patches jointly approximate a high-order nonlinear structure.
    \item Geometry and algebra coincide: the geometric cover movement corresponds to the algebraic coordinate change induced by weight updates.
\end{enumerate}

Dynamic node cover means the node’s coordinates change; see preceding Section~\ref{sec:node_coordinate_change} 

\subsubsection{Manifold World}
A manifold may take the form of a point, a line, a cycle, a triangle, a rectangle, or even an infinite-dimensional Banach or Hilbert manifold—these which we call classical manifolds. However, when confronted with high-order nonlinearity and discontinuity, such manifolds exhibit inherent structural limitations. 

This “connected–stacked–piecewise” manifold structure appears everywhere in daily life. An RGB image is the simplest example: the red, green, and blue channels are stacked together to form a representation of the real world—one that is fundamentally high-order nonlinear and filled with discontinuities. By stacking piecewise-smooth manifolds, we are in fact extending calculus itself, enabling it to operate on such highly nonlinear and discontinuous structures.

\begin{figure}[H]
    \centering
    \includegraphics[width=0.85\linewidth]{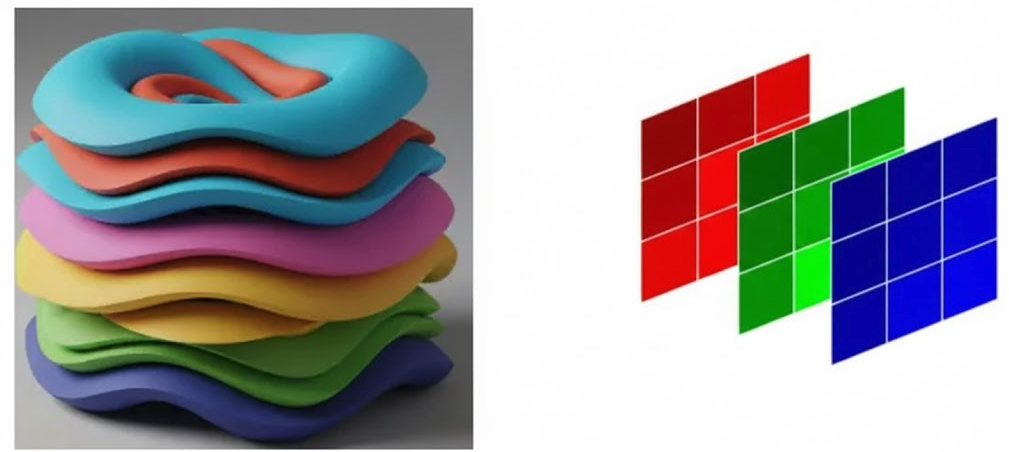}
    \caption{Pancake like, Manifold vs. RGB}
    \label{fig:pancake_manifold_RGB}
\end{figure}

One viable way to build an AI world model is to approximate it piecewise: stacked local manifolds, data-defined boundary conditions, and iterative fixed-point training that stays statistical rather than exact.

\subsection{Neural Network Algebra}
\label{sec:algebra}
\begin{quote}
    \textit{Algebra is the science of operations, the silent one behind all transformation.}
\end{quote}
\subsubsection{Data Efficiency via Coordinate Change}
\label{sec:node_coordinate_change}
In neural networks, weight updates do more than adjust numerical values, they continuously redefine the coordinate system through which data is represented. Each weight matrix functions as a set of basis vectors and each node activation
\begin{equation}
    a_{k,n}(h_k;\theta)
\end{equation}
acts as a coefficient determining how strongly that basis participates in approximating the input. Because the coordinate system itself changes with every weight update, neural networks are not restricted to fitting data in a fixed frame. Instead, they continuously reshape their representational space to align with the structure of the data, effectively bending the coordinates until patterns become linearly or locally separable. 

\begin{figure}[H]
    \centering
    \includegraphics[width=1.0\linewidth]{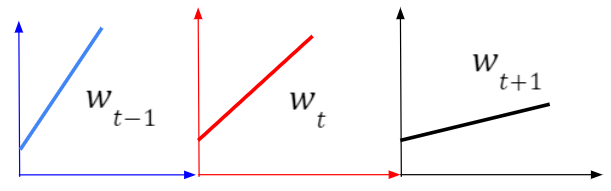}
    \caption{Changing Coordinate}
    \label{fig:coordinate}
\end{figure}

This dynamic reconfiguration grants neural networks an inherent advantage in learnability: the model is not only adjusting parameters but also adapting the algebraic scaffolding that interprets the data. In this sense, training is both curve-fitting and coordinate-crafting, which explains why neural networks can fit highly nonlinear and high-dimensional data where classical methods struggle.

This iteration-driven movement of node covers is the source of \emph{data efficiency}. Instead of learning in a static coordinate frame, the network adapts its coordinate system at each iteration, aligning its local representation with the portion of the data manifold exposed by the minibatch. This adaptive coordinate change greatly reduces the amount of data required for accurate fitting of high-order nonlinear structures.

\vspace{6pt}

Formally, begin with the layer-$k$ manifold (Equation~(\ref{eq:piecewise-manifold})):
\[
M_k = \bigcup_{i=1}^{N_k} M_{k,i}
\]

Each node $n$ in layer $k$ defines a local cover through its activation :
\begin{equation}
a_{k,n} : M_k \times \Theta \rightarrow \mathbb{R},\qquad 
a_{k,n}(h_k;\theta)
\end{equation}

In iteration $t$, the node’s cover region is :
\begin{equation}
U_{k,n}(t)
=
\{\, h_k \in M_k \mid a_{k,n}(h_k;\theta_t) > \tau \,\}, 
\qquad \tau > 0
\end{equation}

These regions form a time-dependent cover of the manifold:
\begin{equation}
M_k \approx \bigcup_{n=1}^{N_k} U_{k,n}(t)
\end{equation}
Each node functions as a local approximator on its patch:
\begin{equation}
f_k(h_k)
\approx
\sum_{n\in \mathcal{N}_k}
a_{k,n}(h_k;\theta_t)\,\psi_{k,n}(h_k)
\end{equation}

Parameters update per iteration as:
\begin{equation}
\theta_{t+1}
=
\theta_t - \eta_t\nabla_\theta L(\theta_t)
\end{equation}
which moves the node covers:
\begin{equation}
U_{k,n}(t+1)
=
\{\, h_k \mid a_{k,n}(h_k;\theta_{t+1}) > \tau \,\}
\end{equation}

Given an input $x$ with a layer-wise trajectory:
\begin{equation}
h_0(x), h_1(x), \dots, h_L(x)
\end{equation}
the data-transit manifold is:
\begin{equation}
\gamma(x) = \{\, h_k(x) \,\}_{k=0}^{L}
\end{equation}

The set of active covers at iteration $t$ is:
\begin{equation}
\mathcal{C}(x,t)
=
\{\, (k,n)\mid h_k(x)\in U_{k,n}(t)\,\}
\end{equation}

\vspace{6pt}

\noindent
\textbf{Data Efficiency}:
Because node covers $\{U_{k,n}(t)\}$ move at every iteration ($t$), the coordinate frame used to represent data also moves. This allows the network to:
\begin{enumerate}
    \item Fit high-order nonlinear regions with fewer samples, since the coordinate system adapts to the manifold fragment exposed by each minibatch.
    \item Sweep through the manifold over many iterations, reconstructing global geometry through accumulated local approximations.
    \item Expand effective representation capacity without increasing parameter count, because coordinate movement multiplies usable degrees of freedom.
\end{enumerate}

Thus, neural networks achieve data efficiency not merely through depth or scale, but through \emph{iteration-driven coordinate change}, a mechanism fundamentally distinct from classical numerical methods operating in fixed coordinate systems.

\subsubsection{Iterated Integral}
\label{sec:iterated_integral}
In classical analysis, iterated integrals are understood as compositions of linear operators. Because integration is linear, the order of composition does not matter; Fubini’s theorem guaranties equivalence. This embodies a foundational assumption in traditional mathematics: the iteration of well-behaved operations collapses into a commutative structure, one in which complexity is smoothed rather than generated.

\begin{figure}[H]
    \centering
    \includegraphics[width=0.75\linewidth]{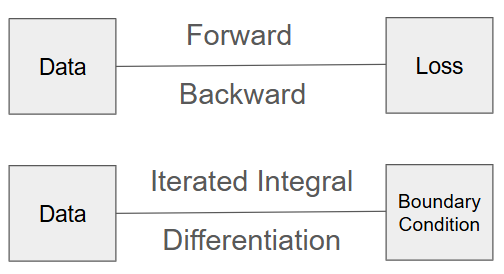}
    \caption{Iterated Integral}
    \label{fig:iterated_integral}
\end{figure}

By contrast, neural networks are explicitly built on the basis of function composition. Each layer combines a linear transformation with a nonlinear activation, producing operators whose compositions are inherently noncommutative and order-sensitive. Depth thereby enhances the capacity for hierarchical representation, where each transformation depends explicitly on the previous output. Iteration in this context no longer collapses structure, but instead accumulates and amplifies it across multiple sub-manifolds. We hypothesize that this mechanism constitutes the underlying source of in-context learning: whereas classical integration reduces compositions, neural networks exploit ordered, hierarchical iteration to transform the sequence.

\emph{Iterated Integral with Boundary Condition:} 
Let the prompt $p$ define the boundary condition $\partial\Omega(p)$ and the
initial embedding $h_0(p)=E(p)\in\mathbb{R}^m$.  
Inference proceeds as an iterated integral over prompt-bounded regions of the 
stacked, piecewise manifold introduced in Section~\ref{sec:stretched_stacked_piecewise_manifolds} 
At each depth $k$, the boundary $\partial\Omega_k(p)$ selects one slice of the 
stacked manifold as Equation (\ref{eq:piecewise-manifold})
\[
M_k \;=\; \bigcup_{i=1}^{N_k} M_{k,i},
\]
so that the integral in layer $k$ operates over the appropriate "pancake"
region determined jointly by the prompt and the previous activation.

The forward pass is described by Equation~(\ref{eq:forward_integral}):
\begin{equation}
\begin{aligned}
h_0(p) &= E(p) \in \mathbb{R}^m,\\[3pt]
h_k(p) &= 
\int_{\partial\Omega_k(p)} 
    f_k\!\big(h_{k-1}(p),x_k\big)\,
    \mathrm{d}\boldsymbol{\mu}_k(p)(x_k),
    \qquad k = 1,\dots,L,\\[3pt]
h_L(p) &\in \mathbb{R}^m \text{ is the final output.}
\end{aligned}
\label{eq:forward_integral}
\end{equation}

Although Equation~(\ref{eq:forward_integral}) does not explicitly display the piecewise manifolds, the 
selection of the integration boundary $\partial\Omega_k(p)$ and the 
prompt-conditioned measure $\mathrm{d}\mu_k(p)$ implicitly determine which 
sub-manifold $M_{k,i}$ is active at each layer.

The measure itself is defined by Equation~(\ref{eq:integral_boundary_condition})

\begin{equation}
\mathrm{d}\mu_k(p)(x_k)
  \;=\;
  W_k(p,x_k)\,\mathrm{d}x_k,
\label{eq:integral_boundary_condition}
\end{equation}

where $W_k(p,x_k)\in\mathbb{R}^{m\times m}$ is the learned weighting or gating 
matrix.  This weighting governs how the network integrates in the selected slice of the 
stacked manifold at depth $k$, adjusting the local geometric contribution below 
the boundary induced by the stimulus.

What makes the neural network extremely powerful is its \emph{primitive integral} nature.   In classical numerical computation, integration is often carried out piecewise or by near-piece approximations to ensure stability.  A neural network performs a similar process: each layer approximates a local integral of the data manifold, while the composition of layers forms a global, hierarchical integration with boundary condition, Section~\ref{sec:nn_boundary_condition}.  Unlike classical schemes, however, the network continuously \emph{self-corrects} these approximate integrals through backpropagation, the backward derivative, so that accumulated local errors are iteratively minimized.  This dynamic coupling of forward approximation and backward correction allows neural networks to act as adaptive numerical integrators over high-dimensional, nonlinear domains.

\begin{quote}
    \textit{The follow-up question is: where is the integral path? see Section~\ref{sec:intrinsic_pathway}}
\end{quote}

\subsubsection{The Most Basic Algebraic Operation: Counting}
\label{sec:counting}
In classical mechanics or fluid dynamics, we compute intrinsic physical quantities: stress, strain, pressure, velocity, each grounded in the internal state of the system and governed by conservation laws.

Neural networks, by contrast, do not compute intrinsic quantities in this sense. The loss function, often treated as a guiding objective, is not intrinsic to the network itself, but is an external signal introduced to steer convergence. Because neural networks lack a static convergence point, the loss cannot be interpreted as measuring any absolute or conserved property of the system. This raises a foundational question: What exactly are neural networks computing? In other words, what property of the data is encoded in their activations?
\vspace{1em}
\begin{figure}[H]
    \centering
    \includegraphics[width=0.75\linewidth]{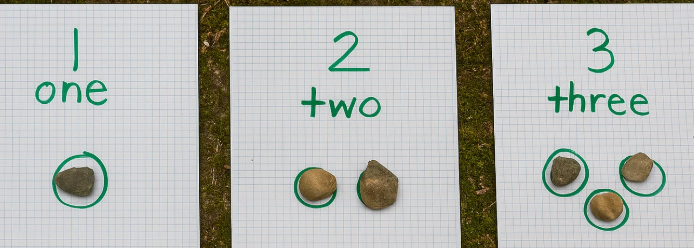}
    \caption{Counting}
    \label{fig:counting}
\end{figure}

If we examine the heads of Transformer models, their operation is fundamentally
classificatory. The vocabulary defines the discrete class space, and every activation
represents a local stage of counting: an accumulation of evidence for one token
over others. The final logits are therefore not intrinsic semantic quantities but
the result of a continuous counting process defined over the manifold established
by the forward pass.

In the Deep Manifold framework, inference is not performed over token
positions but as an iterated integral across a sequence of learned manifolds.
Each layer acts as an integral operator that aggregates and transports evidence:
\begin{equation}
h_{\ell}(\xi)
=
\int_{M^{(\ell-1)}}
K_{\ell}(\xi,\eta)\,
\sigma\!\big( W_{\ell} h_{\ell-1}(\eta) \big)\,
d\mu_{\ell-1}(\eta)
\label{eq:manifold_layer_integral}
\end{equation}

Stacking these operators yields a full $L$-layer manifold integration:
\begin{equation}
h_{L}(\xi_L)
=
\int_{M^{(0)}}\!\!\cdots\!\!\int_{M^{(L-1)}}
\left(
\prod_{\ell=1}^{L}
K_{\ell}(\xi_{\ell},\xi_{\ell-1})
\right)
\Phi(p)\,
d\mu_{L-1}(\xi_{L-1})\cdots d\mu_{0}(\xi_0)
\label{eq:iterated_manifold_integral}
\end{equation}
where $\Phi(p)$ is the embedding of the boundary in prompt condition and
$M^{(\ell)} = \bigcup_i M^{(\ell)}_i$ is the piecewise stacking manifold at depth~$\ell$.

The categorical decision for the next token is then obtained by integrating
the final manifold representation through a class-specific counting field:
\begin{equation}
z_c
=
b_c
+
\int_{M^{(L)}}
\theta_c(\xi)\, h_L(\xi)\, d\mu_L(\xi),
\qquad
p(c \mid p)
=
\frac{e^{z_c}}{\sum_{k\in V} e^{z_k}}
\label{eq:final_logit_counting}
\end{equation}

Thus, the familiar statement that \enquote{LLMs use logits for next-token classification}
becomes, in the Deep Manifold view, the final step of manifold counting:
a continuous accumulation of geometric evidence across  piecewise \& layer-wise manifolds,
normalized over the vocabulary.

\noindent

\vspace{1em}
Equations above describe inference in the \emph{Deep Manifold framework}:
each layer integrates local evidence over the manifold, base functions define the piecewise manifold's topology and curvature, and the output softmax performs categorical normalization: the continuous limit of discrete counting.

\subsubsection{Propertyless Property}
\label{sec:propertyless}
Unlike classical mathematics, where abstraction captures richer structures and conserved quantities, neural networks reduce abstraction downward to the simplest algebraic unit: counting. Counting itself is propertyless: each tally has no inherent meaning beyond its accumulation. Yet, it is precisely this propertyless property that grants neural networks remarkable flexibility. Because activations need not encode intrinsic semantics, they can serve as a universal medium of representation. We introduced the concept of “feature bits” in our \emph{Part 1} paper\cite{deepmanifoldpart1}.  The property of feature bits is propertyless. 

Text, images, audio, and other modalities can therefore be mapped to the same activation space, mixed, and transformed through the same count algebra. In this way, the very absence of intrinsic property becomes the foundation of multimodal unification.

\subsection{Neural Network Equation}
\begin{quote}
    \textit{An equation is the quiet connector that enables computation}
\end{quote}
\subsubsection{Fixed-Point Residual as the Primitive Equation}
A dog can be represented as text or as an image. Once embedded, both enter the
same neural operator:
\begin{equation}
x_{\text{text}} = E(\text{\enquote{dog}}), \qquad
x_{\text{image}} = E(\text{dog pixels})
\end{equation}
After passing through the network, both representations map close to themselves:
\begin{equation}
f(x) \approx x.
\end{equation}
This is the fixed-point equation in its most primitive form:

\begin{equation}
f(x) - x = e(x), \qquad \min_{\,\theta} e(x)
\label{eq:fixed_point_residual}
\end{equation}

The network learns by reducing this residual so that \enquote{dog},between modalities,
stabilizes as a fixed point of the learned manifold operator.

Let $f_{\theta}$ be the neural operator parameterized by $\theta$.  
For any embedded input~$x$, the fundamental equation of neural learning is Equation~(\ref{eq:fixed_point_residual}) above. 
Training minimizes this residual over the data:
\begin{equation}
\theta^{\ast}
= \arg\min_{\theta} \;
\mathbb{E}_{x}\,\bigl\| f_{\theta}(x) - x \bigr\|.
\end{equation}
A converged representation satisfies the fixed-point condition:
\begin{equation}
x^{\ast} = f_{\theta^{\ast}}(x^{\ast}).
\label{eq:fixed_point_dot}
\end{equation}
Every stable concept, feature, or manifold region is therefore a fixed point of the
operator.

The equations above express the fixed point condition in its most primitive
form: learning proceeds by minimizing the residual $f_\theta(x)-x$ so that
representations stabilize as fixed points of the operator. However, this
formulation does not yet account for the architectural or domain constraints
under which the network evolves. In practice, the weights $\theta$ do not move
freely: normalization, attention structure, parameterization, and data geometry
impose boundary conditions that restrict the admissible region of optimization.
These constraints reshape the pathway toward the fixed point and determine which
fixed points are reachable or stable. Introducing these constraints in variational
form naturally extends the primitive fixed-point equation into the Lagrangian
formulation of Equation~(\ref{eq:lagrangian_formulation}) below, where the residual becomes the energy term and the
constraints enter as boundary terms enforced by the dual variable $\lambda$.

\subsubsection{Lagrangian Formulation of Neural Fixed Points}
\label{sec:lagrangian_formulation}

The fixed point residual can also be written in variational form.  
Let the residual define the functional energy and let the architectural or data constraints be $g(\theta)=0$.  
The Lagrangian is
\begin{equation}
L(\theta,\lambda)
= \mathbb{E}_x\,\|f_\theta(x)-x\|^{2}
+ \lambda\, g(\theta)
\label{eq:lagrangian_formulation}
\end{equation}

\paragraph{Saddle-Point Stationarity.}
The variational equilibrium enforces stationarity simultaneously in the parameter and constraint directions:
\begin{equation}
\nabla_\theta L(\theta,\lambda)=0,
\qquad
\nabla_\lambda L(\theta,\lambda)=0
\end{equation}

\paragraph{Variational--Fixed-Point Identity.}
When the gradient with respect to $\theta$ vanishes, the forward operator ceases to deform $x$.  
This stationarity coincides exactly with the representational fixed point:

\begin{equation}
\nabla_\theta L(\theta,\lambda) = 0 
\;\;\Longrightarrow\;\;
\text{critical point of } \mathbb{E}_x \|f_\theta(x) - x\|^2
\label{eq:representational_fixed_point}
\end{equation}
and in the ideal limit $f_\theta(x) = x$
\paragraph{Role of $\lambda$ Relative to the Weights.}
The weights $W \subset \theta$ shape the geometry of the forward operator, while the dual variable $\lambda$ imposes the admissible region in which such deformations may occur:
\begin{enumerate}
\item $W$ determines how the manifold bends, stretches, and transforms input.
\item $\lambda$ enforces the constraint $g(\theta)=0$ and regulates the allowable movement of the weights.
\end{enumerate}
Thus,
\[
\lambda\ \text{constrains the motion of}\ W;\qquad
W\ \text{realizes the constraint imposed by}\ \lambda.
\]

\begin{table}[H]
\caption{Lagrangian Formulation of Neural Fixed Point}
\centering
\begin{tabular}{p{5cm} p{7cm}}
\textbf{Item} & \textbf{Description} \\
\hline
Traditional name & Lagrange multiplier method \\
Variational Fixed-Point name & Enforcing constraints via dual variables \\
What’s happening mathematically 
& Saddle-point stationarity: $\nabla_\theta L = 0$, $\nabla_\lambda L = 0$.  
The Lagrange equilibrium corresponds exactly to the representational fixed point $f_\theta(x)=x$. \\
\end{tabular}
\end{table}

\paragraph{Deep Manifold Interpretation}:~The variational formulation Equations~(\ref{eq:lagrangian_formulation}) -- (\ref{eq:representational_fixed_point}) operate not on a \emph{single smooth
manifold}, but on a \emph{stacked piecewise manifold} as defined in
Section~\ref{sec:piecewise_manifold_formalization}. Each layer contains multiple smooth \enquote{pancake} regions, and the
forward operator moves activations across these regions while the parameter
update moves the manifold pieces themselves. This structure alters the
classical interpretation of Lagrange saddle-point equilibrium.

Classical Lagrange theory assumes optimization over a single differentiable
manifold with a well-defined surface on which the saddle point is taken. Neural
networks do not satisfy this assumption. The equilibrium is taken over many
local surfaces, each corresponding to a piecewise region of the stacked manifold.

\begin{enumerate}
\item \emph{Local curvature.}  
The residual $\|f_\theta(x)-x\|$ defines the curvature \emph{within a single piece}
$M_{k,i}$.

\item \emph{Piecewise boundary constraints.}  
The constraint $g(\theta)=0$ determines where each piece of the manifold can move,
consistent with the stacked geometry.

\item \emph{Layer-wise manifold evolution.}  
Training updates both Equations~(\ref{eq:representation-update}) and (\ref{eq:manifold-evolution})
\[
h_{k+1}=f_k(h_k),
\qquad
M_{k+1,j}=\Phi_{k,i\to j}(M_{k,i})
\]
So, the saddle-point condition is taken on a moving, piecewise cover, not a
static manifold.

\item \emph{Equilibrium between stacked pieces.}  
A fixed point exists when representation updates and manifold-piece motions
cease to deform,Equation~(\ref{eq:representational_fixed_point})
\[
\nabla_\theta L(\theta,\lambda)=0
\;\Longleftrightarrow\;
f_\theta(x)=x
\]
but this identity holds \emph{piecewise}, region by region, across the stacked
union, Equation~(\ref{eq:piecewise-manifold})
\[
\mathcal{M}_k = \bigcup_{i=1}^{N_k} \mathcal{M}_{k,i}
\]
\end{enumerate}

Thus, the neural network satisfies a dual, stacked identity: The fixed point in each representation piece \emph{is the Lagrange equilibrium} on its corresponding parameters.

Neural networks therefore operate simultaneously as residual fixed point solvers
on each local manifold piece and as variational Lagrangian systems whose
equilibrium must hold across all stacked piecewise manifold. 

\begin{quote}
    \textit{This explains why deep networks support multiple fixed points, multiple convergence pathways, and overlapping   attractor regions: a geometric consequence of stacked piecewise manifolds rather than a single smooth surface. Neural stability is therefore distributed: equilibrium is achieved across a hierarchy of interconnected manifold that evolve throughout depth.}
\end{quote}

\subsection{Neural Network Stochastic}
\label{sec:nn_stochastic}
\begin{quote}
    \textit{A stochastic world is an inequality world, but it is real.}
\end{quote}
\subsubsection{Inequality Boundaries, and Sum-Based Statistics}
The real world is stochastic by nature. It operates within the realm of probability and statistics, which we interpret as an inequality problem. Neural networks handle this naturally and effectively, often without explicit design. Stochasticity is embedded in their random initial weights as the initial distribution/condition, and because these weights are trainable, the network can absorb and represent the stochastic structure of the data itself. In this sense, neural networks offer one of the simplest and most elegant frameworks for modeling uncertainty and variability.

In the \textit{Deep Manifold} view, this stochasticity manifests most clearly through:

\begin{enumerate}
    \item Learnable stochastic representation of data: Training data contains measurement noise, sampling bias, and incomplete information. Neural networks internalize these probabilistic variations through their trainable stochastic weights.
    \item Boundary conditions as inequalities: Real-world constraints are often expressed as inequalities (e.g., $a \le x \le b$), as in \enquote{inequality theory} in Shi's latest work known as \enquote{contact theory}.  This frame captures feasible sets of outcomes rather than a single deterministic state.
\end{enumerate}

\begin{figure}[H]
    \centering
    \includegraphics[width=0.75\linewidth]{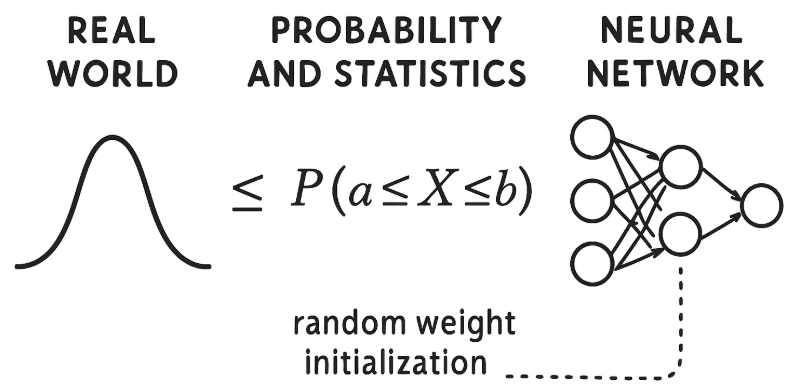}
    \caption{Neural Network Stochastic}
    \label{fig:nn_stochastic}
\end{figure}

A distinctive feature of stochasticity in neural networks is that each activation is not a single random variable, but a \emph{sum} over many stochastic contributions. For a node at depth $k$,
\begin{equation}
a_{k,n} = \sum_j W_{k,nj}\, h_{k-1,j}
\end{equation}

the activation reflects the aggregated influence of many random inputs filtered through many random weights on each individual piecewise manifold, Section~\ref{sec:piecewise_manifold_formalization}.  Its expectation and variance therefore arise from this superposition:
\begin{equation}
\mathbb{E}[a_{k,n}] = \sum_j \mathbb{E}[W_{k,nj} h_{k-1,j}]
\end{equation}
\begin{equation}
\mathrm{Var}(a_{k,n}) = \sum_j \mathrm{Var}(W_{k,nj} h_{k-1,j}) + \text{covariance terms}
\end{equation}
Each term in the summation corresponds to a local manifold piece $M_{k-1,i}$ within the stacked structure, Equation~(\ref{eq:piecewise-manifold})
\[
M_k = \bigcup_{i=1}^{N_k} M_{k,i}
\]
Thus, neural activations are \emph{group statistics}, shaped by many piecewise manifolds  acting together rather than by a single isolated random effect. This sum-based structure is fully consistent with the piecewise ``pancake'' geometry developed in Section 3.1, and it forms the basis for the error-bounding behavior described next.

\subsubsection{Union Bound (Boole’s Inequality) on Stacked Manifolds}
We begin with the classical union bound (Boole’s inequality).  
For any finite or countable family of events $\{A_i\}_{i=1}^N$,
\begin{equation}
\mathbb{P}\!\left( \bigcup_{i=1}^N A_i \right)
\;\le\;
\sum_{i=1}^N \mathbb{P}(A_i)
\label{eq:boole}
\end{equation}

In the Deep Manifold setting, \emph{each single piecewise manifold } $M_{k,i}$
corresponds to a single event:
\begin{equation}
A_i := \{\, h_k \notin M_{k,i} \,\}
\end{equation}

A deviation from the entire layer manifold is the union of deviations from its
individual slices:
\begin{equation}
\{\, h_k \notin M_k \,\}
\;=\;
\bigcup_{i=1}^{N_k}
\{\, h_k \notin M_{k,i} \,\}
\label{eq:union_deviation}
\end{equation}

Applying Boole’s inequality, Equation~\eqref{eq:boole} yields the correct stochastic
structure for a stacked piecewise manifold:
\begin{equation}
\mathbb{P}(h_k \notin M_k)
\;\le\;
\sum_{i=1}^{N_k}
\mathbb{P}(h_k \notin M_{k,i})
\label{eq:union_bound_manifold}
\end{equation}

This union-bound structure means that a deviation at depth $k$ is never a
single-path event, but a union of piecewise manifold level events. Each manifold contributes additively, not multiplicatively, to the total deviation probability. This stands in contrast to single-chain exponential, error models, which assume independent per-step drift on a global manifold. 

In a deep network, the activations are group statistics and the geometry is piecewise, so the deviation propagates additively across all piecewise manifolds. The inequality guaranties that even when each manifold piece fluctuates, the combined deviation remains bounded: errors cannot grow exponentially. The stacked piecewise  manifold disperses and dilutes deviations, providing an intrinsic geometric stabilizing effect. Furthermore, union bound affects fixed-point stability, Section~\ref{sec:nn_fixed_point}.

This structure stands in contrast to the assumptions behind LeCun's critique of autoregressive models. His argument presumes a single global manifold, a single chain of predictions, and a single per-token error probability. In that view, producing a sequence of length $n$ requires a success at $n$ independent steps, resulting in
\begin{equation}
P_{\mathrm{correct}}(n) = (1 - \varepsilon)^n
\end{equation}
The appeal of this formula lies in its simplicity, but its simplicity depends on assumptions that do not reflect the actual representational mechanics of deep networks. It assumes that each hidden state is governed by an individual statistic, that deviations propagate linearly along one trajectory, and that nothing in the model's internal geometry reshapes or corrects the evolving state.

In practice, none of these assumptions holds. Because activations are group statistics, not individual statistics, they do not inherit a single independent probability of error but a superposition of many partial probabilities. Because the underlying manifold is piecewise and stacked, representations do not travel along a unique path, but traverse overlapping geometries, where perturbations are absorbed or redirected. Because deviation events form a union, not a chain, the correct bounding principle is the Bonferroni inequality, not an exponential decay law.

The geometric character of the stacked piecewise manifold further reinforces this discrepancy. For a fixed layer $k$, as activations move through the \emph{stacked piecewise manifold depth} of 
the stacked manifold, they are repeatedly drawn toward semantically coherent regions 
$B_{k,j}$, producing an empirical contraction across stacked piecewise manifold:
\begin{equation}
    \mathbb{E}[e_{j+1}] 
    \;\le\; 
    \rho_{k,j}\,\mathbb{E}[e_j] + \xi_{k,j},
    \qquad 
    0 \le \rho_{k,j} < 1,
\end{equation}
where $e_j = d(h_{k,j}, B_{k,j})$ measures the deviation from the appropriate 
manifold region at stacked piecewise manifold depth $j$. This contractive behavior arises directly from 
the sum-based statistics and the stacked manifold structure. Rather than compounding 
errors, the network repeatedly re-centers its internal states within each layer, 
allowing deviations to diminish across slice depth.

For these reasons, the exponential-error argument cannot describe the behavior of modern neural networks. The model imagined by that argument, a single-path, independently drifting process, does not exist inside a deep network. What exists instead is a multi-layered, piecewise geometric system whose stochasticity is inherently distributed. Its deviations are governed by union-event bounds and corrected by manifold contraction, not magnified step by step. The phenomenon commonly described as self-correction is therefore not an accidental artifact of training but a natural consequence of the network's geometry and its group-statistical structure.

\subsection{Neural Network Fixed Point}
\label{sec:nn_fixed_point}
\begin{quote}
    \textit{Fixed point theory is the theory of iteration, until fixed}
\end{quote}
\subsubsection{Fixed Point Class Theory: Completing the Rigor of Calculus}
In the 1960s, the theory of fixed point classes \cite{kiang1980} emerged to address three fundamental questions in algebraic, integral, and differential equations. 

\begin{enumerate}
    \item Does a solution exist?
    \item Is the solution unique?
    \item Is the solution stable?
\end{enumerate}

\begin{figure}[H]
    \centering
    \includegraphics[width=0.65\linewidth]{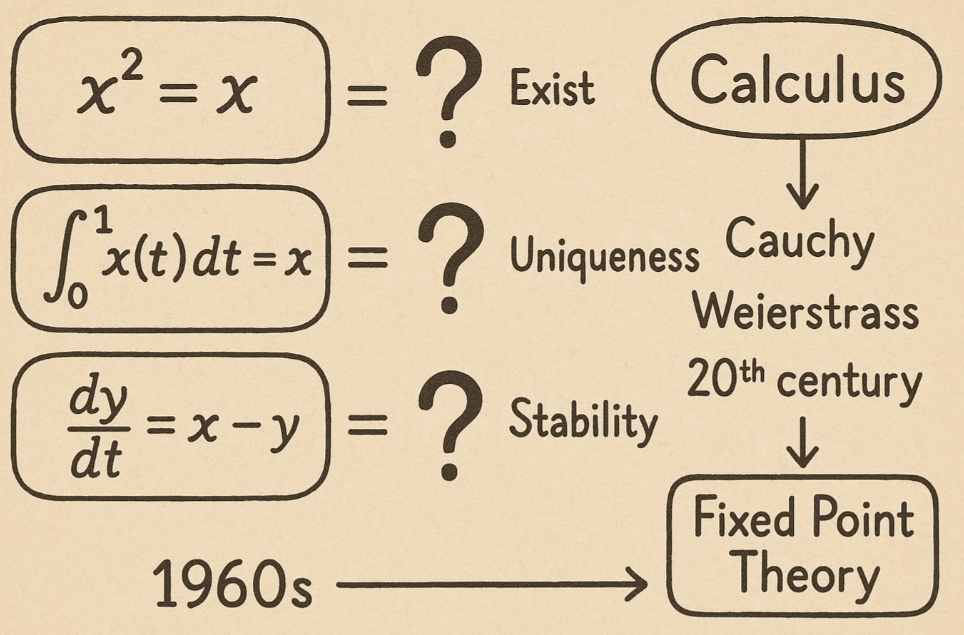}
    \caption{Theory of Fixed Point Classes}
    \label{fig:fixed_point_classes}
\end{figure}

Calculus underpins differential and integral equations through its core operations. Newton's calculus was powerful, but lacked full mathematical rigor. It wasn't until the development of real analysis, especially through Cauchy, Weierstrass, and later 20th century formalizations (e.g. measure theory, functional analysis, fixed point theory), that the foundations of calculus reached full rigor. The development of the theory of fixed point class marked one of the last milestones in establishing the full mathematical rigor of calculus in the 1960s, nearly 300 years later.

\subsubsection*{Fixed Point Theory on Stacked, Piecewise Manifolds}

The fixed point theory is iteration theory. Neural networks operate not on a single
smooth manifold, but on the stacked, piecewise manifold, Equation~(\ref{eq:piecewise-manifold}).
\[
\mathcal{M} = \bigcup_{i=1}^{N} \mathcal{M}_i.
\]

\paragraph{Piecewise Error Dynamics.}

\begin{figure}[H]
    \centering
    \includegraphics[width=0.6\linewidth]{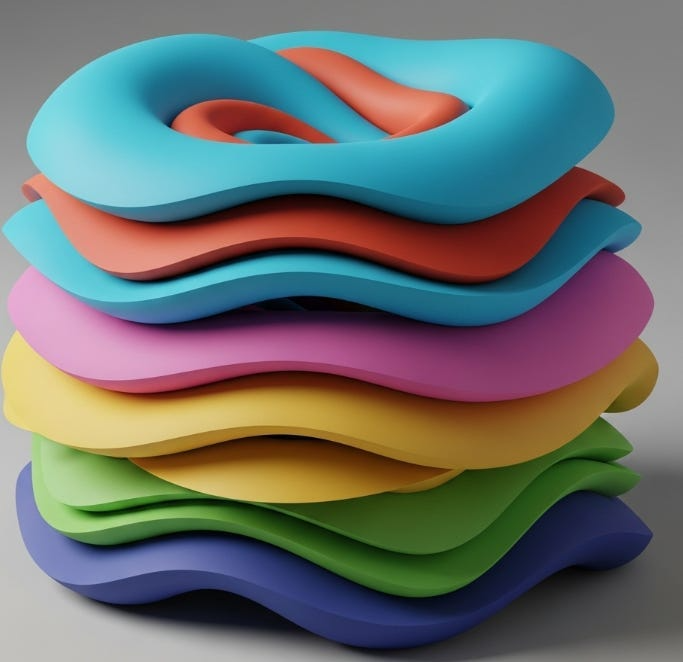}
    \caption{Pancake like Stacked Manifold}
    \label{fig:pancake_like_stacked_manifold}
\end{figure}

Within the piecewise manifold $\mathcal{M}_i$, the local iteration is
\[
h_{t+1} = f_i(h_t), \qquad h_t \in \mathcal{M}_i
\]
as in Equation (\ref{eq:representation-update}), 
Linearization near a fixed point at the piecewise manifold level $h_i^{*}$ yields
\begin{equation}
e^{(i)}_{t+1}
    = J_{f_i}(h^{*}_i)\, e^{(i)}_t 
      + O\!\left(\|e^{(i)}_t\|^2\right)
\label{eq:piecewise-linearization}
\end{equation}
while cross piecewise manifold transitions satisfy
\begin{equation}
e^{(j)}_{t+1}
    = J_{f_j \circ f_i}(h^{*})\, e^{(i)}_t 
\label{eq:cross-slice-jacobian}
\end{equation}
Global piecewise stability requires
\begin{equation}
\rho\bigl(J f_i(h_i^\ast)\bigr) < 1
\quad\text{for all piecewise manifolds } i
\label{eq:piecewise-contraction}
\end{equation}

\noindent
The spectral radius $\rho$ of the Jacobian being less than 1 is the standard condition for a contraction mapping (Banach Fixed Point Theorem).
 
\paragraph{Perturbation as Error Control.}
The neural iteration always includes the perturbation:
\begin{equation}
h_{t+1} = f_{i(t)}(h_t) + \delta_t 
\label{eq:perturbed-iteration}
\end{equation}
Thus,
\begin{equation}
e^{(i)}_{t+1}
    = J_{f_i}(h_i^{*})\, e^{(i)}_t + \delta_t 
\label{eq:piecewise-perturbation}
\end{equation}
and across slices,
\begin{equation}
e^{(j)}_{t+1}
    = J_{f_j \circ f_i}(h^{*})\, e^{(i)}_t + \delta_t 
\label{eq:cross-slice-perturbation}
\end{equation}

\paragraph{Perturbation Acceleration.}
The structured perturbation modifies each piecewise manifold operator:
\begin{equation}
f_{i,\varepsilon}(x)
    = f_i(x) + \varepsilon\, g_i(x)
\label{eq:structured-perturbation}
\end{equation}
with Jacobian
\begin{equation}
J_{f_{i,\varepsilon}}
    = J_{f_i} + \varepsilon J_{g_i}
\label{eq:perturbed-jacobian}
\end{equation}
Acceleration occurs when
\begin{equation}
\rho\!\left(J_{f_{i,\varepsilon}}\right)
    < \rho\!\left(J_{f_i}\right), 
    \qquad \forall i 
\label{eq:accelerated-convergence}
\end{equation}

\paragraph{Union-Bound Stability Across Slices.}
Since $e_t$ may lie on any piecewise manifold,
\begin{equation}
e_t \in \bigcup_i e^{(i)}_t 
\end{equation}
The probability of deviation obeys the union bound, Equation~\ref{eq:union_bound_manifold},
\[
P(e_t \notin \mathcal{M})
    \le \sum_{i} P\!\left(e^{(i)}_t \notin \mathcal{M}_i\right)
\label{eq:union-bound}
\]

\paragraph{Unified Stacked–Manifold Iteration.}
The general piecewise iteration is
\begin{equation}
h_{t+1}
    = f_{i(t)}(h_t) + \delta_t,
    \qquad h_t \in \mathcal{M}_{i(t)} 
\label{eq:stacked-iteration}
\end{equation}
A global fixed point satisfies the following requirements.
\begin{equation}
h^{*} = f_{i(k)}(h^{*}), \qquad \forall k .
\label{eq:stacked-fixedpoint}
\end{equation}
Convergence is guaranteed when
\begin{equation}
\rho\!\left(J_{f_i}\right) < 1 \ \text{for all slices},
\qquad
\sup_t \|\delta_t\| < \infty 
\label{eq:global-stability}
\end{equation}

\subsubsection{The Second-Order Derivative}

High-order nonlinearity is the default operating regime of neural networks. In
 stacked piecewise manifolds, the forward operator is
\begin{equation}
h_{t+1} = f_t(h_t)
\label{eq:fp-forward-operator}
\end{equation}
does not evolve on a single smooth surface but across many manifold slices.  
In such regimes, first-order derivatives provide only \emph{velocity}.  
They indicate the tangent direction on the current slice, but they do not encode 
curvature and, therefore, do not determine whether the iteration moves toward, 
away from, or orthogonally to the true fixed-point basin.

\paragraph{First-Order Limitation.}
Linearization on the piecewise manifold $M_i$ yields
\begin{equation}
e^{(i)}_{t+1} 
= J f_i(h_i^{\*})\, e^{(i)}_t + O(\|e_t\|^2)
\label{eq:fp-linearization}
\end{equation}
where convergence requires
\begin{equation}
\rho\!\left(J f_i\right) < 1
\label{eq:fp-spectral-radius}
\end{equation}
A first-order gradient step uses only
\begin{equation}
g_t = \nabla_\theta L(\theta_t)
\label{eq:first-order-gradient}
\end{equation}
which provides directional slope information but no curvature.  
In high-order nonlinear regions, this produces:
(i) misaligned updates across the piecewise manifold,  
(ii) oscillation, and  
(iii) slow or stalled convergence.

\paragraph{Second-Order Guidance.}
The second-order Derivative: AdamW (diagonal curvature), Adagrad-matrix (full 
accumulated curvature), Shampoo (matrix square-roots of layerwise curvature) 
and Muon—approximately the Hessian
\begin{equation}
H_t \approx \nabla^2_{\theta} L(\theta_t)
\label{eq:curvature-hessian}
\end{equation}
Mathematically, AdamW etc. do not approximate the Hessian directly; they approximate curvature information via moments. The correct convergence direction is the Newton-type update
\begin{equation}
\Delta\theta_t \propto - H_t^{-1} \nabla_\theta L(\theta_t)
\label{eq:newton-direction}
\end{equation}
.
\paragraph{Manifold-Corrected Update.}
Second-order optimizers produce the curvature adjusted step
\begin{equation}
\theta_{t+1}
= \theta_t - \eta\, G_t^{-1}\, \nabla_\theta L(\theta_t)
\label{eq:second-order-update}
\end{equation}
Here $G_t^{-1}$ denotes the optimizer-derived curvature proxy,
constructed from gradient moment statistics (AdamW, Shampoo, Adagrad-matrix, M\={o}UN).
The update induces an effective iteration of the Jacobian
\begin{equation}
J_{\mathrm{eff}}
= I - \eta\, G_t^{-1} H_t
\label{eq:effective-jacobian}
\end{equation}
and convergence requires
\begin{equation}
\rho(J_{\mathrm{eff}}) < 1
\label{eq:effective-convergence}
\end{equation}

\paragraph{Fixed-Point Interpretation.}
In the piecewise manifold $M_i$, the fixed point satisfies
\begin{equation}
h^{*} = f_i(h^{*})
\label{eq:slice-fixed-point}
\end{equation}
Second-order updates provide the correct orientation toward this fixed point 
because curvature determines the contraction direction on each piece of the 
stacked manifold. First-order methods supply only the velocity, while second-order 
methods supply the missing geometric information required for stable convergence.

In the Deep Manifold framework, curvature-aware optimization is therefore not 
an enhancement but a necessity: high-order nonlinearity and stacked piecewise 
geometry demand second-order information to achieve correct fixed-point 
orientation across piecewise manifolds.

\subsubsection{Two Type Iteration}

Boundary conditions play a decisive role in both classical analysis and neural  networks. In fixed point theory, they define the admissible region of motion  and orient the trajectory toward convergence. Without boundary conditions, iteration has no direction, much less an anchor; with them, iteration gains both scope and orientation.

In neural networks, boundary conditions emerge in two complementary iterative 
processes:

\begin{enumerate}
    \item \textit{Forward pass, inference.} 
    Each layer applies an iterative transformation to its input. In inference, the input or prompt of the inference, instructions, and the provided examples act as boundary conditions: they 
    delimit the representational space and orient the trajectory of the activations toward outputs consistent with the given context (Section~\ref{sec:iterated_integral}).

    \item \textit{Backward pass, backpropagation.}
    The reverse iteration of the error signals is similarly constrained. Architectural structure, normalization, skip connections, and gradient constraints regulate the propagation of gradients across layers, suppressing divergence or collapse, and allowing stable convergence toward fixed points.
\end{enumerate}

In both directions—forward activations and backward gradients—boundary 
conditions make the iteration meaningful. They constrain and orient the dynamical 
trajectory so that neural networks, like classical fixed point systems, can 
stabilize into coherent and useful representations.

\paragraph{Training vs.\ Inference: Two Coupled Fixed-Point Iterations.}

Training and inference are not separate mathematical processes; they are two 
fixed-point iterations operating in different spaces, each governed by its own 
boundary conditions.

\begin{enumerate}
    \item \textit{Training: boundary in parameter space}
    \begin{equation}
        \theta_{t+1} = \theta_t - \eta \nabla_\theta \mathrm{Loss}(\theta_t)
    \end{equation}
    The loss function defines the boundary of admissible parameter motion, 
    shaping the fixed point in~$\theta$.

    \item \textit{Inference: boundary in representation space}
    \begin{equation}
        h_{t+1} = f_t(h_t), 
        \qquad 
        h_t \in \partial\Omega_t(p)
    \end{equation}
    The prompt~$p$ defines the representational boundary, selecting the intrinsic 
    pathway through the stacked piecewise manifolds.
\end{enumerate}

Thus, both training and inference are realized as boundary-conditioned fixed 
point iterations over the same stacked manifold geometry—one evolving the 
parameters~$\theta$, the other evolving the activations~$h$. Their 
complementarity is structural: the forward iteration builds activations consistent 
with the prompt boundary, and the backward iteration shapes parameters consistent 
with the training boundary. Together, they form a \emph{dual fixed-point system} 
that governs the behavior of the model.

\subsubsection{$\mathbb{N}$ Fixed Points And Stacked Manifold}
\label{n_fixed_points}
 Due to the stacked nature of its manifold covers, the neural network possesses a unique ability to model the complex transfer of data between layers. Each cover acts as a local patch, capturing a different nonlinearity, and when composed together, they allow trajectories of data to flow, bend, and stabilize. Within this layered geometry, the network does not converge to a single universal state, but rather to a collection of  $\mathbb{N}$ fixed points, each representing a distinct mode of stability.

These fixed points embody how the network reconciles high-order nonlinearities: data entering different regions of the manifold may be absorbed into different approximators, yet all remain governed by the same global architecture. In practice, this manifests itself as multiple representation equilibria: stable embeddings, recurrent attractors, or converged weight configurations that coexist and interact. From the \textit{Deep Manifold} perspective, the multiplicity of fixed points is the very mechanism through which neural networks achieve generalization, interpolation, and emergent reasoning.

Neural networks can be viewed as layered compositions of smooth manifolds, like pancakes, progressively transforming complex data distributions into linearly separable forms. This perspective helps explain how deep architectures decompose nonlinearity into locally tractable structures. Convexity or non-convexity is defined with respect to a single, smooth global manifold. 

\begin{figure}[H]
    \centering
    \includegraphics[width=0.75\linewidth]{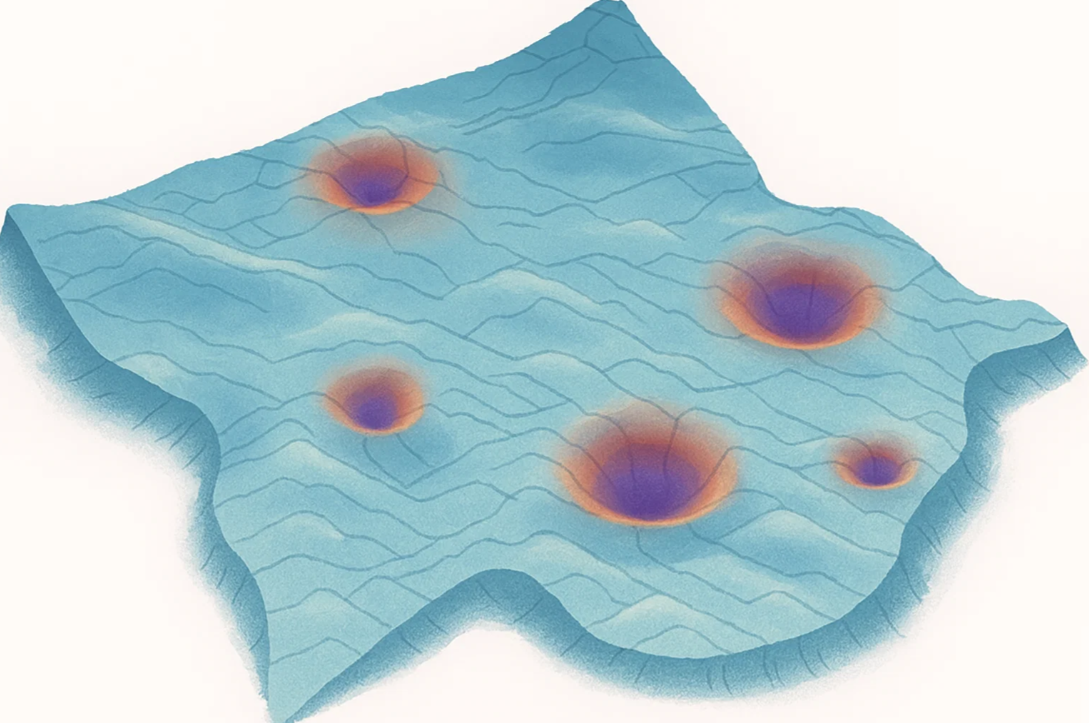}
    \caption{Foundation Model Fixed Point Landscape}
    \label{fig:foundation_model_fixed_point_landscape}
\end{figure}

Let the representational space of the network be the stacked union of
manifold pieces, Equation~\eqref{eq:piecewise-manifold}
\[
\mathcal{M} = \bigcup_{i=1}^{N} \mathcal{M}_i
\]
and let \(F_{\theta} : \mathcal{M} \to \mathcal{M}\) denote the trained forward
operator. The fixed-point set is defined by
\begin{equation}
\mathrm{Fix}(F_{\theta}) =
\{\, x \in \mathcal{M} \mid F_{\theta}(x) = x \,\}
\end{equation}

\begin{equation}
\mathrm{Fix}(F_{\theta})
= \{ x_1^{*}, \ldots, x_N^{*} \},
\qquad
x_i^{*} \in \mathcal{M}_i
\end{equation}

\begin{equation}
|\mathrm{Fix}(F_{\theta})| = N
\end{equation}

This expresses that a deep network operating in stacked piecewise manifolds
supports $N$ distinct fixed points, each tied to its corresponding manifold
region and each admitting its own convergence pathway during training and
inference.

However, neural networks implicitly construct and operate over many local manifolds, layered in a way that resembles a covering space in topology. This is part of their fundamental magic: they decompose a high-order, nonlinear global manifold into interconnected, smooth submanifolds, each more amenable to optimization and representation. Each layer incrementally transforms the geometry, unfolding the complexity into locally able structures—many of which exhibit convex behavior. As a result, deep networks operate in a hybrid regime: a mixture of convexity and non-convexity, dynamically balanced across depth and scale.

Pathways are formed through the iterated integral process, Section~ \ref{sec:iterated_integral} as Intrinsic Pathways, Section~\ref{sec:intrinsic_pathway} during training, where each update gradually constructs a path through the manifold. If no fixed point exists, the pathway remains uninitialized or evolves along the wrong trajectory. When fixed points do exist, the pathway has been properly initialized through iterated integration and aligned with the correct direction of convergence. This distinction helps explain the observed difference in prompt behavior between reasoning and non-reasoning models: reasoning models support a larger set of well-formed iteration pathways for a given prompt.

The next question is: what is the best prompt-engineering strategy? See blow.

\subsection{Neural Network Boundary Condition}
\label{sec:nn_boundary_condition}
\begin{quote}
    \textit{Boundary conditions give iteration purpose and direction.}
\end{quote}
\subsubsection{Three Type Boundary Condition of Backpropagation}
\label{three_model}
The Lagrangian formulation of Section~\ref{sec:lagrangian_formulation} 
established neural networks as fixed-point systems operating in stacked piecewise 
manifolds. The residual $f_\theta(x)-x$ defines the energy; 
architectural and data constraints enter through $g(\theta)=0$; and 
stationarity of the Lagrangian corresponds to a representational fixed point 
on each piecewise manifold. Because the geometry itself shifts during training, fixed points are not predetermined, they must be constructed through boundary-conditioned iteration.

Deep Manifold reveals that modern neural networks realize three types of 
fixed points, each determined solely by the boundary conditions imposed 
during the corresponding backward-pass regime. These stages are not 
different algorithms, they are different boundary structures guiding the 
same Lagrangian fixed-point mechanism.

\paragraph{Stage 0 --- Pre-Training: Weak Fixed-Point Iteration.}

Pre-training imposes a weak statistical boundary condition. 
\textit{The next-token objective provides only a weak statistical boundary, 
exemplified by the next-token conditional likelihood.}

\begin{equation}
\theta^{*} \approx 
\arg\min_{\theta} KL\!\left(p_{\text{data}} \,\|\, q_\theta\right)
\end{equation}

No explicit target geometry is supplied. Through iterated-integral updates 
over stacked piecewise manifolds, the system gradually constructs an initial 
fixed-point landscape. This stage forms the statistical envelope within which 
stronger boundaries will operate later.

\paragraph{Stage 1 --- Instruction Fine-Tuning: Intended Fixed-Point Iteration.}

Instruction fine-tuning introduces explicit intention-shaping boundary 
conditions. Unlike Stage 0, which is statistical, Stage~1 supplies clear 
targets that anchor the backward iteration toward a well-formed intended 
fixed point:

\begin{equation}
\theta_{t+1} 
= \theta_t - \eta_t \nabla_\theta L_{\text{SFT}}(\theta_t)
\end{equation}

These boundaries carve out stable attractor regions and align the manifold 
geometry with human-specified semantics. Convergence becomes purposeful: the 
fixed point becomes an intended fixed point, shaped by explicit instructions 
and example traces.

\paragraph{Stage 2 --- Reinforcement: Perturbed Fixed-Point Iteration.}

Reinforcement-style updates introduce weak and discrete boundary conditions. 
Rewards and preferences are sparse, low-magnitude signals that apply local 
perturbations on top of the manifold shaped by Stages~0 and~1. Let the weak 
boundary measure be:

\begin{equation}
d\mu_\varepsilon(x) 
= \sum_j \varepsilon_j \delta_{x_j}(x)
\end{equation}

and the accumulated perturbation functional:

\begin{equation}
B(\theta) 
= \sum_j \varepsilon_j \, r\bigl(\Phi_\theta(x_j)\bigr)
\end{equation}

The backward update becomes:

\begin{equation}
\theta_{t+1} 
= \theta_t - \eta 
\bigl[
\nabla_\theta L_{\text{SFT}}(\theta_t) 
+ \lambda \nabla_\theta B(\theta_t)
\bigr]
\end{equation}

yields a perturbed fixed-point iteration. These perturbations do not reshape 
the global manifold; they slightly tilt it at discrete points, refining the 
intended fixed point while preserving stability.

\paragraph{Unified Boundary-Condition Formula.}

Let:
\begin{enumerate}
\item $\alpha$ = implicit boundary; Stage 0: iteration of weak fixed-point
\item $\beta$  = sim-structured boundary; Stage 1: iteration of a fixed-point intended
\item $1-\alpha-\beta$ = explicit boundary; Stage 2: iteration of perturbed fixed-point 
\end{enumerate}

Then all three backward-pass regimes unify under a single fixed-point 
objective:

\begin{equation}
\theta^{*}
\in \arg\min_{\theta} 
\mathbb{E}_p
\Big[
\alpha \, KL(p_{\text{data}} \!\parallel q_\theta)
+ \beta \, C\!\left(\Phi_\theta(p)\right)
+ (1 - \alpha - \beta) \, \ell(q_\theta, y)
\Big]
\label{eq:unified_boundary_condition}
\end{equation}

\begin{table}[H]
\centering
\caption{Three stages of backpropagation Fixed Point Iteration.}
\vspace{1em}
\begin{tabular}{ll p{4.5cm} p{4.5cm}}
\textbf{\#} & \textbf{Stage Name} & \textbf{Boundary Type} & \textbf{Iteration Type} \\[2pt] \hline
0 & Pre-Training & Implicit boundary & Weak fixed-point iteration \\[4pt]
1 & SFT & Smi-Structured boundary & Intended fixed-point iteration \\[4pt]
2 & RL  & Explicit boundary & Perturbed fixed-point iteration \\
\end{tabular}
\end{table}

Across all stages, the network remains the same Lagrangian fixed-point system 
on stacked piecewise manifolds. Boundary conditions---weak, intended, or 
perturbed---determine which fixed point the backward iteration constructs:

\[
\text{Weak Fixed Point}
\;\longrightarrow\;
\text{Intended Fixed Point}
\;\longleftrightarrow\;
\text{Perturbed Fixed Point}.
\]

Although the three stages may operate interchangeably: especially 
Stage~1 and Stage~2, Stage~0 remains indispensable: it lays the statistical 
and geometric foundation upon which all subsequent fixed-point iterations 
depend. Without this foundation, neither intended nor perturbed fixed points 
can be stably formed.

\begin{figure}[H]
    \centering
    \includegraphics[width=0.9\linewidth]{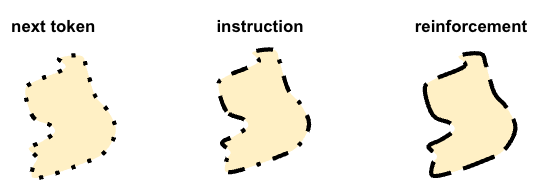}
    \caption{Foundation Model Boundary Conditions}
    \label{fig:foundation_model_boundary_conditions}
\end{figure}

\begin{quote}
    \textit{In this perspective, neural networks do not begin with fixed points; 
backpropagation builds them through stage-specific boundary conditions.}
\end{quote}

\subsubsection{Symmetric Boundary Conditions}
\label{sec:symmetric_boundary_conditions}

Symmetric boundary conditions yield the fastest and most stable form of fixed-point
iteration. This is especially important for neural networks because:
\begin{quote}
    \textit{Neural networks do not know where their fixed points are.}
\end{quote}

The fixed point is unknown, dynamic, and constructed during training. With no
predefined equilibrium or closed-form geometry, the boundary condition becomes
the sole mechanism that orients iteration on stacked piecewise manifolds.

\begin{figure}[H]
    \centering
    \includegraphics[width=0.75\linewidth]{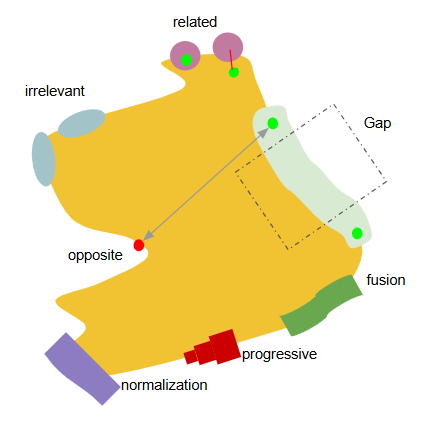}
    \caption{Symmetric Boundary Condition}
    \label{fig:symmetrical_boundary_condition}
\end{figure}

The objective of a unified boundary-condition (Equation~\ref{eq:unified_boundary_condition}) is

\[
\theta^{*} \in 
\arg\min_{\theta}
\mathbb{E}_{p}\!\left[
\alpha\, \mathrm{KL}(p_{\text{data}}\Vert q_{\theta})
+ \beta\, C(\Phi_\theta(p))
+ (1-\alpha-\beta)\,\ell(q_\theta,y)
\right]
\]

When the fixed point is unknown, the symmetric boundaries provide the following:
\begin{enumerate}
    \item balanced constraint from both sides,
    \item prevention of directional drift as the manifold moves,
    \item stability across stacked manifold slices,
    \item suppression of oscillation in high-order nonlinear regions,
    \item centered trajectories even under dynamic geometry.
\end{enumerate}

These properties create the narrowest and most reliable convergence corridor for
discovering a fixed point.

\textit{Contrastive learning} is the exact symmetric specialization of
Equation~(\ref{eq:unified_boundary_condition}). Setting

\begin{equation}
\alpha_{\text{pos}}=\alpha_{\text{neg}},\qquad
\beta=0,\qquad
1-\alpha_{\text{pos}}-\alpha_{\text{neg}}=0,
\end{equation}
reduces to the pure symmetric boundary pair expressed by the contrastive
objective:

\begin{equation}
L_{\mathrm{con}}
= -\mathbb{E}\!\left[
\log 
\frac{\exp(-d(h,h^{+}))}
{\exp(-d(h,h^{+})) + \sum_j \exp(-d(h,h^{-}_j))}
\right].
\end{equation}

The positive example forms an attraction boundary, while the negative example
forms a mirrored repulsion boundary. Together, they produce a two-sided symmetric
constraint, the canonical boundary condition for fast, stable iteration when
the fixed point is unknown and dynamic.

\subsubsection{Weak and Discrete Boundary Condition}
\label{sec:weak_discrete_boundary_condition}
In the Deep Manifold framework, a neural network does not begin with known fixed 
points. Unlike classical fixed point systems, where equilibrium location and 
geometric structure are predetermined, the neural network must \emph{discover} its 
fixed points entirely through training. With no prior knowledge of where these 
fixed points lie, the iteration cannot rely on internal geometry; it must be 
guided exclusively by boundary conditions.

Section~\ref{sec:symmetric_boundary_conditions} established that \emph{symmetric boundary conditions} provide the most stable and effective guidance when fixed-point locations are unknown.  However, symmetric boundaries require paired, balanced constraints 
:conditions rarely available in practical training. Most real-world signals are incomplete, 
asymmetric, or human-generated and therefore cannot provide the geometric symmetry 
necessary for the ideal convergence corridor.

When symmetric boundaries are not available, the iteration must rely on 
\emph{weak and discrete boundary conditions}. These appear most clearly in 
reinforcement-style updates: rewards are sparse, discrete, and low magnitude. 
Such signals do not impose strong geometric structure; instead, they produce 
\emph{minimal, non-disruptive adjustments} that preserve the manifold shaped by 
earlier training phases.

\begin{figure}[H]
    \centering
    \includegraphics[width=0.75\linewidth]{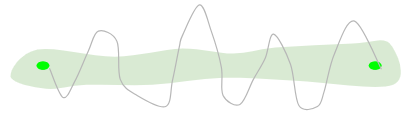}
    \caption{Weak and Discrete Boundary Condition}
    \label{fig:weak_discrete_boundary_condition}
\end{figure}

Figure~\ref{fig:weak_discrete_boundary_condition} illustrates this behavior: weak 
localized adjustments are superimposed on the existing manifold geometry. These 
signals act as pointwise nudges rather than continuous deformations 
, local corrections that accumulate gently across space without bending the manifold 
globally.

In the language of the Unified Boundary-Condition Formula (Equation~\ref{eq:unified_boundary_condition}):

\[
\theta^{*} \in 
\arg\min_\theta\;
\mathbb{E}_p\!\left[
\alpha\,\mathrm{KL}(p_{\rm data}\Vert q_\theta)
+\beta\,C(\Phi_\theta(p))
+(1-\alpha-\beta)\,\ell(q_\theta,y)
\right]
\]

RL-style boundary conditions correspond to a \emph{residual, low-magnitude term} 
added on top of the dominant pre-training and SFT components. Their role is not 
to reshape the manifold, but to perturb it slightly at discrete points where 
external signals exist.

This can be formalized as a \emph{boundary measure in space}. Let boundary 
signals occur at discrete points $\{x_j\}_{j=1}^N \subset \mathcal{M}$, each with 
small weight $\varepsilon_j$ and reward $r_j$. Define a weak, discrete boundary 
measure:
\begin{equation}
d\mu_\varepsilon(x) = \sum_{j=1}^N \varepsilon_j\, \delta_{x_j}(x)
\end{equation}

The accumulated weak boundary functional is then:

\begin{equation}
B(\theta)
=
\int_{\mathcal{M}} r\!\big(\Phi_\theta(x)\big)\,\mathrm{d}\mu_\varepsilon(x)
=
\sum_{j=1}^N \varepsilon_j\,r_j\!\big(\Phi_\theta(x_j)\big)
\end{equation}

This expression captures the idea from Figure~\ref{fig:weak_discrete_boundary_condition}: many tiny discrete boundary touches accumulating across the manifold. Each individual 
signal is weak, but together they produce a subtle global inclination in the 
iteration pathway.

The full training objective becomes

\begin{equation}
L_{\text{total}}(\theta)
=
L_{\text{base}}(\theta)
+
\lambda\,B(\theta)
\end{equation}

with update rule:

\begin{equation}
\theta_{t+1}
=
\theta_t
-
\eta\Big[
\nabla_\theta L_{\text{base}}(\theta_t)
+
\lambda\,\nabla_\theta B(\theta_t)
\Big]
\end{equation}

Here $L_{\text{base}}$ encodes the strong manifold structure formed by 
pre-training and SFT, while $B(\theta)$ represents the accumulated weak and 
discrete boundary contributions that gently adjust the network's behavior without 
destabilizing the geometry.

Thus, \emph{in the absence of known fixed-point locations} and \emph{in the 
absence of symmetric boundaries}, weak and discrete boundaries are the \emph{only 
effective choice}. They provide the minimal guidance necessary to refine the 
iteration trajectory without imposing excessive geometric distortion. Their 
strength comes from their weakness: they preserve the manifold, while still 
nudge the iteration in desirable directions. Only at the end may we note that such weak perturbations can influence 
reasoning-like behavior. But this is strictly a secondary effect. The primary 
mathematical function of weak and discrete boundaries is to provide gentle, 
spatially accumulated corrections when no stronger stabilizing structure is 
available within the Deep Manifold.

\subsection{Neural Networks Advance Mathematics}
\begin{quote}
    \textit{Neural networks advance mathematics—concretely through practice, subconsciously through theory—driven largely by non-mathematicians. }
\end{quote}
\subsubsection{Primitive Mathematics}

What makes neural network mathematics primitive is not a lack of structure,
but that every structure, like variables, operators, moves.
This shifting foundation violates nearly every instinct of classical mathematics. However, neural networks are not a black box: they rest on a solid mathematical foundation
built from \emph{piecewise manifolds, fixed-point theory, and iterated calculus}.

\begin{enumerate}
    \item The variables, coefficients, and even coordinates change. Everything is in flux. This is not how mathematicians are trained. It would be impossible for mathematicians to come up with such a design, see section \ref{dynamic_node_cover}.

    \item Mathematicians tread carefully around composite functions with more than two layers, wary of the many pitfalls, yet neural networks solve them effortlessly, almost nonchalantly, see Section \ref{sec:iterated_integral}.
    
    \item Solving the forward problem (positive time) and the inverse problem (negative time) together has always been a desire of mathematicians, but they have never known where to begin. However, neural networks tackle this problem naturally; see Section 3.1  \cite{deepmanifoldpart1}.

    \item In contrast to our traditional mathematics training, where abstraction typically moves upward to encompass wider ranges of real-world phenomena, neural networks seem to move in the opposite direction. Their abstraction runs downward, reducing complexity to the simplest algebraic unit: counting. see Section \ref{sec:counting}.

    \item The real world is inherently stochastic, shaped by probability and statistics. Stochasticity manifests as inequality,  outcomes are distributed, not equal. Neural networks capture this naturally and effectively, often without explicit design in the form of inequality, see Section \ref{sec:nn_stochastic}.

    \item High-dimensional, high-order nonlinearity is often treated as a complexity problem in classical mathematics. Neural networks, instead of predefined manifold formulations, implicitly build stacked piecewise manifolds.This bottom up construction contrasts with the top-down sophistication of traditional mathematical transformations, see Section \ref{sec:stretched_stacked_piecewise_manifolds}.

    \item The \textit{propertyless} nature of counting is precisely what gives neural networks tremendous power. Because counting itself does not carry intrinsic semantics, it provides a universal substrate for integration—allowing text, images, audio, and other modalities to be represented, mixed, and learned within a single model; see Section \ref{sec:propertyless}.
\end{enumerate}
Furthermore,  Neural networks have stacked covers, mathematically speaking, whereas the \emph{Numerical Manifold Method} typically uses only 3 to 4. In contrast, neural networks stack hundreds or even thousands of such covers,  never imagined anyone would take it that far.

\begin{figure}[H]
    \centering
    \includegraphics[width=0.6\linewidth]{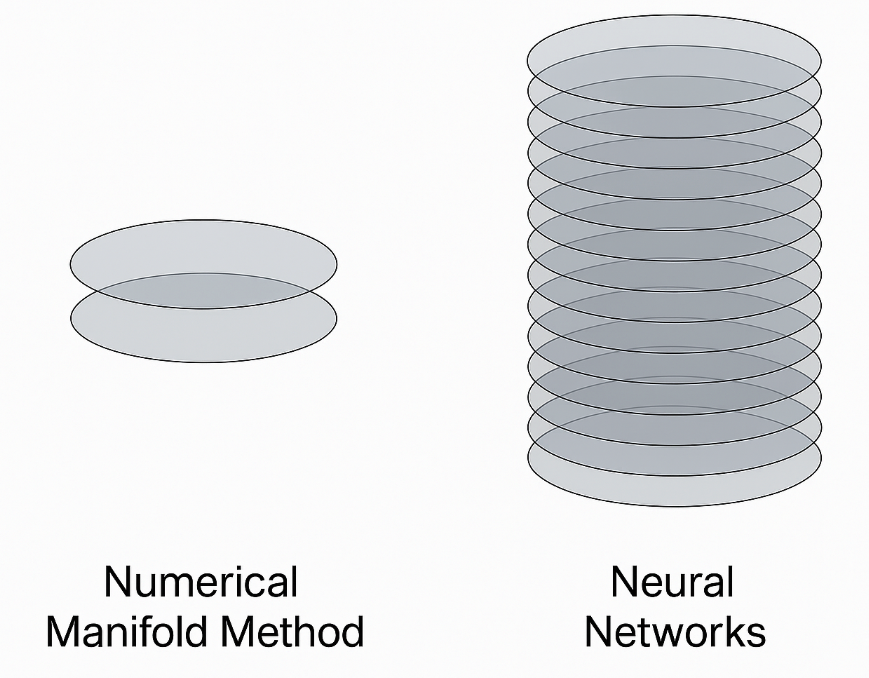}
    \caption{Manifold Node Covers}
    \label{fig:manifold_node_covers}
\end{figure}

In many ways, history repeats itself. Some of the most significant advances in mathematics have historically come from outside the discipline, pioneered by physicists, engineers, or other scientists, rather than pure mathematicians. Here are a few examples:

\begin{enumerate}

    \item Pierre de Fermat – Number Theory (1637).
    \item Blaise Pascal – Probability Theory (1654 ).
    \item Isaac Newton – Calculus (1666).
    \item János Bolyai – Non-Euclidean Geometry (1832).
    \item Claude Shannon – Information Theory (1948).
\end{enumerate}

Newton developed calculus to express the laws of motion, not to solve abstract math. Today’s AI revolution continues this tradition.

\subsubsection{Grand Calculus}
By examining neural networks through piecewise-smooth stacked manifold structures and fixed-point theory, we have opened a window into what was once a black box. Across geometry, algebra, equations, randomness, fixed points, and boundary conditions, we find that neural networks are not only grounded in solid mathematical foundations, but also expand and advance those foundations through practice, gradually forming what we call a grand calculus.

The most counterintuitive aspect of this grand calculus is that it is fundamentally approximate, not exact. Its power comes from continuous correction under data-defined boundary conditions, driven primarily by massive-scale cross-entropy. If classical calculus may be understood as a mathematical form of “divide and conquer,” then neural networks extend this idea to billions or even trillions of piecewise smooth, performing computations over stacked manifolds. This is the true strength of neural networks: a data-driven, massively scaled divide-and-conquer calculus that operates entirely through approximation.

Another unexpected discovery is that the activation values of a neural network are themselves propertyless. It is precisely this lack of inherent attributes that enables neural networks not only to maintain unified computability across different modalities but also to operate freely across mixed modalities: mirroring the real world, which is woven together from many such modalities. 

More importantly, this propertylessness provides the mathematical footing that allows neural networks to extend calculus. Without such an unpinned representation, differentiation and integration could not operate coherently across signals of different physical or semantic nature without inserting ad-hoc transformations.

This grand calculus opens an entirely new conceptual space, which we call the \textbf{Manifold Federation} based on the simplest concept of “divide and conquer”.  Building a complete AI world model remains our ultimate goal, and we believe that the “Manifold Federation” is the essential bridge connecting the real world to its computational world model.

\subsubsection{Deep Manifold: Mathematical Lineage }

The \emph{Numerical Manifold Method} (NMM, \cite{shi1991manifold}), grounded in differential geometry,  overcomes these limitations by employing connected, stacked, piecewise-smooth manifolds. On these piecewise-smooth patches, integration and partial differentiation remain valid, thereby preserving the computability of the overall system.

This mathematical progress of Numerical Manifold Method did not emerge from the quiet offices of a mathematics department but from the engineering realities of the construction of hydroelectric dams. There, engineers must confront the high-order nonlinearity and discontinuity inherent in rock foundations, deep tunnels, vertical shafts, and steep slopes, where many decisions are literally matters of life and death. These extreme practical demands forced the development of an entirely new mathematical structure.

As one of the key founders of the Theory of Fixed Point Classes in the late 1960s, co-author Gen-hua Shi combined nearly a decade of frontline engineering experience in dam construction with deep mathematical expertise to create NMM. Professor Shiing-Shen Chern, widely regarded as the father of modern differential geometry, served as one of the three members of the Shi PhD dissertation committee, providing academic validation for the mathematical framework of Shi and laying the foundation for the subsequent development of the theory.

The concept of the Deep Manifold arose from our attempt to understand neural networks through the lens of the Numerical Manifold Method, as discussed in Deep Manifold, Part I: Anatomy of Neural Network Manifolds. What surprised us was that, in real training data, high-order nonlinearity and discontinuity are nearly ubiquitous. Their intensity not only far exceeds what is observed in brittle rock masses but also exhibits even higher stochasticity.
\section{Neural Network: Learnable Numerical Computation}
\label{sec:computation_mathematics}
Neural networks are often portrayed as mysterious “black boxes,” but at their core they operate within the well-established tradition of numerical computation.  While mathematics seeks solutions that hold across all admissible domains; numerical computation seeks solutions that remain stable under concrete boundary conditions. Their difference lies in three operating principles:

\begin{enumerate}
    \item \textit{Approximation} --- replaces exact formulas with functions that are “close enough’’ within a bounded region of interest.
    \item \textit{Discretization} --- decomposes a continuous domain into finitely many slices where calculus remains applicable.
    \item \textit{Iterative correction} --- repeatedly adjusts the residual on these slices until the system settles into a tolerable fixed point.
\end{enumerate}

Together, these principles allow numerical computation to resolve problems that pure mathematics cannot express in a closed universal form.

Their flexibility lies in the ease with which nodes, parameters, and empirical components can be introduced, just as in numerical schemes. In this way, neural networks are not an alien paradigm, but a continuation of numerical practice applied to high-dimensional, nonlinear data.

\subsection{Lagrangian Fixed-Points as the Neural Network Equation}

Neural networks become numerically computable the moment their behavior is expressed as a \emph{fixed point residual wrapped in a Lagrangian form}, Equation~(\ref{eq:fixed_point_residual})
\[
f(x) - x = e(x), \qquad \min_{\,\theta} e(x)
\]
 The Lagrangian is the mechanism that makes this equation solvable in stacked piecewise manifolds. In Section~\ref{sec:lagrangian_formulation}, this structure is made explicit: the residual defines the energy, the constraints are entered as multipliers, and the saddle-point condition provides the stationarity needed for an iterative numerical method

Once expressed as a Lagrangian fixed point problem, Section~\ref{sec:nn_fixed_point}, the model behaves as any numerical system would: its updates become residual corrections, its gradients become constraint enforcement, and its convergence becomes the emergence of a stable fixed point on the manifold shaped during training. Backpropagation is simply the iterative correction step applied to this variational structure.

\subsection{Numerical Computation on Stacked, Piecewise Manifolds}
\label{sec:numerical_computation_stacked_piecewise_manifold}
Once the network equation is written in its Lagrangian fixed-point form, Section~\ref{sec:lagrangian_formulation} on a stacked piecewise manifold, Section~\ref{sec:piecewise_manifold_formalization}, its numerical structure follows directly. The key is that computation does \emph{not} occur on a single smooth domain, but on \emph{stacked, piecewise manifolds}, each acting as a local discretized region

A single manifold slice \( \mathcal{M}_k \) carries a local update driven by the fixed-point residual:
\begin{equation}
r_k(x) = f_{\theta,k}(x) - x 
\end{equation}
and the variational energy over that piecewise manifold is
\begin{equation}
E_k(\theta) = \int_{\mathcal{M}_k} \| r_k(x) \|^2 \, \mathrm{d}\mu_k(x)
\end{equation}

The full network behaves as a \emph{stacked numerical system}, aggregating these local energies:
\begin{equation}
E(\theta) = \sum_{k} E_k(\theta)
\end{equation}

The Lagrangian formation of Section~\ref{sec:lagrangian_formulation} now operates \emph{in a piecewise manifold}:
\begin{equation}
\mathcal{L}(\theta,\lambda) =
\sum_{k} 
\int_{\mathcal{M}_k} 
\| f_{\theta,k}(x) - x \|^2 \, \mathrm{d}\mu_k(x)
\;+\;
\lambda\, g(\theta)
\end{equation}

The stationarity conditions therefore become a \emph{distributed numerical constraint}:
\begin{equation}
\nabla_\theta L(\theta,\lambda) 
= \sum_k \int_{M_k} \nabla_\theta \|f_{\theta,k}(x) - x\|^2\, d\mu_k(x)
+ \bigl(\nabla_\theta g(\theta)\bigr)^\top \lambda = 0
\end{equation}

\begin{equation}
\nabla_\lambda \mathcal{L}(\theta,\lambda) = g(\theta) = 0
\end{equation}

These equations make the connection explicit:

\begin{enumerate}
    \item each piecewise manifold acts like a discretized region;
    \item each local residual is an approximation error;
    \item the integral over each piecewise manifold is the numerical assembly step;
    \item the summation over piecewise manifold is the global coupling;
    \item and the gradient update is the iterative correction step.
\end{enumerate}

Thus, the network computes by repeatedly applying the following:
\begin{equation}
\theta^{(t+1)} 
= 
\theta^{(t)} 
- 
\eta 
\left(
\sum_{k} 
\int_{\mathcal{M}_k}
\nabla_\theta \| f_{\theta,k}(x) - x \|^2 \, \mathrm{d}\mu_k(x)
\right)
\end{equation}
which is simply residual-driven numerical iteration carried out on stacked, piecewise manifolds.

Training sculpts the geometry so that the fixed point becomes reachable; inference traverses this geometry to settle into the corresponding numerical solution.

\subsection{Galerkin Method Analogy in Neural Networks}
\label{sec:galerkin_method}
The Galerkin method is one of the foundational principles of modern numerical computation. Its power comes from a simple idea: instead of solving a differential equation in its exact analytic form, one expresses the problem in a variational (weak) form and requires the residual to be orthogonal to a chosen function space. This converts an otherwise intractable continuous problem into a numerically solvable one, where approximation, discretization, and iterative correction become both natural and effective. Finite Element Methods, spectral methods, and most modern PDE solvers are direct descendants of this principle; their structure is built on the insight that \emph{numerical solvability emerges from a variational residual formulation}, Section~\ref{sec:iterated_integral} (iterated integrals resemble numerical assembly).

The connection to neural networks is structural rather than literal:

\begin{enumerate}
    \item \emph{Residual correspondence} --- the neural residual \(f_\theta(x) - x\) mirrors the PDE residual in variational formulations.
    \item \emph{Discretized geometry} --- stacked piecewise manifolds function like discretized regions in numerical methods.
    \item \emph{Piecewise integration} -- integrals over each manifold piece resemble the numerical assembly of local contributions.
    \item \emph{Iterative correction} --- gradient-based updates play the role of residual-driven correction steps.
    \item \emph{Learned basis} --- unlike classical Galerkin, the basis, operator, and test functions are not prescribed, but learned during training.
    \item \emph{Overparameterization as fine-grained discretization} --- in numerical computation, a finer mesh or more basis functions expand the function space in which the residual is minimized. Neural networks behave in the same way: more parameters mean more learned basis functions and finer discretization of the manifold. What deep learning calls ``overparameterization'' is numerically just a \emph{higher resolution discretization}, making residual reduction easier and fixed-point convergence more stable.
\end{enumerate}

In neural networks, there is no explicit test-function space. The network implicitly learns its own basis and its own ``test functions'' through iteration. This is why the analogy to Galerkin is structural, not literal: the role of the test function, probing residual consistency, exists, but it is learned, not prescribed.

In this setting, the \emph{basis} corresponds to the activation-generated functions
\begin{equation}
\phi_i(x;\theta) = \sigma(w_i^\top x + b_i)
\end{equation}
which shift every iteration as the node covers and coordinates evolve. The \emph{operator} corresponds to the layer mapping
\begin{equation}
\mathcal{F}_\theta : x \mapsto \sigma(Wx + b)
\end{equation}
or, globally, the stacked composition
\begin{equation}
\mathcal{F}_\theta^{(L)} = f_{\theta_L} \circ \cdots \circ f_{\theta_1}
\end{equation}
which transforms one manifold piece into the next. Both the basis and the operator are learned---not prescribed---reinforcing that neural networks extend the Galerkin principle by making its components learnable while preserving its variational--residual--iteration structure.

In this sense, neural networks extend the Galerkin principle by making the basis, the operator, and the effective test-function space learnable.

\subsection{Numerical Stabilization: Dropout, Skip Connections, and Normalization}
\label{sec:numerical_stabilization}
Interpreting neural networks as learnable Galerkin systems clarifies the role of several key architectural mechanisms, Section~\ref{sec:galerkin_method}. In classical Galerkin methods, the basis, operator, and test-function space are fixed; numerical stability is guaranteed by construction. Neural networks, however, learn all three through iteration, which requires explicit stabilizers to maintain a coherent variational--residual--iteration process.

\begin{enumerate}
    \item \emph{Dropout as a perturbation of the learned basis} \\
    Classical methods rely on fixed basis functions whose stability never changes. Neural networks instead learn the basis
    \begin{equation}
    \phi_i(x;\theta)
    \end{equation}
    whose orientation changes with each iteration. Dropout introduces controlled stochastic perturbations that prevent basis collapse, maintain elasticity across manifold pieces, and ensure stable residual correction.

    \item \emph{Skip connections as preservation of residual structure} \\
    Variational and residual formulations are central to Galerkin-style computation. Skip connections explicitly encode the residual form:
    \begin{equation}
    x_{l+1} = x_l + f_{\theta_l}(x_l)
    \end{equation}
    turning each layer into a residual update step. This stabilizes deep iterative correction, reducing numerical drift, and enabling consistent updates across many stacked piecewise manifolds.

    \item \emph{Normalization as stabilization of function-space geometry} \\
    Galerkin methods assume a well-behaved function space with consistent norms. Because neural bases and operators evolve during iteration, this geometry must be actively controlled. LayerNorm, RMSNorm, and related methods restore consistent local norms, ensuring that residual evaluation and correction occur in a stable manifold neighborhood.
\end{enumerate}

Together, these mechanisms provide the stability that classical Galerkin systems automatically obtain from fixed functional components. Dropout maintains the elasticity of the base, skip connections preserve the residual form, and normalization stabilizes the local geometry of the learned function space. Their presence reinforces the view that neural networks extend the Galerkin principle by making its components learnable while requiring explicit stabilizers to support iterative residual minimization on stacked piecewise manifolds to delay plasticity, Sections 3.6 \cite{deepmanifoldpart1} and~\ref{sec:plasticity_source}.

\subsection{Babuška's Paradox}
\label{sec:Babuska_paradox}

Babuška (1962) showed that a simply supported circular plate, when approximated by regular 
polygons, exhibits \emph{increasing} error as the number of sides grows. Although the polygonal 
boundary converges geometrically to the circle, the finite-element solution converges to a 
\emph{different} limit, because each polygon imposes boundary constraints that are not equivalent 
to those of the circular simply supported edge. Thus, geometric refinement changes not only the 
boundary but also the variational manifold itself, causing the solution to drift away from the intended 
problem.

From the Deep-Manifold viewpoint, this pathology arises from the \emph{single, globally defined  manifold} in classical FEM: geometry and function approximation share the same discretization. Any perturbation of the boundary therefore perturbs the entire manifold and alters the effective boundary conditions.

The new method based on the numerical manifold method (NMM, \cite{shi1991manifold}) 
prevents this mechanism by operating in \emph{stacked and overlapping piecewise manifolds} ~\cite{Zheng2013}. The \emph{mathematical cover (MC)} provides a high-quality approximation manifold independent of geometry, and changes the boundary geometry do \emph{not} modify the underlying approximation manifold. Consequently, the problem definition remains stable and convergence is preserved, even for highly curved or nonpolygonal boundaries.

In essence, Babuška’s paradox reveals the fragility of single-manifold discretizations. The NMM  stacked piecewise manifolds , Section~\ref{sec:piecewise_manifold_formalization}, fundamental to the Deep-Manifold framework, provide stacked piecewise numerical stability, Section~\ref{sec:numerical_computation_stacked_piecewise_manifold}, remove this coupling, 
ensuring that geometric refinement does not change the mathematical problem itself.

\begin{quote}
    \textit{Just as in the supercomputing era of the 1970s–1990s, where scaling saturated despite rapidly increasing computing power, this example shows the same structural ceiling: classical numerical methods, lacking any real understanding of high-order nonlinearity, could not handle large displacement or strong nonlinear behavior, and increasing FEM node counts merely amplified geometric and variational inconsistencies rather than resolving them.}
\end{quote}
\section{Data}
\label{sec:data}
\begin{quote}
    \textit{No one seems to ask the most basic question in the simplest mathematical terms: is it linear, low-order nonlinear, or high-order nonlinear? What about data complexity ? }
\end{quote}
\subsection{The Mathematical Nature of Real-World Data}
As discussed in Section~\ref{sec:nonlinear_complexity}, real-world data have some linear, never low-order, but are mostly high-order nonlinear, high-dimensional and nearly infinite in scope and scale, properties that fundamentally shape how neural networks must learn from it.
\subsubsection{High Order Nonlinearity}
We believe that AI training data are inherently high-order nonlinear, characterized by recursion, long-range dependencies, symbolic discontinuities, and deeply compositional structures. In Our Part 1 Paper, \cite{deepmanifoldpart1}, we provide a mathematical definition of high-order nonlinearity. The figure below illustrates a simple classification of linear, low-order nonlinear, and high-order nonlinear cases. 
\begin{figure}[H]
    \centering
    \includegraphics[width=0.7\linewidth]{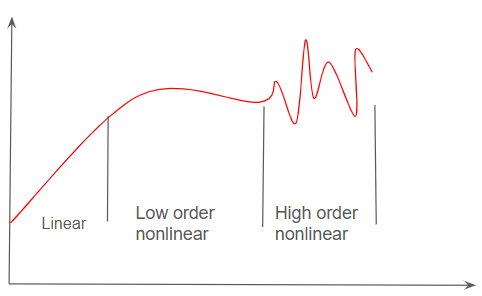}
    \caption{Linear, Low-order Nonlinear and High-order Nonlinear}
    \label{fig:nonlinear_curve}
\end{figure}

Take the image of the U.S. national flag as an example: we see sharp color jumps between stripes, and between the stars and the blue background. Such abrupt changes can be considered high-order nonlinearities

\begin{figure}[H]
    \centering
    \includegraphics[width=0.7\linewidth]{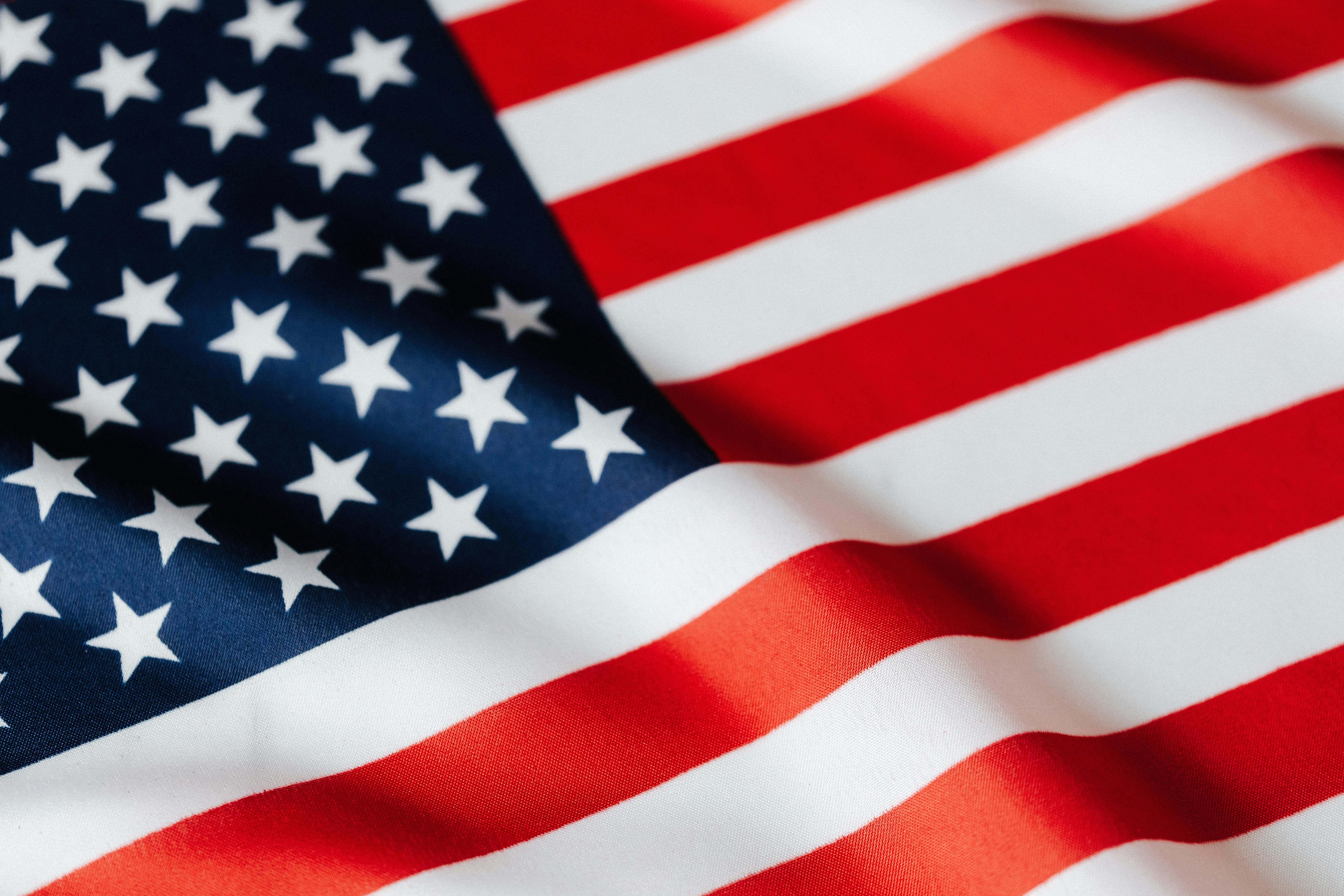}
    \caption{US National Flag, Discontinued Color}
    \label{fig:usa_flag}
\end{figure}
We can take a more technical approach to understanding data relationships. For images, applying different filters will reveal that if no high-order nonlinearity is present, the high-order nonlinearity filter returns none. The figure below illustrates this.
\begin{figure}[H]
    \centering
    \includegraphics[width=0.9\linewidth]{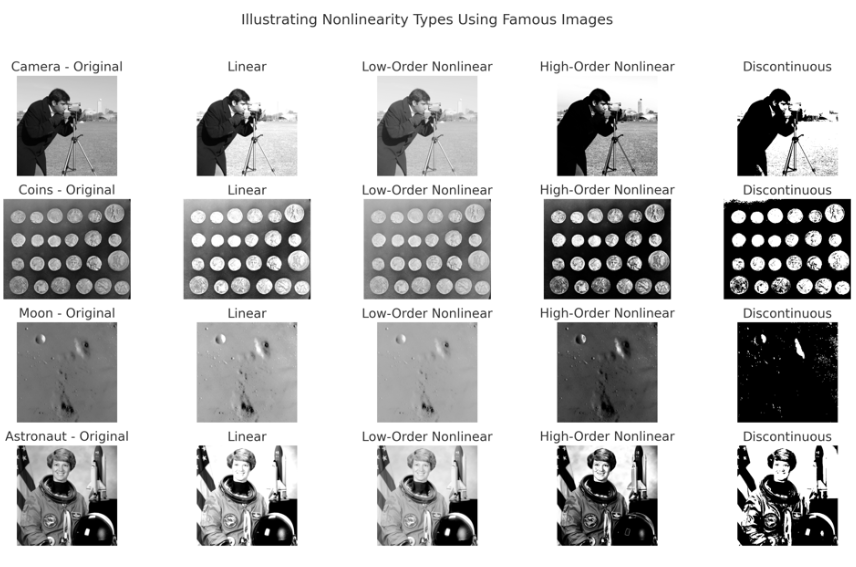}
    \caption{Linear and Nonlinear}
    \label{fig:image_nonlinearity}
\end{figure}

For language and text, see the example below.
\begin{table}
\centering
\caption{NLP Nonlinearity}
\label{tab:relationships}
\begin{tabularx}
{\textwidth}{@{} 
  >{\hsize=0.54\hsize}X
  @{\hspace{10pt}}
  >{\hsize=1.00\hsize}X
  @{\hspace{10pt}}
  >{\hsize=1.46\hsize}X 
@{}}
\toprule
\textbf{Type} & \textbf{Example Sentence} & \textbf{Relationship Description} \\
\midrule
Linear & ``More sugar makes it sweeter.'' & Direct, proportional relationship between input and output. \\
\addlinespace 
Low Order Nonlinearity & ``A little wine relaxes, too much ruins the night.'' & Smooth, curved effect --- like a quadratic or saturating response. \\
\addlinespace
High Order Nonlinearity & ``I never said she stole the money.'' & Meaning changes based on multi-token interaction or emphasis. \\
\addlinespace
Discontinuous & ``Not bad'' means ``good.'' & Small token change causes sudden semantic shift or inversion. \\
\bottomrule
\end{tabularx}
\end{table}

\subsubsection{Nonlinear Complexity}
\label{sec:nonlinear_complexity}
After reviewing thousands of AI papers published over the past five years, we found that data high order nonlinearity is almost never explicitly discussed but instead is often subsumed under the term ‘complexity"’ or “high order”. While high dimensionality is widely acknowledged, it is high-order nonlinearity that defines the true complexity of AI datasets. When you break down complexity in terms of time, dimension, and nonlinearity, nonlinearity appears to play a critical role, see below.

\begin{table}
\centering
\caption{Complexity Breakdown}
\begin{tabularx}{\textwidth}{@{} l X @{}}
\toprule
\textbf{Concept} & \textbf{Complexity Theory View} \\
\midrule
Time & How long an algorithm takes (time complexity) \\
\addlinespace 
Dimension & Higher input/state space dimension $\rightarrow$ more computation \\
\addlinespace
Nonlinearity & Nonlinear systems are often harder to analyze and solve \\
\bottomrule
\end{tabularx}
\label{tab:complexity}
\end{table}

From a purely mathematical perspective, linear functions form the baseline of complexity: their structure is simple, predictable, and solvable with cost proportional to input size n. Low-order nonlinear functions extend this baseline by introducing polynomial terms of finite degree k, whose complexity grows as $O(n^k)$. High-order nonlinear functions require unbounded expansions or exhibit chaotic, multi-scale behavior, placing them beyond any fixed polynomial class. Discontinuous functions cannot be globally approximated by polynomials at all, demanding piecewise covers that may scale exponentially with $n$. Thus, functional complexity can be indexed as the deviation from the linear case, with each class reflecting a progressively harder mathematical structure.

\begin{table}
\centering
\caption{A Classification of Function Complexity}
\label{tab:function_complexity}
\begin{tabularx}{\textwidth}{@{} l l c X @{}}
\toprule
\textbf{Class} & \textbf{Definition} & \textbf{Complexity index $k(f)$} & \textbf{Scaling with input size $n$} \\
\midrule
L (Linear) & $Ax + b$ & $0$ & $O(n)$ \\
\addlinespace
P (Low-order polynomial) & degree $k$ & $k$ & $O(n^k)$ \\
\addlinespace
H (High-order nonlinear) & infinite expansion & $\infty$ & grows faster than any fixed $O(n^k)$ \\
\addlinespace
D (Discontinuous) & piecewise / jumps & $\perp$ & exponential partitions possible \\
\bottomrule
\end{tabularx}
\end{table}

\begin{figure}[H]
    \centering
    \includegraphics[width=0.9\linewidth]{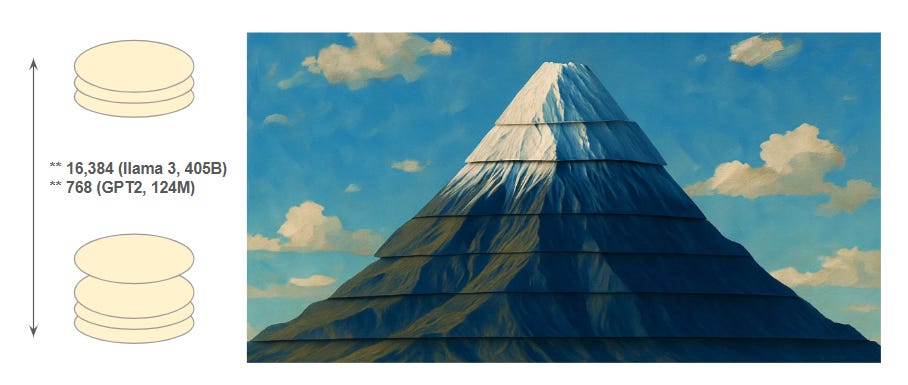}
    \caption{Chain Of Thought}
    \label{fig:sliced_mountain}
\end{figure}

The geometry of real-world data resembles a steep, irregular mountain: a global manifold
characterized by high-order nonlinearity, rapidly varying curvature, and discontinuous
features. Neural networks do not approximate this mountain as a single smooth surface.
Instead, they represent it through \emph{stacked, overlapping piecewise manifolds}, analogous
to slicing the mountain horizontally into locally regular layers. Each slice is smooth and
low-order, while the global structure emerges from their coordinated assembly.

This construction yields two structural consequences. First, decomposition into $s$ slices
reduces the effective nonlinear degree. A global complexity of order
\begin{equation}
    \mathcal{C}_{\text{global}} \sim  O\!\left(n^{k}\right)
\end{equation}

could be heuristically replaced by $s$ local components of complexity;
\begin{equation}
\mathcal{C}_{\text{stacked}} \sim  s\,O\!\left(n^{k/s}\right)
\end{equation}
The global, high-order geometry is thereby reconstructed from locally tractable pieces.
Second, the overlap requirement inherent to good-cover structures necessitates a
\emph{high-dimensional embedding space}. The high dimension does not correspond to a
single monolithic manifold; it provides the ambient capacity for many low-order slices to
coexist, overlap, and align without geometric interference.

\paragraph{Mathematical Representation of Nonlinear Complexity}
Nonlinear complexity refers to the intrinsic difficulty arising from the high-order, discontinuous, and multi-scale nonlinear structures of real-world data. We noted in above that the real world is not governed by simple smooth low-order nonlinearities; instead, it contains:
\begin{enumerate}
    \item high-order nonlinear transitions,
    \item piecewise smooth but globally fractured geometry,
    \item local discontinuities,
    \item rapid variation in curvature across scales.
\end{enumerate}

Such data cannot be captured by a single global function; they require \emph{overlapping piecewise manifolds}, each approximating locally but collectively forming a high-order nonlinear structure.

Formally, if the underlying data manifold $\mathcal{M}$ is decomposed into smooth pieces $\{\mathcal{M}_i\}$, then nonlinear complexity reflects the variation of curvature, discontinuities, and transition structure between these pieces:
\begin{equation}
\mathcal{C}_{\mathrm{nonlinear}}
=
\sum_{i}
\left(
    \|\nabla^{2} f\|_{\mathcal{M}_i}
    +
    \Delta_{\partial \mathcal{M}_i}
\right)
\label{eq:nonlinear_complexity}
\end{equation}
where $\nabla^{2} f$ measures local curvature (high-order nonlinearity) and $\Delta_{\partial \mathcal{M}_i}$ measures discontinuity across boundaries.

\subsection{Intrinsic Pathway}
\label{sec:intrinsic_pathway}
The forward pass of the neural network is an iterated-integral process (Section~\ref{sec:iterated_integral}). Section \ref{sec:nn_boundary_condition} introduces the integral boundary conditions. The remaining question then is: \textit{where is the integral path?}

\begin{figure}[H]
    \centering
    \includegraphics[width=0.8\linewidth]{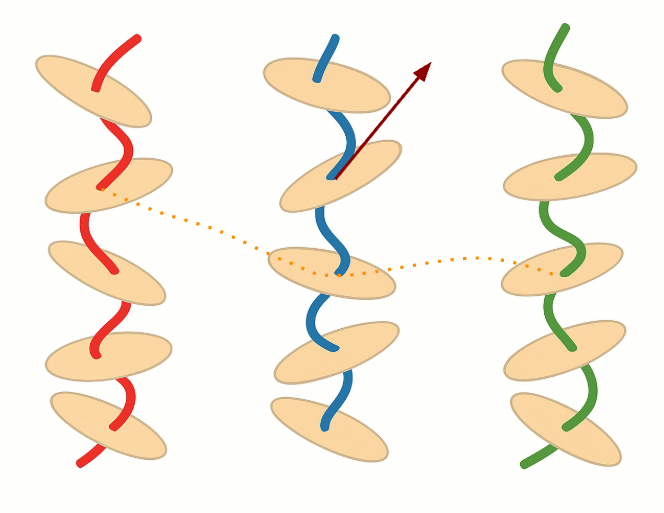}
    \caption{Piecewise Manifold Dimensionality and Intrinsic Pathway}
    \label{fig:piecewise_manifold_dimension}
\end{figure}

In large language models, token embeddings embody exactly this structure. Their high
embedding dimension simultaneously supports the overlap geometry of stacked piecewise
manifolds and reduces the effective nonlinearity of each local representation. Thus,
expressive power arises not from modeling one globally complex manifold but from
organizing many overlapping, locally low-order manifolds within a high-dimensional
ambient space.

This illustration (Fig.~\ref{fig:piecewise_manifold_dimension}) captures the geometry at three given activations. Conventional views treat each embedding node as encoding a targeted feature; in the Deep Manifold view, it is instead a \emph{feature bit} \cite{deepmanifoldpart1}, Section~3.7). The tangent direction (dark-brown arrow) on the central activation curve represents the \emph{local dimension}, sub dimension, required to for full dimension to model high-dimensional data. The dashed connectors between the piecewise manifolds form the network’s \emph{intrinsic pathway}, the actual iterated-integral trajectory taken through the model.

\begin{quote}
    \textit{Inference input or prompt instruction serves as the boundary condition of this iterated integral, determining which intrinsic pathway is activated.}
\end{quote}

\subsection{Emergent Properties And Multi-Turn Intrinsic Pathway}
Large models contain more stacked piecewise manifolds and, therefore, more potential intrinsic pathways. With greater training data, the iterated integral process in Section \ref{sec:iterated_integral} has more opportunities to form stable pathways across the overlapping piecewise manifolds described in Sections \ref{sec:data_complexity} and \ref{sec:intrinsic_pathway}. Each additional data fragment refines these overlaps, reducing the effective nonlinearity locally and increasing the number of viable routes through the model. This accumulation of intrinsic pathways is the structural reason larger models exhibit emergent behaviors: the geometry allows more trajectories, and training pressure gradually stabilizes them into usable fixed-point routes.

\vspace{1em}
\begin{figure}[H]
    \centering
    \includegraphics[width=0.75\linewidth]{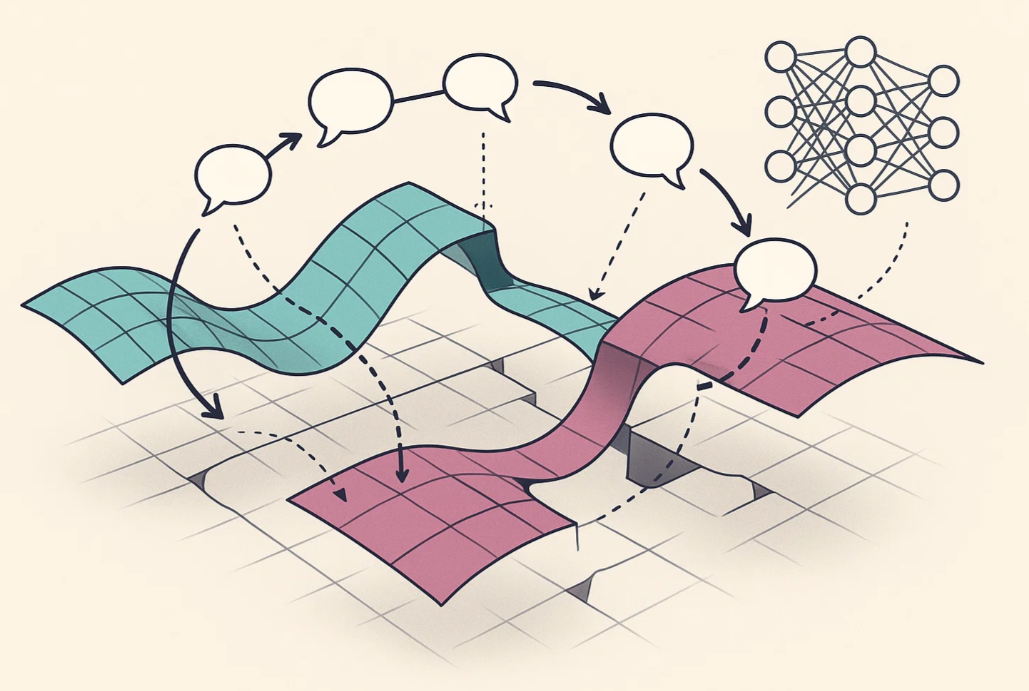}
    \caption{Multi-Turn Intrinsic Pathway (Chain-of-Thought)}
    \label{fig:chain_of_thought}
\end{figure}

In this view, Chain-of-Thought is not an external skill but a boundary condition for the iterated integral itself, Section~\ref{sec:iterated_integral}. The instruction sequence constrains which intrinsic pathway is activated, determining how the model moves across the stacked piecewise manifolds during inference. As models scale and training data deepen these pathways, the prompt’s boundary condition becomes increasingly effective in selecting long, coherent trajectories: what we call reasoning. Emergent properties arise naturally once the piecewise-manifold structure contains enough intrinsic pathways for the boundary condition to “lock onto’’ a stable iterated integral route.

\subsection{Intrinsic Dimension}
\label{sec:Intrinsic Dimension}

Discussion of dimensionality requires a precise specification of the underlying
space. Neural networks operate simultaneously in two distinct notions of
dimension. The \emph{ambient dimension} is determined by the chosen
representation space, for example, RGB values in $\mathbb{R}^3$ or token
embeddings in $\mathbb{R}^m$. By contrast, the \emph{intrinsic dimension}
reflects the degrees of freedom present in the data itself. A uniform image,
although embedded in $\mathbb{R}^3$, possesses zero intrinsic dimension; more
generally, the intrinsic dimension is content-dependent and does not need to match the
ambient coordinate size.

The intrinsic dimension interacts directly with the boundary conditions imposed by
the data or the prompt. Together, they define the \emph{learning space} 
(Part~1 \cite{deepmanifoldpart1}, Section~3.4), the effective region of the manifold within which
iteration can proceed. A smaller learning space, i.e., lower intrinsic 
dimension under well-posed boundary conditions, reduces geometric variability 
and, therefore, simplifies learning.

This distinction aligns with dimension theory in topology, in which an 
$m$-dimensional region may be described by a good cover whose intersections have
maximal depth:
\begin{equation}
\text{dimension } m \;\Rightarrow\; \text{at most } m+1 \text{ sets overlap}
\end{equation}
The quantity $m+1$ characterizes the overlap structure rather than the number of
covering sets. In the Deep Manifold framework, stacked piecewise manifolds 
function analogously as overlapping covers: locally low-order representations
whose boundary interactions encode the geometry of the underlying 
high-dimensional data.

The high dimensionality itself is not the principal source of complexity. A
nonlinear curve embedded in a large ambient space possesses a well-defined
tangent direction, and this tangent provides the effective local dimension 
governing the computation. Consequently, the primary difficulty arises from 
high-order nonlinearity, not from the size of the ambient space. Once a neural
network learns this nonlinear structure, it simultaneously acquires the
dimensional organization induced by that structure.

For this reason, large language models rarely encounter the classical \enquote{curse of
dimensionality}. High ambient dimension supplies representational capacity,
while learning progresses along low-dimensional intrinsic pathways determined by nonlinear geometry.

Because the activations are propertyless, the network is free to express both high-order nonlinearity and high-dimensional tangent structure in the same computational space. This is why high order and high dimension become naturally unified.

It can be written as:

\[
\text{High-order nonlinearity}
\;\Longleftrightarrow\;
\text{High-dimensional tangent structure}
\]

or

\begin{equation}
\dim_{\mathrm{effective}}(\mathcal{M})
=
\mathrm{rank}\!\left(
\frac{\partial^{k} f}{\partial x^{k}}
\right)
\end{equation}

This expresses that the effective dimension is determined by higher-order
derivatives (the curvature structure), rather than by the input
dimensionality.

\subsection{Semantic–Symbolic Dual Representation}
\label{sec:semantic_symbolic}
It is important to understand the relationship between symbolic and semantic data. In real-world settings, most enterprise data are tabular, hence symbolic, yet they must be reconciled with the dominant semantic world of ChatGPT

\subsubsection{Semantic-Symbolic Pairing}
\begin{quote}
    \textit{Semantic–symbolic data can be regarded as a data-scale issue; see Data Complexity, Section~\ref{sec:data_complexity}}
\end{quote}
Symbolic data can be understood as discrete values, such as a cell in an enterprise database or a fixed field in an application form. Humans operate comfortably on such symbolic data. From the perspective of neural network algebra and propertyless counting, the distinction between the two largely disappears.

We therefore regard semantic and symbolic as interchangeable, differing only in dimension size  through which they are expressed. This motivates the notion of semantic-symbolic pairing. In practical applications, the key diagnostic question is whether the target outcome should be treated as semantic or symbolic. For this reason, we remain skeptical that neuro-symbolic AI approaches can advance very far.

\begin{figure}[H]
    \centering
    \includegraphics[width=0.4\linewidth]{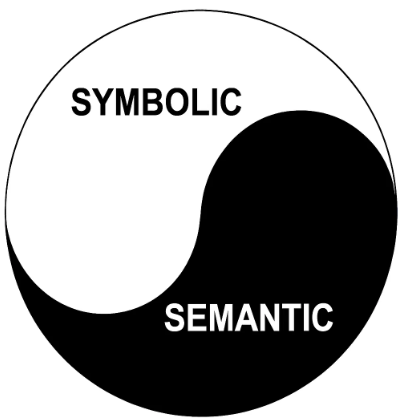}
    \caption{Symbolic Semantic Pairing}
    \label{fig:symbolic_semantic_pairing}
\end{figure}

In the Deep Manifold view, symbolic and semantic data are all propertyless, Section~\ref{sec:propertyless}, and represent two coordinate systems of the same underlying manifold. Symbolic data occupy a discretized coordinate optimized for enumeration and relational consistency, whereas semantic data unfold for continuous embeddings shaped by contextual interaction. The semantic-symbolic \emph{pairing} process therefore serves as a manifold mapping between discrete algebraic indices and continuous representational geometry. This mapping ensures that symbolic precision and semantic generality coexist within a unified numerical framework, enabling a seamless transition between database logic and neural meaning.

\begin{equation}
\begin{aligned}
\text{Symbolic space:} \quad 
    & S = \{ s_i \}_{i=1}^N, \quad s_i \in \mathbb{Z}^n, \\[3pt]
\text{Semantic space:} \quad 
    & M = \{ \boldsymbol{m}_i \}_{i=1}^N, \quad \boldsymbol{m}_i \in \mathbb{R}^m, \\[3pt]
\text{Pairing map:} \quad 
    & \Phi: S \rightarrow M, \qquad 
      \boldsymbol{m}_i = \Phi(s_i), \\[3pt]
\text{Inverse map:} \quad 
    & \Psi: M \rightarrow S, \qquad 
      s_i = \Psi(\boldsymbol{m}_i), \\[3pt]
\text{such that} \quad 
    & \Psi \circ \Phi \approx \mathrm{Id}_S, \quad
      \Phi \circ \Psi \approx \mathrm{Id}_M.
\end{aligned}
\end{equation}

This Figure~\ref{fig:symbolic_semantic_pairing} expresses \emph{semantic–symbolic pairing} as a near-subjective manifold correspondence,
where symbolic discreteness and semantic continuity are dual projections of the same informational structure.

Most enterprise systems—and even daily human environments—operate in a 
fundamentally symbolic way. Databases, forms, APIs, spreadsheets, and 
business logic all exist as symbolic structures. Although foundation models can 
process symbolic data, they are trained primarily on semantic signals, which 
makes symbolic tasks inherently inefficient. In this sense, symbolic tools, 
whether existing or future, act as external boundary conditions: they guide 
iteration toward stable fixed points far more efficiently than relying on 
semantic inference alone.

\subsection{Data Complexity}
\label{sec:data_complexity}

Neural networks face one fundamental challenge: real-world data possess near-infinite scope across several interacting axes, illustrated in Fig.~\ref{fig:data_scale_scope_time}. The first is \emph{vertical scope} (Z-axis), the axis of scale. Similarly, as Fig.~\ref{fig:data_scope_across_scales} ranges from a mountain down to rock microsamples or from public health down to biochemistry, data exist across many resolution levels. A minibatch exposes only one thin slice of this hierarchy, limiting how much curvature the model can absorb per iteration.

\begin{figure}[H]
    \centering
    \includegraphics[width=0.6\linewidth]{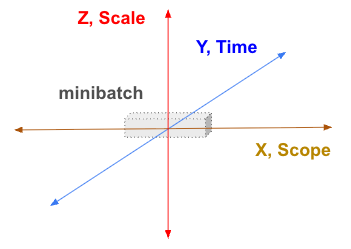}
    \caption{Data Scope}
    \label{fig:data_scale_scope_time}
\end{figure}

The second is \emph{horizontal scope} (X-axis), the breadth of semantic and symbolic variation at a fixed scale. Training never encounters the full distribution at once; instead, it sees fragmented, discontinuous slices scattered across the manifold, precisely the horizontal fragmentation depicted in Fig.~\ref{fig:data_scale_scope_time}. This constrains how the network reconstructs the global structure from local exposures.

The third is \emph{temporal scope} (Y-axis). Real-world data evolve, and the manifold drifts over time. As Fig.~\ref{fig:data_scope_across_scales} suggests through scientific scale progression, systems change over years as well. Without explicit temporal cues, the model forms an implicit internal timeline shaped by the data distribution; as noted in Part~1, Section 3.1 \cite{deepmanifoldpart1}, this can misalign with actual events. For example, ``How many World Cups has Argentina won?'' should yield \emph{three}, but if most training data predate 2022, the model may implicitly anchor to an earlier temporal slice and answer \emph{two}.

\begin{figure}[H]
    \centering
    \includegraphics[width=1.0\linewidth]{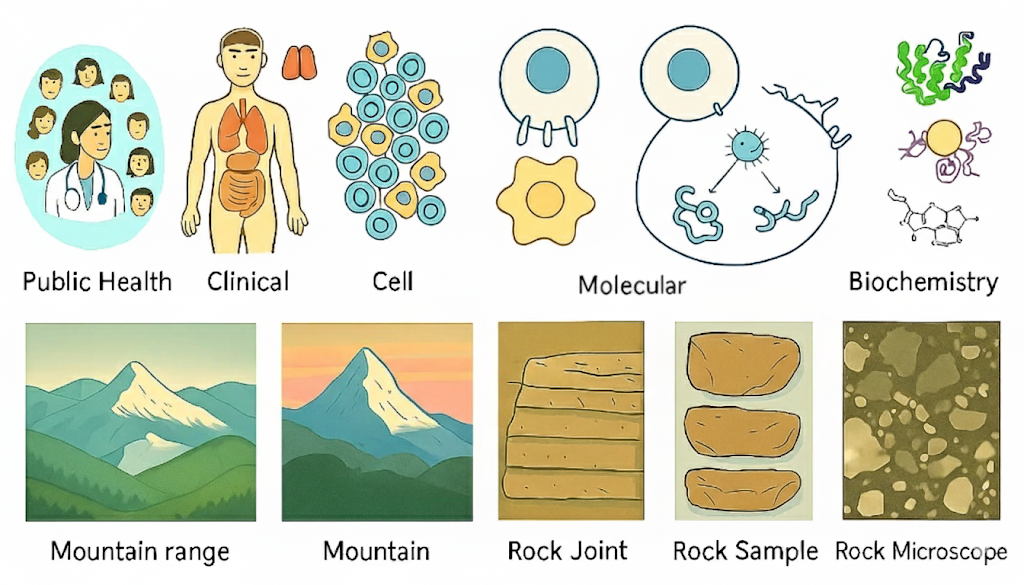}
    \caption{Data Scope Across Scales}
    \label{fig:data_scope_across_scales}
\end{figure}

A fourth axis is \emph{nonlinear complexity} (n-axis), reflecting high-order curvature, piecewise transitions, and discontinuities (Section~\ref{sec:nonlinear_complexity}). Such a structure cannot be represented by a single global approximation and instead requires overlapping, locally adaptive manifolds.

A fifth axis is \emph{modality} (m-axis): modern datasets integrate text, images, audio, and other signals into a unified manifold, enabled by the propertyless nature of neural activations.

\paragraph{(b) Minibatch exposure.} Across all axes $(X, Y, Z, m, n)$, minibatch training observes only a sparse and discontinuous subset of the entire data manifold. Each iteration samples a tiny structural neighborhood---insufficient to reveal global curvature, cross-scale relationships, temporal drift, nonlinear transitions, or multimodal alignment. This sparsity is intrinsic to the learning process and fundamentally limits manifold reconstruction, accelerating rigidity, and diminishing plasticity.

Together, these axes and the minibatch sampling mechanism define \emph{data complexity}: the structural, temporal, nonlinear, multimodal, and sampling-induced difficulty of learning from real-world data.

\paragraph{Mathematical Representation of Data Complexity}
The true structural complexity of the data manifold is as follows.
\begin{equation}
\mathcal{C}_{\mathrm{data}}
=
\int_{\mathcal{M}}
\mathcal{S}_{\mathrm{data}}(x,y,z,m,n)\,
d\rho_{\mathrm{data}}(x,y,z,m,n)
\label{eq:data_complexity_full}
\end{equation}
where
$x$ denotes semantic/symbolic scope,
$y$ denotes temporal evolution,
$z$ denotes scale (resolution hierarchy),
$m$ indexes modality, and
$n$ denotes nonlinear complexity (curvature, discontinuity, and high-order structure).

Minibatch exposure provides only a discrete, sparse approximation:
\begin{equation}
\mathcal{C}_{\mathrm{data}}^{(b)}
=
\sum_{(x,y,z,m,n) \in b}
\mathcal{S}_{\mathrm{data}}(x,y,z,m,n)
\label{eq:data_complexity_batch}
\end{equation}
with the sampling relationship;
\begin{equation}
b \ll \rho_{\mathrm{data}},
\label{eq:data_sampling_relation}
\end{equation}
implying that each minibatch provides only a highly sparse, discontinuous approximation of the full data measure.

Thus, $\mathcal{C}_{\mathrm{data}}^{(b)}$ is the structural complexity observed during training, rather than the full complexity of Equation~\eqref{eq:data_complexity_full}.

\subsection{Data Driven Architecture}
\label{sec:data_driven_architecture}
Because of data high-order nonlinearity and near infinite scope, We simply do not see frontier models handling the lowest granularity of a given domain or discipline. The deeper the nonlinearity, the less effective a universal model becomes at capturing fine-grained detail. 

\paragraph{Model Size}

Even when a frontier model manages lowest-level granularity, it comes at very high cost: requiring massive model size and increasingly complicated prompt text. Such approaches are inefficient and unstable. From the Deep Manifold view, this reflects a mismatch: using a global cover to approximate highly local, nonlinear structures. Data-driven architecture provides a more direct path by aligning model design with the intrinsic granularity of the domain.
We see the future as different size classes of models under neural network mathematics 

\begin{enumerate}
    \item Task models handle narrow and well-defined problems.
    \item Domain models capture broader, but still specialized, scopes.
    \item Discipline models integrate across domains within a field.
    \item Frontier models span multiple disciplines, but with limited granularity and high cost.
\end{enumerate}

\begin{figure}[H]
    \centering
    \includegraphics[width=0.8\linewidth]{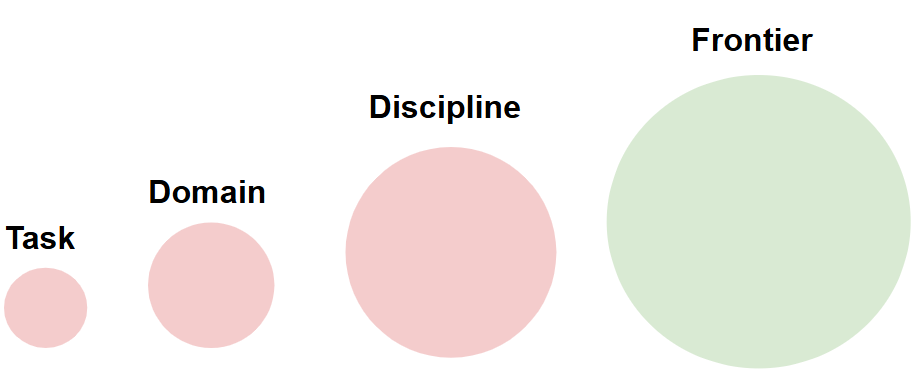}
    \caption{Model Size}
    \label{fig:model_size}
\end{figure}

Further Data-Driven Architecture depends on the learning space, as defined in \textit{Deep Manifold Part 1} \cite{deepmanifoldpart1} , Section 3.4. The size of a model should be primarily driven by the degree of nonlinearity of the data within that learning space.

\subsection{Data Normalization \& Learning Space}
Synthetic and distilled data are not necessarily real-world data; they are normalized data. The synthetic data engine or large model used for distillation serves as a normalizer, transforming complex high-dimensional reality into forms that fit the model’s training manifold. This process not only brings consistency, but also removes diversity and uncertainty, the very properties that define the real world. One major benefit is the reduction of \emph{learning Space}, Section~3.4, \cite{deepmanifoldpart1}.

In recent years, expert data has gained renewed attention. It is seen as high-value because it carries domain reasoning and contextual judgment that ordinary data lack. However, in practice, experts themselves act as normalizers. Across many national strategic and security projects around the world we have done many years as experts in rock engineering, we have observed that even among top experts in the same discipline, the views often diverge, sometimes even conflict directly. Each expert projects reality through their own manifold of experience, theory, and intuition. 

Thus, “expert knowledge” is not the truth of the ground, but a structured normalization of experience. Filter chaos into a stable, communicable form, but also imposes its own bias and loss. The Bitter Lesson reappears here: human or machine, normalization simplifies reality for computation, but at the cost of richness and adaptability. The more we normalize, the more we drift away from the true world manifold—and the less we learn from it.

\begin{quote}
    \textit{Data operations such as synthetic generation, distillation, and “expert data” can be illusionary: they effectively reduce the learning space, but also shift the model away from the real world. It is normalized data.}
\end{quote}

\section{Neural Network Learnability}

\begin{quote}
    \textit{Scaling laws arise from empirical observation; learnability reflects the model’s actual capacity. We also need to understand learning complexity; see Section~\ref{sec:learning_complexity}}
\end{quote}
\subsection{Trainability and Learnability}
Because node coordinates change with every iteration, the network continuously redefines its underlying representation space. This means that the neural network possesses, in principle. an infinite degree of freedom. Such flexibility grants the network unbounded trainability: it can keep adapting indefinitely without exhausting its representational capacity. In this sense, neural networks can be trained “forever”, continually reshaping their internal coordinate system to fit data, refine patterns, or absorb new tasks. 
\vspace{1em}
\begin{figure}[H]
    \centering
    \includegraphics[width=0.6\linewidth]{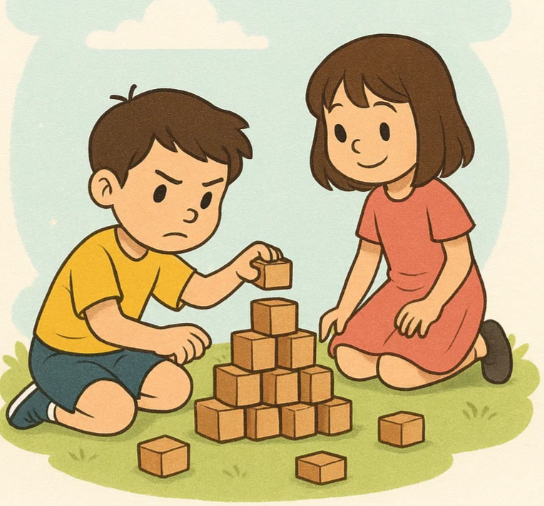}
    \caption{Neural Network Trainability and Learnability}
    \label{fig:nn_trainability_learnability}
\end{figure}

However, \textit{trainability} does not equal learnability. During training, the network develops a form of neural plasticity, but its learnability becomes discrete and bounded by both the network’s architecture and its boundary conditions. Figure~\ref{fig:nn_trainability_learnability} shows: 

\begin{quote}
    \textit{Bricks can be moved around(Trainability), but that does not mean that they can be used to build a structure(Learnability)}
\end{quote}

\subsection{Data Efficient Learning and Neural Plasticity}
\label{sec:data_efficient}
It is  our view that current Transformer designs, whether dense or mixture-of-experts (MoE), exhibit a degree of rigidity in their learnability. Among them, MoE models provide slightly greater flexibility than dense models, but both remain structurally constrained in how they convert \textit{trainability} into genuine learnability.  Given the high-order nonlinearity in data, neural network architectures should be highly elastic, like play-doh, able to absorb and mold themselves around knowledge.

In high-order nonlinear dynamical systems, once rigidity sets in and the system loses adaptability, bifurcations, or singularities are natural outcomes. This aligns with well-established principles in nonlinear dynamics and catastrophe theory.  This also helps explain certain observations of model collapse: as rigidity accumulates within a high-order nonlinear system, the network can encounter bifurcations or singularities that manifest as collapse.

Neural networks train on minibatches, each exposing only a microscopic slice of the data manifold, often far less than one–millionth of the full distribution. Such a narrow exposure contributes little to the sustainability of plasticity: the model absorbs only local curvature, while large portions of the manifold remain unshaped. Instead of expanding elasticity, minibatch-driven updates reinforce the geometry of the few slices repeatedly seen in training, allowing curvature to accumulate unevenly. Over time, this imbalance accelerates rigidity, causing the network to overfit local manifold fragments even as the global manifold remains largely unformed.

\begin{figure}[H]
    \centering
    \includegraphics[width=1\linewidth]{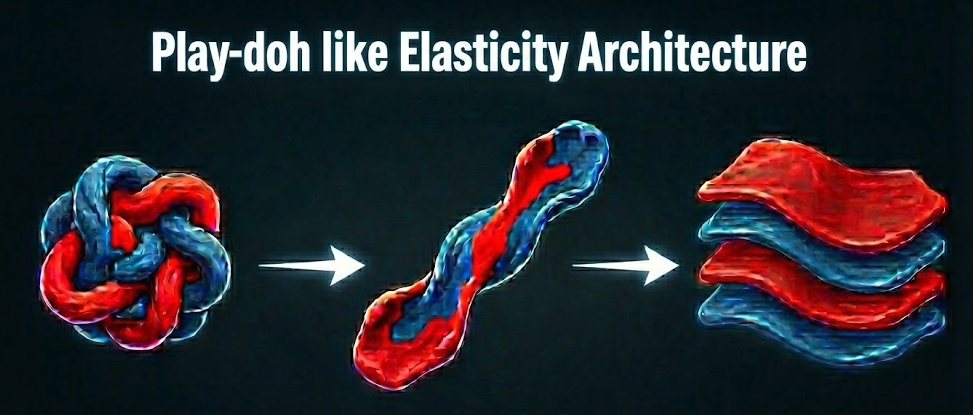}
    \caption{Play-doh like Elasticity Architecture}
    \label{fig:play-doh}
\end{figure} 

In this sense, it is important to maintain a Play-Doh–like flexibility in model behavior, which raises another interesting question: how can we measure model rigidity? One way is through a solid understanding of the learning capacity of the model, since rigidity can be seen as a loss of effective capacity even when the nominal capacity remains.

Learning capacity was discussed at length in our part 1 paper \cite{deepmanifoldpart1}. During training, Deep Manifold spaces develop; yet as the designed capacity is approached, the actual capacity declines, since node covers lose flexibility, even though, by design, they possess infinite degrees of freedom. This phenomenon is commonly known as the neural network bottleneck. The AI community has, often unknowingly, sought to delay or mitigate it through dropout, skip connections, and data engineering for weak learning.

\vspace{1em}
\begin{figure}[H]
    \centering
    \includegraphics[width=1.0\linewidth]{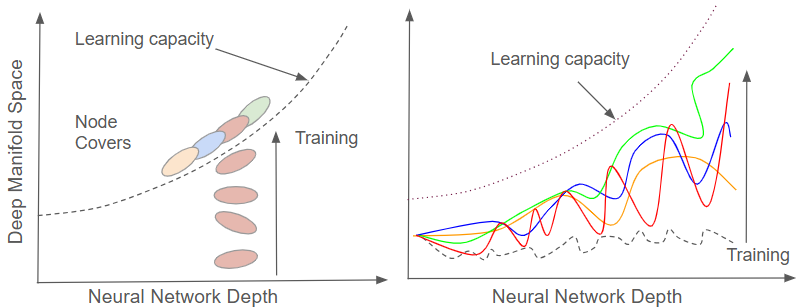}
    \caption{Learning Capacity}
    \label{fig:learning_capacity}
\end{figure}

\begin{quote}
    \textit{For high-order nonlinear data, learning must proceed in a high-order nonlinear manner. Neural-network elasticity is what preserves learning efficiency in this regime.}
\end{quote}

\subsection{Source of Plasticity: Interconnected Toroidal Geometry}
\label{sec:plasticity_source}
Independent of neural architecture, high-order nonlinearity itself gives rise to
neural plasticity. As we describe through the concept of the interconnected
toroidal structure, first introduced in our Part~1 paper \cite{deepmanifoldpart1}, training naturally
produces knotted and intertwined tori within the representational space. These
structures arise from repeated folding, stretching, and alignment of the
manifold under iterative updates.

Drawing inspiration from Gaussian level curves in continuous field models
(Sayama, 2015, \cite{sayama2015}), we model these structures as a Gaussian-symmetric manifold
composed of interconnected tori. Each torus corresponds to a locally smooth
region of the high-order nonlinear space, while their interlocking pattern encodes
the accumulated curvature generated during training.
\vspace{1em}
\begin{figure}[H]
    \centering
    \includegraphics[width=0.75\linewidth]{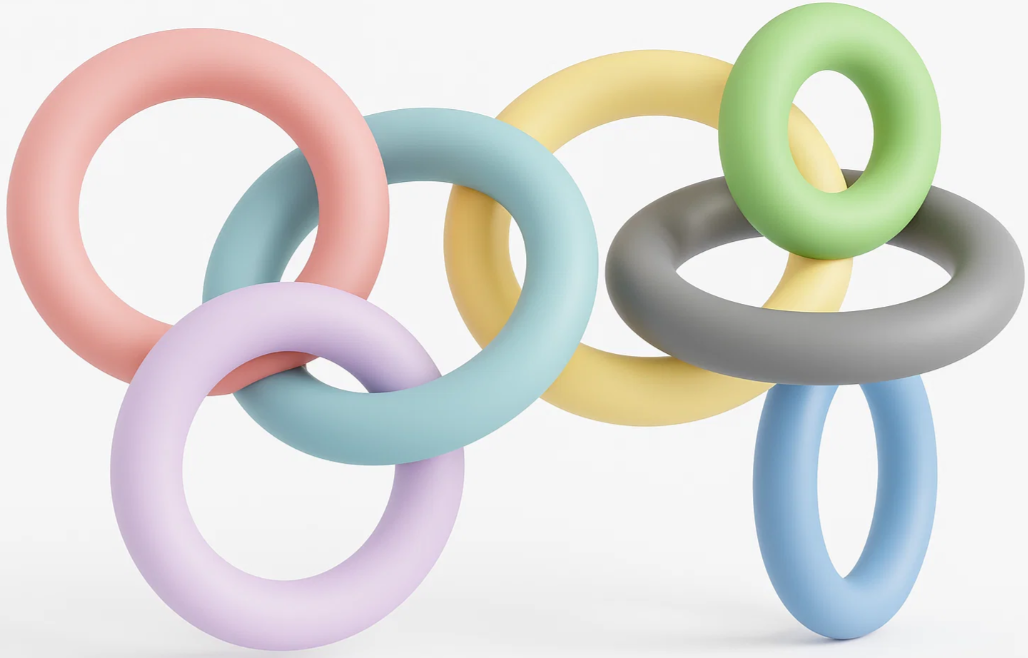}
    \caption{Neural Network Knotted Torus Structure}
    \label{fig:knotted_toru}
\end{figure}

\paragraph{Mathematical Representation on Plasticity.}
Let each local manifold component be described by a Gaussian field
\begin{equation}
\phi_i(x) =
\exp\!\left(
 -\tfrac{1}{2}(x-\mu_i)^{\!\top}\Sigma_i^{-1}(x-\mu_i)
\right),
\qquad i = 1,\dots,N,\; x \in \mathbb{R}^d
\end{equation}

Each interconnected torus is approximated as a set of levels of $\phi_i$:
\begin{equation}
\mathcal{T}_i
\;\approx\;
\{\, x \in \mathbb{R}^d \,\mid\, \phi_i(x)=c_i \,\}
\qquad 0 < c_i < 1
\end{equation}

The knotted toroidal manifold generated during training is then
\begin{equation}
\mathcal{M}_{\mathrm{torus}}(\theta)
=
\bigcup_{i=1}^N \mathcal{T}_i(\theta)
\end{equation}
where the dependence on $\theta$ reflects that each parameter update deforms,
translates, and interlocks the tori.

Training evolves both parameters and geometry:
\begin{equation}
\theta_{t+1}
=
\theta_t - \eta_t \nabla_\theta L(\theta_t)
\qquad
\mathcal{M}_{\mathrm{torus}}^{(t+1)}
=
\mathcal{M}_{\mathrm{torus}}(\theta_{t+1})
\end{equation}

\paragraph{Interconnected Tori as the Source of Plasticity.}
The central point is that these toroidal structures are not geometric byproducts;
they \emph{are the source of neural plasticity}. When the tori are loosely curved,
the manifold remains flexible and easily reconfigurable, enabling new learning.
As training progresses, the curvature accumulates and the tori become more tightly
interconnected. We quantify this accumulated curvature through functional analysis.
\begin{equation}
\mathcal{R}(\theta)
=
\sum_{i=1}^N
\int_{\mathcal{T}_i(\theta)}
\left\| \nabla^2 \phi_i(x) \right\|^2
\, \mathrm{d}\mu_i(x)
\end{equation}
which increases as the manifold becomes more twisted, knotted, and eventually rigid and fragile \cite{deepmanifoldpart1}, Section 1.

This accumulated curvature creates internal geometric resistance: although the
parameters $\theta$ can continue to move freely, the manifold itself resists further
deformation. Effective learning capacity is therefore bounded by
\begin{equation}
\mathcal{C}_{\mathrm{eff}}(\theta)
=
\frac{\mathcal{C}_0}{\,1 + \mathcal{R}(\theta)\,}
\end{equation}
capturing the fact that while infinite trainability exists in principle, the geometry
of high-order nonlinearity imposes a natural ceiling on what the network can
meaningfully learn.

\begin{quote}
    \textit{Data high-order nonlinearity generates numerous interconnected tori, causing the model manifold to stiffen. As each feature progressively converges toward the center of its torus, the manifold loses elasticity and becomes more plastic}
\end{quote}

\subsection{Training Dynamics}
\label{sec:training_dynamics}

Training is the dynamical process of solving the neural network equation introduced in Section~2.3.2. Once the network is expressed as a Lagrangian fixed-point system, its behavior is not static optimization but the continuous relaxation of a residual under changing boundary conditions on a stacked piecewise manifold. Each update reshapes both the representation and the manifold itself, producing a time-dependent trajectory toward equilibrium.

This dynamic mirrors classical relaxation systems. In soil consolidation, an external load acts as a boundary condition, instantly imposed, while the system relaxes only gradually: pore pressure carries the initial response, and deformation emerges later as the boundary condition propagates through the medium. Neural networks behave analogously. During training, the model is repeatedly subjected to \emph{different and evolving boundary conditions}---minibatch selection, loss terms, normalization layers, stochastic perturbations, and even curriculum-like data ordering. Early updates adjust only local slices of the manifold (memorization), while the global geometry remains misaligned with the true fixed point.

Grokking is a direct expression of this delayed relaxation. The model achieves near-perfect training accuracy long before the correct fixed point is formed. Generalization appears abruptly when accumulated curvature, minibatch-driven reorientation, and stochastic perturbations finally allow the manifold to reorganize under the applied boundary conditions and settle into a coherent intrinsic pathway.

Double descent follows the same principle. Near the interpolation threshold, the manifold is in a boundary-driven unstable regime: the curvature is high, the residual correction inefficient, and the training oscillatory. Far beyond this regime, \emph{overparameterization} (Section~3.4) acts as a finer discretization of the stacked manifold, restoring stability and allowing faster relaxation toward fixed points induced by the boundary.

Two additional components shape the trajectory of this relaxation. First, \emph{batch selection} creates a highly discontinuous sequence of boundary conditions: at each iteration, the manifold is pulled toward a different sparse projection of the full data measure, causing alternating curvature and directional drift. Second, the \emph{learning rate schedule} controls the stiffness of the relaxation process itself. Large learning rates inject strong geometric perturbations, increasing plasticity and accelerating early deformation, while decaying learning rates reduce excitation and allow the manifold to settle into a stable fixed-point basin.

Thus, training dynamics capture how the Lagrangian neural equation is solved through iterative relaxation under evolving boundary conditions. Curvature accumulates, dissipates, and re-aligns; plasticity emerges and decays; stability arises only after sufficient manifold reconfiguration. These behaviors set the stage for the structural basis for the next learning.

\paragraph{Mathematical representation on Training Dynamics.}
Let $b_k$ denote the boundary condition at iteration $k$, including minibatch selection, loss terms, normalization, and other stochastic influences. Let the Lagrangian from Section~2.3.2 be
\begin{equation}
L(\theta,\lambda; b_k),
\end{equation}
with fixed-point conditions;
\begin{equation}
\nabla_\theta L(\theta^\ast,\lambda^\ast; b_k) = 0,
\qquad
\nabla_\lambda L(\theta^\ast,\lambda^\ast; b_k) = 0.
\end{equation}
Training dynamics are the iterative relaxation
\begin{equation}
(\theta_{k+1}, \lambda_{k+1})
=
(\theta_k, \lambda_k)
-
\eta_k \,\nabla_{(\theta,\lambda)} L(\theta_k,\lambda_k; b_k)
\end{equation}
where $b_k$ changes with minibatch selection and $\eta_k$ is the learning-rate schedule.
The stacked piecewise manifold is the time-dependent geometric object induced by the trajectory
\begin{equation}
\{(\theta_k,\lambda_k,b_k,\eta_k)\}_{k \ge 0}
\end{equation}
as it relaxes toward (possibly moving) fixed points.

\subsection{Learning Triangle}
\label{sec:learning_triangle}

Learning emerges only from the interaction of three structural forces:
\emph{data complexity}, \emph{training dynamics}, and \emph{architecture efficiency}.
Each side constrains the model from a different direction, and the learnability appears
only when these constraints overlap. The Learning Triangle therefore provides
the local structure that governs why a model learns, what it learns, and how large
the model must be to learn it.

\begin{figure}[H]
    \centering
    \includegraphics[width=0.60\linewidth]{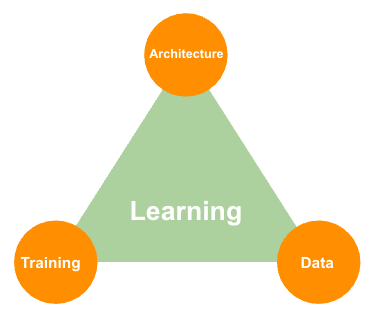}
    \caption{Learning Triangle}
    \label{fig:learning_triangle}
\end{figure}

\paragraph{Data.}
Data define the geometry of the learning space.
As discussed in Section~\ref{sec:data_complexity}, real-world data possess
high-order nonlinearity and wide semantic--symbolic scope. These axes determine
the manifold curvature and the intrinsic learning space. Cleaner or narrower data
shrink the manifold; heterogeneous or nonlinear data expand it, increasing
the representational burden.

\paragraph{Training.}
Training supplies evolve boundary conditions.
Section~\ref{sec:training_dynamics} shows that minibatch fragmentation exposes
only sparse, discontinuous slices of the full data measure, while learning-rate
schedules, perturbations, and normalization determine how these fragments are
integrated into an iterative pathway. Training therefore not only moves across
the manifold, it \emph{orients and constructs} the manifold through iteration.

\paragraph{Architecture.}
Architecture governs the efficiency and elasticity of representation.
Section~\ref{sec:data_efficient} emphasizes that architectural choices determine
curvature capacity, plasticity retention, and the effective degrees of freedom
available for learning. Even with favorable data and training, a rigid
architecture limits learnability; elastic architectures preserve curvature under
high-order nonlinearity.

Together, these three forces form the \emph{Learning Triangle}, the local
structure from which model quality and model size emerge. No single side
determines learnability; it is the \emph{triangular equilibrium} that enables a
model to approximate the manifold implied by data and training.

\paragraph{Mathematical Representation on Learning Triangle}

\paragraph{Data as manifold geometry.}
Following the formulation in Section~\ref{sec:data_complexity}, the data define a
structural manifold integral:
\begin{equation}
M_{\text{data}}
=
\int_M
S_{\text{data}}(x,y,z,m,n)\, d\rho_{\text{data}}
\end{equation}

\paragraph{Training as evolving boundary conditions.}
Each iteration applies its own boundary operator, reflecting minibatch
selection and schedule:
\begin{equation}
\partial\Omega_{\text{train}}(k) = b_k 
\end{equation}

\paragraph{Architecture as mapping geometry.}
Architecture defines how curvature is transported and represented:
\begin{equation}
h_{k+1} = \Phi_{\text{arch}}(h_k) 
\end{equation}

\paragraph{Composite Learning Operator}

Learnability is realized only through the alignment of these three components.
The fundamental operator induced by the Learning Triangle is the composite map:
\begin{equation}
h_{k+1}
=
\Phi_{\text{arch}}
\circ
\partial\Omega_{\text{train}}
\circ
M_{\text{data}}
\, (h_k)
\end{equation}
linking manifold geometry, boundary conditions, and mapping efficiency into a
single iterative process. The interior of the Learning Triangle:the region
where curvature, boundary, and elasticity coincide, is precisely where a model
becomes learnable.

\subsection{Transferability, Structural Redundancy And Feature Abundance}
\label{sec:transferability_redundancy}
A first and central idea of \emph{the Deep Manifold} is that a neural network constructs a 
\emph{stacked, piecewise manifold}. Each layer forms multiple locally smooth 
regions, and across depth these regions accumulate into a hierarchical manifold 
cover. This geometry is not a secondary feature---it is the organizing principle 
behind many observed behaviors in modern neural networks. Among them, 
\emph{redundancy} and \emph{transferability} emerge as two tightly connected 
consequences of this structure.

Formally, each layer decomposes its representation space into manifold pieces, Equation~\ref{eq:piecewise-manifold}:
\[
M_k = \bigcup_{i=1}^{N_k} M_{k,i}
\]
with each $M_{k,i}$ a smooth local patch. Because training proceeds through 
mini-batches and never sees the full data distribution at once, these patches 
inevitably overlap:
\begin{equation}
M_{k,i} \cap M_{k,j} \neq \varnothing 
\;\;\Rightarrow\;\;
f_\theta(M_{k,i}) \approx f_\theta(M_{k,j})
\end{equation}
This overlap is the geometric origin of \emph{structural redundancy and feature abundance}. The network 
develops multiple approximators for the same region simply because different 
batches induce different partial trajectories across the manifold. Abundance is 
therefore not an artifact of overparameterization; it is a necessary consequence 
of high-order nonlinearity and limited data scope.

Once redundancy is understood as overlapping covers, the feasibility of 
\emph{pruning} and \emph{merging} becomes clear. Pruning removes parameters 
associated with a local cover, yet if neighboring covers approximate the same 
region, the manifold remains intact:
\begin{equation}
M_k \setminus M_{k,i}
\;\text{still covers}\;
\operatorname{supp}(p(x))
\quad\text{when}\quad
\exists j:\; M_{k,i} \subseteq M_{k,j}
\end{equation}
Merging is the geometric analog: collapsing two redundant slices into a single cover,
\begin{equation}
M_{k,i}^{\text{new}} = M_{k,i} \cup M_{k,j},
\qquad
f_\theta(M_{k,i}^{\text{new}})
\approx 
f_\theta(M_{k,i})
\approx 
f_\theta(M_{k,j})
\end{equation}
Both operations work only because the manifold is stacked and piecewise. Without 
this structure, pruning would destroy essential geometry, and merging would corrupt 
curvature. This explains why pruning and merging succeed for task-oriented models, 
but must be approached cautiously for frontier models, whose manifolds are deeply 
stacked and highly nonlinear.

The same geometry explains \emph{transferability}. The learned representation 
evolves through iterative transformations,
\begin{equation}
h_{k+1} = f_k(h_k), 
\qquad
M_{k+1,j} = \Phi_{k,i\to j}(M_{k,i})
\end{equation}
The early layers tend to encode the global curvature of the manifold. Because this 
curvature is captured redundantly across overlapping patches, the base model 
preserves a flexible geometric scaffold that can be re-aligned to new tasks, 
domains, or modalities:
\begin{equation}
\Phi_{k,i\to j}^{\text{new}} 
\approx
\Phi_{k,i\to j}^{\text{old}}
\quad\text{for early layers}.
\end{equation}
Later layers become closer to linear classifiers and are more rigid:
\begin{equation}
\frac{\partial \Phi_k}{\partial \theta}
\;\text{is large for small } k,
\qquad
\frac{\partial \Phi_k}{\partial \theta}
\;\text{is small for large } k.
\end{equation}
RoLA leverages exactly this property by adapting deeper layers within the 
high-order nonlinear region where curvature is most expressive.

Redundancy and transferability are also manifest in the structure of fixed points. 
Deep networks possess multiple stable attractors, and their basins often overlap:
\begin{equation}
\mathcal{A}(x_i^{\*}) \cap \mathcal{A}(x_j^{\*}) \neq \varnothing
\end{equation}
These shared attraction regions allow models to prune parameters without losing 
stability and to adapt representations without requiring full retraining. They 
also form the basis for merging different models, as their fixed points lie within 
compatible manifold neighborhoods.

In this light, redundancy and transferability are not separate ideas, but two 
expressions of the same geometric mechanism. The stacked piecewise manifold 
provides overlapping local covers (redundancy) and flexible global curvature 
(transferability). This duality explains why the pruning and merging function is effective, why 
transfer learning is effective, and why learnability is preserved across domains. 
The model's structure itself, not any specific architecture or heuristic, 
gives rise to these capabilities.

\begin{quote}
    \textit{This is the foundational insight of the deep manifold: the geometry of the connected, stacked, piecewise manifold governs the model’s structural redundancy, feature abundance, and learnability, forming the basis for transfer learning, model pruning, and model merging.}
\end{quote}

\subsection{Causal Inference}
Large Language Models already exhibit strong causal-inference behavior. Their training corpora contain abundant causal patterns, scientific studies, case analyses, and real-world narratives, which allow the models to internalize statistical and linguistic cues of cause–effect structure. At the same time, the architecture aligns contextual dependencies across massive text distributions, reconstructing directional relations without explicit symbolic modeling. Together, data richness and neural learnability yield causal reasoning as an emergent form of manifold alignment rather than a predefined causal graph.

Causal inference is often framed symbolically: graphs, logical dependencies, interventions, but within the Deep Manifold view its structure becomes geometric and numerical. As with neural operators, causal relations do not arise from predefined analytic forms; they emerge from \emph{learned manifold geometry}, \emph{boundary-induced orientation}, and \emph{piecewise continuity} under high-order nonlinearity.

\paragraph{Local Structure: Causality as Ordered Manifold Geometry.}
Classical causal models impose fixed graphs, whereas neural networks learn an \emph{ordered topology}. Each causal relation defines a partial order that induces an Alexandrov-type topology: neighborhoods encode causal accessibility, closures encode causal past/future, and separability reflects statistical identifiability. This mirrors the contrast between the local basis of Taylor and the fixed global basis of Fourier: Classical causal graphs are rigid, but neural networks form \emph{local causal neighborhoods} in stacked pieces.

\paragraph{Cross-Piece Structure: Algebra of Dependency and Transition.}
Piecewise manifolds unify metric geometry in lattice order. Each slice provides a smooth metric region; the lattice order encodes dependency or precedence; and the cross-slice transitions represent interventions or context shifts. Where Fourier-based operators fail at discontinuities, classical causal models fail in fragmented structures. Stacked piecewise manifolds naturally support fragmented causal geometry and its recombination.

\paragraph{Global Behavior: Causal Reasoning as Geometric Stability.}
Causation emerges when a mapping across manifold slices is \emph{invariant under deformation:robust}  to perturbation, noise, and coordinate shifts. Correlations appear as local alignments, but true causal mechanisms emerge only when manifold flow preserves continuity across slices. Interventions act as boundary-condition shifts that redirect the flow, just as in a fixed-point iteration. The network infers causality by stabilizing mappings that remain coherent under repeated iteration.

\paragraph{Unified View.}
Causal inference becomes a numerical--geometric process: distance encodes influence strength, order encodes causal direction, topology encodes stability, and piecewise manifold composition encodes how causal structure propagates across discontinuity and scale. In this view, causality is not symbolic; it is a \emph{topological invariant} revealed through the same iteration--integration-correction cycle that governs deep manifold dynamics.

\begin{quote}
\textit{Causality is an ordered manifold geometry. Neural networks can discover it not by deduction, but by approximation mapping  that survive perturbation, distortion, and coordinate drift.  What persists across deformation becomes causal; what collapses under iteration is merely correlation.   In this sense, causal inference emerges as a topological invariant of the deep manifold, but a structure revealed by learning in an ordered manifold geometry}
\end{quote}

\subsection{Neural Network Mirage \& Learability}
There are many research efforts suggesting, or at least assuming, that neural networks can be explained through established mathematical and statistical frameworks. Some view them as Monte Carlo methods or Gaussian fields, others as Bayesian inference machines. Still others interpret them through the lens of Lie groups or even within the abstractions of category theory. Physics-based approaches also have their merit: they anchor networks in a physical carrier, governed by equations with high-order nonlinearity. More speculative perspectives go further, attributing to neural networks cognitive abilities such as reasoning and memory.

We regard these as Neural Network Mirages: illusions of built-in intelligence. The underlying mathematics of neural networks is, in fact, more primitive than any of these frameworks. Their geometry and algebra show that they are fundamentally machinic rather than theoretical. Yet, it is precisely this primitiveness that gives them unique power: the \textit{learnability} to absorb, approximate, and embed higher-level structures, statistical, physical, or cognitive, through exposure to vast amounts of text and imagery. 

\begin{figure}[H]
    \centering
    \includegraphics[width=0.8\linewidth]{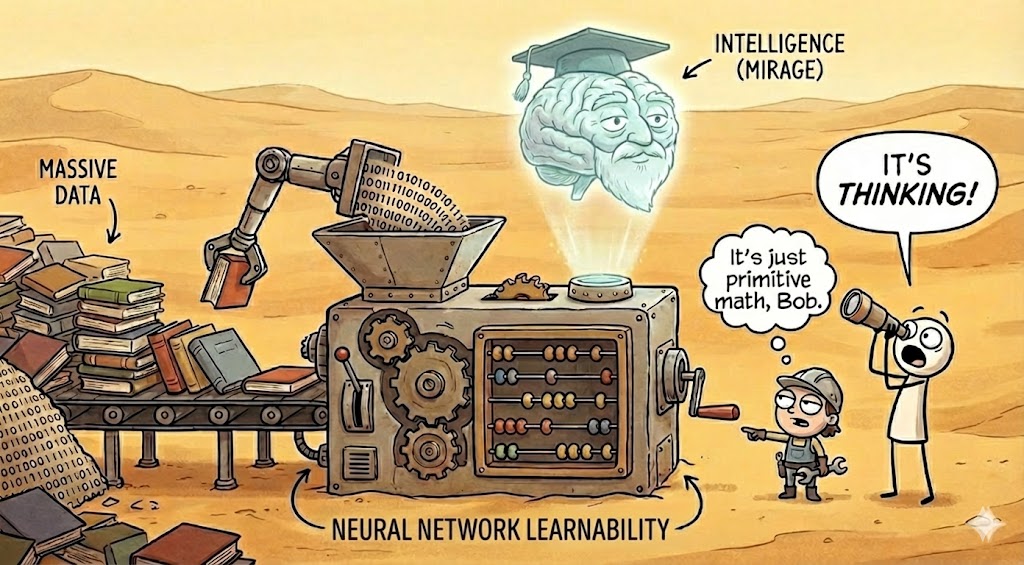}
    \caption{Neural Network  Mirage \& Learability}
    \label{fig:neural_network_mirage}
\end{figure}

Neural networks themselves do not possess intrinsic “magic.” What appears as magic capabilities: reasoning, memory, even creativity are not innate properties of the network, but learned approximations drawn from data. The model exhibits these functions because the training has embedded them, not because the architecture intrinsically contains them. For this reason, the focus should not be on whether a model “has” reasoning, but on its learnability of reasoning: its ability to approximate reasoning patterns when exposed to sufficient and structured data. 

For this reason, we doubt any suggestion that neural networks are intrinsically similar to the biological brain. Our understanding of the human brain is fragmentary, perhaps less than fivepercent, and projecting such complexity onto the models of today risks conflating what emerges from scaled iteration with what has evolved in biology. The mathematics of neural networks aligns instead with Rich Sutton’s Bitter Lesson: progress comes not from embedding human theories in machines, but from leveraging simple mechanisms at scale.

\begin{quote}
\textit{Neural networks do not contain intelligence or cognitive ability; they learn from data via stochastic approximation through learnable numerical computation.
What appears as reasoning or memory is a mirage formed by manifold alignment under massive data, not an intrinsic cognitive structure.
Capabilities emerge only when training pressure carves stable patterns into a primitive geometric system.
In this light, learnability, not intelligence, is the true essence of the neural network.}
\end{quote}

\section{AI for Science and Engineering}
\subsection{Forward and Inverse Combined}
Solving forward problems (propagation in positive time) and inverse problems (reconstruction in negative time) has long been a central challenge for mathematicians and engineers. Traditionally, forward problems are well-posed and tractable, while inverse problems are often poorly posed, unstable, and mathematically elusive. Neural networks, however, approach both naturally within the same framework. Neural networks solve forward and inverse problems within the same boundary–conditioned iteration on the stacked piecewise manifold \cite{deepmanifoldpart1}.

\paragraph{Forward problem.}
Given input $x$ \emph{and manifold slices}, Section~\ref{sec:piecewise_manifold_formalization}, Equation~\ref{eq:piecewise-manifold}
\[
M_k = \bigcup_i M_{k,i}
\]
the forward problem computes the output;
\begin{equation}
y = F_{\theta}(x),
\end{equation}
by propagating the activation through these stacked piecewise regions (Equation~\ref{eq:initial-embedding} and \ref{eq:representation-update}):
\[
h_{k+1} = f_k(h_k), 
\qquad 
h_0 = E(x)
\]

In the iterated--integral formulation of Section~\ref{sec:iterated_integral}, each layer integrates over a 
prompt--selected manifold slice (Equation~\ref{eq:forward_integral}):
\[
h_{k+1}(x)
=
\int_{\partial\Omega_k(x)}
f_k\!\big(h_k(x), \xi\big)\,
\mathrm{d}\mu_k(x)(\xi)
\]

\paragraph{Inverse problem.}
Given an observation $y$, the inverse problem seeks an input $x$ that satisfies
\begin{equation}
F_{\theta}(x) = y
\end{equation}
This corresponds to moving through the same stacked manifold in the reverse 
direction:
\begin{equation}
h_k = f_k^{-1}(h_{k+1}),
\end{equation}
or, in reverse--integral form,
\begin{equation}
h_k(y)
=
\int_{\partial\Omega^{-1}_k(y)}
g_k\!\big(h_{k+1}(y), \xi\big)\,
\mathrm{d}\nu_k(y)(\xi),
\end{equation}
where $g_k$ and $\mathrm{d}\nu_k$ denote the learned inverse kernels and inverse 
boundary measures.

\paragraph{Unified view.}
Forward and inverse problems are not distinct mechanisms but two orientations of a 
single boundary--conditioned manifold operator:
\begin{equation}
h_{k+\varepsilon}
=
\int_{\partial\Omega^{(\varepsilon)}_k}
\mathcal{K}^{(\varepsilon)}_k(h_k,\xi)\,
\mathrm{d}\mu^{(\varepsilon)}_k(\xi)
\qquad 
\varepsilon \in \{+1,-1\}
\end{equation}

\paragraph{Inverse as fixed point.}
The inverse solution is the fixed point of the reverse operator:
\begin{equation}
x^{\ast} = \Psi_{\theta}(x^{\ast}),
\qquad 
\Psi_{\theta} = F_{\theta}^{-1}
\end{equation}

Thus, inverse reconstruction is simply:
\begin{equation}
x^{\ast}
=
\Phi_{\theta}^{(-)}
\circ \cdots \circ 
\Phi_{\theta}^{(-)}(y)
\end{equation}

In this formulation, forward simulation and inverse reconstruction 
are not separate tasks but two orientations of the same 
stacked--manifold iteration, driven entirely by boundary conditions.

\begin{quote}
    \textit{By learning from data, they acquire a unique capacity to approximate mappings in both directions, forward simulation and inverse reconstruction,without requiring explicit analytical formulation. This makes them particularly powerful tools for tackling inverse problems that classical methods struggle to handle.}
\end{quote}
\subsection{Propertyless Neural Network}
The \emph{propertyless} nature of neural networks, see Section \ref{sec:propertyless}, offers a unique capacity to incorporate additional observations in the real-world directly into the solution process. In classical science and engineering, nearly all fundamental formulations—including the seventeen famous physical equations, exist as abstractions of reality, where geometry enters only through boundary conditions. However, neural networks are not bound by fixed formulations. 

They can flexibly integrate extra geometric or observational constraints as boundary conditions at any stage of training. This ability to embed richer constraints is one of the main reasons neural networks can achieve orders-of-magnitude speedup (often cited as 100×) compared to classical numerical computation: each additional boundary condition accelerates convergence by guiding the network more efficiently toward the fixed point solution.

Beyond accelerating convergence like additional boundary conditions, the same \emph{propertyless} character also enables multi-disciplinary investigation and computation.They can seamlessly unify observations across physics, chemistry, biology, and engineering into a single computational framework. This opens a path toward solving problems that span multiple disciplines, which classical methods cannot easily accommodate.
    
\begin{quote}
    \textit{The propertyless nature of neural networks remains widely unrecognized and significantly undervalued.}
\end{quote}

\subsection{Discoverability: Interpolation and Extrapolation}

Most scientific discoveries and engineering advances arise not from a single domain
, but from \emph{multiple disciplines working together}, including the extraction, recombining,
and cross-pollination of ideas. Classical sciences progress when mathematics informs
physics, physics inspires engineering, biology influences computation, and so on.
The discovery emerges when the fields intersect.

In low-dimensional or narrow problem spaces, the distinction between interpolation
and extrapolation is meaningful: interpolation operates within a bounded region,
while extrapolation steps outside it. But in the \emph{near-infinite dimensions of
the learning space}, shaped by data drawn from countless disciplines simultaneously,
this boundary dissolves. Neural networks operate in a \emph{connected, stacked, piecewise manifold landscape}, where the representational span grows rapidly with dimension,
curvature, iteration, and, critically, with cross-disciplinary diversity.

In such a space, interpolation and extrapolation collapse into the same operation:
\emph{manifold traversal}.

Formally, letting
\begin{equation}
D = \{(x_i, y_i)\}_{i=1}^N \subset X \times Y, \quad f_\theta : M \to Y, \quad
M = \mathrm{span}\{x_i\}
\end{equation}

the classical distinction is:
\begin{equation}
\begin{aligned}
\text{Interpolation: } & \hat{y}(x) = f_\theta(x), \quad x \in \mathrm{conv}\{x_i\}, \\
\text{Extrapolation: } & \hat{y}(x) = f_\theta(x), \quad x \notin \mathrm{conv}\{x_i\}, \\
\text{Unified: } & \hat{y}(x) = f_\theta(x), \quad x \in M = \mathrm{span}\{x_i\}.
\end{aligned}
\end{equation}

As the network learns within the \emph{near-infinite dimensions of learning space},

\begin{equation}
\text{As } \dim(M)\to\infty,\ \ 
\operatorname{conv}\{x_i\} \subset \operatorname{span}\{x_i\}
\end{equation}
and the effective distinction between the two weakens for generalization.\\

the boundary separating interpolation from extrapolation weakens. Discovery becomes
movement along learned manifold directions rather than switching regimes.

Two principles must be explicitly stated:

\begin{enumerate}
    \item \emph{Near-infinite data scope (cross-domain).} Real-world datasets span
    science, engineering, medicine, humanities, economics, mathematics, and culture, each
    providing different geometric slices of the representational world. Every minibatch
    exposes only a fragment of this immense space. Over training, these fragments accumulate
    into a unified, expanding \emph{stacked, piecewise manifold landscape}, making
    interpolation and extrapolation indistinguishable as the manifold grows across domains.
    
    \item \emph{Propertyless neural learnability across disciplines.} Neural networks
    possess no intrinsic attributes: activations, coordinates, and internal representations
    carry no built-in semantics or modality restrictions. This \emph{propertyless} nature
    allows data from any discipline to reshape the coordinate system, basis functions,
    and the \emph{stacked, piecewise manifold landscape}. As a result, the model learns
    \emph{cross-disciplinary generalization directions} that no single field contains on
    its own. Discovery becomes an emergent property of manifold expansion driven by
    multi-disciplinary learning and stochastic approximation.
\end{enumerate}

Thus, interpolation/extrapolation is no longer a dichotomy. In the \emph{near-infinite
dimensions of learning space} (Section~\ref{sec:data}), where multi-disciplinary data and propertyless (Section~\ref{sec:propertyless}) learnability shape the representational geometry, neural generalization is best
understood as \emph{manifold traversal}, governed by curvature, boundary conditions,
and residual correction.

\begin{quote}
\textit{In the near-infinite dimensions of learning space, the boundary between interpolation and extrapolation collapses into a single intrinsic trajectory across stacked piecewise manifolds. Discovery arises not by stepping outside the data, but by following directions induced by neural networks applying boundary conditions distinct from human ‘knowledge,’ shaped solely through their iterative, propertyless manifold-learning process across disciplines.}
\end{quote}

\subsection{Neural Operator}

Taylor expansions converge locally around a point, whereas Fourier expansions
converge globally over a rectangular domain with orthogonal bases. Fourier methods often perform
exceptionally well It is tempting to apply higher-order functions such as Taylor series of high-order \cite{ma1996discontinuous}, Fourier series \cite{ma1995single}, or even the Kolmogorov–Arnold Network (KAN), due to their demonstrated effectiveness in traditional numerical calculations. In these contexts, higher-order expansions improve approximation power, stability, and convergence.

However, neural networks operate in a fundamentally different regime. Because
node coordinates shift throughout training, the representational space
continuously reorients itself. Once an appropriate architecture is chosen, these
representations become almost orthogonal by default and exhibit near-global
convergence over stacked piecewise manifolds. 

Neural networks extend the Galerkin principles in a fundamentally different way.
They make \textbf{all} Galerkin components learnable instead of fixed:

\begin{enumerate}
    \item \emph{Learnable basis}: generated by activations and continually
    reshaped through weight updates;
    \item \emph{Learnable operator}: each layer's transformation evolves during
    training, modifying the manifold itself;
    \item \emph{Learnable test-function space}: the network implicitly learns
    how to probe and reduce residuals;
    \item \emph{Geometry of the learning domain}: represented as stacked piecewise
    manifolds that deform and adapt under data-driven training.
\end{enumerate}

This yields a natural three-tier hierarchy:
\begin{enumerate}
    \item \emph{Classical operators}: fully fixed Galerkin systems (Taylor,
    Fourier, polynomial bases), stable but rigid and domain-limited.
    \item \emph{Neural operators (e.g., FNO)}: partially learnable Galerkin
    systems---trainable operators but fixed bases; effective only when the
    domain matches the basis.
    \item \emph{Deep neural networks}: fully learnable Galerkin systems with basis, operator, test-function space, and geometry all
    learned, naturally adapting to high-order nonlinearity, discontinuity, and
    data-defined curvature.
\end{enumerate}

\begin{quote}
\textit{Neural networks gain their advantage not from increasingly high-order operators, but from general, natural, and inherently “weak” approximation bases. This reinforces the Bitter Lesson: progress in AI does not come from hand-crafted functional sophistication.}
\end{quote}

\subsection{Real World Stochastic Fixed Point}
Neural networks approximate a type of fixed point that is closer to the real world: a stochastic fixed point. In contrast, classical numerical computation, whether grounded in theoretical formulation or empirical modeling, typically yields an average fixed point. This distinction is significant. 

\vspace{1em}
\begin{figure}[H]
    \centering
    \includegraphics[width=1.0\linewidth]{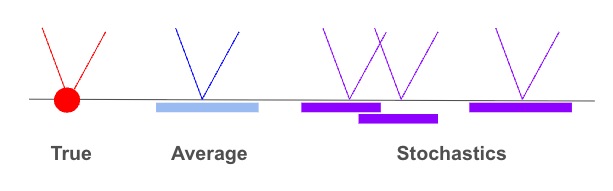}
    \caption{True, Average and Stochastic Fixed Point}
    \label{fig:stochastic_fixed_point}
\end{figure}
The real world operates under stochasticity—driven by uncertainty, fluctuations, and probabilistic dynamics—so a stochastic fixed point is often more faithful to reality than its average counterpart. In this sense, neural networks represent a tremendous advance over traditional numerical computation: they allow solutions to emerge not as static averages, but as dynamic equilibria shaped by randomness itself.

However, we should pay particular attention to how these stochastic fixed points are grounded. Without anchoring them to established theoretical or empirical fixed points, the solutions risk drifting into artifacts of data or training rather than reflecting underlying physical or scientific truth. The power of stochastic fixed points lies in their realism, but their reliability depends on careful grounding.

Mathematically, the three notions in Figure~\ref{fig:stochastic_fixed_point} can be written as

\begin{align}
\text{(1) True fixed point:} \qquad 
& h^{\mathrm{true}} = \Phi\!\big(h^{\mathrm{true}}\big),
\label{eq:true_fixed_point} \\[6pt]
\text{(2) Average fixed point:} \qquad
& h^{\mathrm{avg}} = \Phi\!\big(h^{\mathrm{avg}}\big) + \bar{\delta},
\qquad 
\bar{\delta} = \mathbb{E}[\delta_k],
\label{eq:avg_fixed_point} \\[6pt]
\text{(3) Stochastic fixed point:} \qquad
& H^\ast = \Phi(H^\ast) + \varepsilon,
\qquad 
\varepsilon \sim \mathcal{D}_\varepsilon.
\label{eq:stochastic_fixed_point}
\end{align}

These three forms distinguish the ideal theoretical fixed point, the 
average fixed point produced by numerical computation under systematic 
perturbations, and the stochastic fixed point satisfied only in 
distribution in real-world and neural systems.

\begin{quote}
    \textit{From Statistical design in civil engineering (Uniform Building Code) to Stochastic Differential Equations (SDEs) for dynamic systems with great deal of randomness and uncertainty. Neural-network stochastic fixed points (Section~\ref{sec:nn_stochastic}) offer an elegant, natural, and uniquely powerful approach to these problems, it is second to none.}
\end{quote}

\subsection{PINN and Numerical Computation}
\label{sec:pinn_nc}
Physics-Informed Neural Networks (PINNs) are commonly viewed as a bridge
between deep learning and numerical computation. From the Deep Manifold
perspective, however, PINNs are more precisely characterized as
Galerkin-like residual solvers: they impose a \emph{fixed} operator and
\emph{fixed} domain geometry, while the network provides a highly flexible but
unstructured learnable basis. Their mathematical behavior becomes
transparent when examined through boundary-conditioned fixed-point theory.

To formalize the discussion, we introduce a unified notation:
\begin{enumerate}
    \item $\Omega \subset \mathbb{R}^d$: domain; $\partial\Omega$: boundary.
    \item $\mathcal{M}_\theta$: stacked, piecewise manifold induced by the network.
    \item $u_\theta(x)$: PINN output; $\; p_\theta(y \mid x)$: LLM output.
    \item $\mathcal{N}$: PDE operator; $\mathcal{B}$: boundary operator.
    \item $\mathcal{R}_\theta$: reward/preference operator.
    \item $\mathcal{R}_{\text{pde}},\mathcal{R}_{\text{bc}},\mathcal{R}_{\text{pref}}$:
    respective residuals.
\end{enumerate}

\paragraph{PINN as a Residual Solver on Fixed Geometry}

A PINN seeks $u_\theta$ satisfying the PDE:
\begin{equation}
    \mathcal{N}[u_\theta](x) = f(x), \qquad x \in \Omega
    \label{eq:pinn-pde}
\end{equation}
and the boundary condition:
\begin{equation}
    \mathcal{B}[u_\theta](x) = g(x), \qquad x \in \partial\Omega
    \label{eq:pinn-bc}
\end{equation}

Define the PDE and boundary residuals:
\begin{align}
    \mathcal{R}_{\text{pde}}(x;\theta)
    &= \mathcal{N}[u_\theta](x) - f(x),
        \label{eq:pinn-res-pde} \\
    \mathcal{R}_{\text{bc}}(x;\theta)
    &= \mathcal{B}[u_\theta](x) - g(x)
        \label{eq:pinn-res-bc}
\end{align}

The PINN energy functional becomes:
\begin{equation}
    \mathcal{E}_{\text{PINN}}(\theta)
    =
    \lambda_{\text{pde}}
    \, \mathbb{E}_{x\sim\Omega}
    \!\left[\|\mathcal{R}_{\text{pde}}(x;\theta)\|^2\right]
    +
    \lambda_{\text{bc}}
    \, \mathbb{E}_{x\sim\partial\Omega}
    \!\left[\|\mathcal{R}_{\text{bc}}(x;\theta)\|^2\right]
    \label{eq:pinn-energy}
\end{equation}

In the Deep Manifold view, PINNs impose \emph{strong, continuous} boundary
conditions on a non-learnable operator $\mathcal{N}$ and fixed geometry
$(\Omega,\partial\Omega)$, while the network supplies a learnable basis on the
manifold $\mathcal{M}_\theta$. The misalignment between learned geometry and PDE
geometry explains common pathologies: drift in the boundary-layer, stiffness,
gradient imbalance, slow convergence, and failure of discontinuities.

\paragraph{Boundary-Condition Equivalence Between PINN and GRPO}

GRPO enforces pairwise preference boundaries rather than analytic PDE
boundaries. The learned reward model is as follows:
\begin{equation}
    r_\theta(y)
    = \beta \log p_\theta(y \mid x)
    \label{eq:grpo-reward}
\end{equation}
producing a preference residual:
\begin{equation}
    \mathcal{R}_{\text{pref}}(y_w,y_l;\theta)
    =
    r_\theta(y_w) - r_\theta(y_l)
    \label{eq:grpo-pref-res}
\end{equation}

The GRPO objective is:
\begin{equation}
    \mathcal{E}_{\text{GRPO}}(\theta)
    =
    -
    \mathbb{E}_{(y_w,y_l)\sim\mathcal{D}}
    \left[
        \log \sigma\!\left(
            \mathcal{R}_{\text{pref}}(y_w,y_l;\theta)
        \right)
    \right]
    \label{eq:grpo-energy}
\end{equation}

\paragraph{Unified interpretation.}

\begin{enumerate}
    \item PINN imposes \emph{strong, continuous, and analytic} boundary conditions.
    \item GRPO imposes \emph{weak, discrete, behavioral} boundary conditions.
    \item Both modify neural convergence through \emph{boundary-conditioned fixed-point iteration}.
\end{enumerate}

PINN solves physics via strong continuous boundaries; GRPO aligns behavior
via weak discrete boundaries. Both reshape neural convergence through boundary-conditioned iteration on stacked piecewise manifolds. The success of GRPO is rooted in the use of weak and discrete boundary conditions, as described in Section~\ref{sec:weak_discrete_boundary_condition}. PINNs should follow the same principle.

\begin{quote}
    \textit{ PINNs are fundamentally different from traditional numerical methods, and the two should be viewed as complementary systems, not replacements for one another.}
\end{quote}

\subsection{Data-Driven Constitutive Modeling}
\begin{quote}
    \textit{The constitutive relationship of a physical system corresponds to an intrinsic relationship within an AI model.}
\end{quote}
\paragraph{Constitutive Law as a Learnable Manifold}
In high order nonlinear regimes, especially in the plasticity domain, the 
constitutive law is an \emph{intrinsic relationship} that is often difficult 
to obtain from either in--situ measurement or laboratory testing. Neural 
networks replace the closed--form mapping
\begin{equation}
    \sigma = f(\varepsilon)
\end{equation}
with a learned constitutive manifold
\begin{equation}
(\varepsilon,\sigma)\in M_\theta, 
\qquad 
M_\theta = \{(\varepsilon, f_\theta(\varepsilon))\}
\end{equation}
making the intrinsic law \emph{learnable} rather than prescribed.  
Although illustrated in solid mechanics, the same principle extends naturally 
to biology, chemistry, and other systems where intrinsic constitutive 
relationships are unknown or inaccessible.

\paragraph{Constitutive Law as Fixed Point Iteration}
Plasticity introduces high order nonlinearity that obscures the intrinsic 
constitutive relationship. A neural network treats the constitutive update 
as a fixed--point problem:
\begin{equation}
(\varepsilon,\sigma) = \Phi_\theta(\varepsilon,\sigma)
\qquad 
R_\theta(\varepsilon) = \sigma - f_\theta(\varepsilon)
\end{equation}
Training solves the problem
\begin{equation}
\theta^\ast = \arg\min_\theta \,\mathbb{E}\,\|R_\theta(\varepsilon)\|^2,
\end{equation}
thereby recovering the intrinsic law as a numerically learned property.  
Although derived from solid mechanics, this framework generalizes to biological 
tissues, chemical reactions, and other nonlinear material systems that also 
possess intrinsic constitutive relations.

\paragraph{Constitutive Update as Projection onto a Learned Data Manifold}
In high--order nonlinear materials, the constitutive mapping remains intrinsic 
but does not admit a clean analytic formula. A neural network formulates the 
constitutive update as a projection onto a learned data manifold:

\begin{equation}
(\epsilon, \sigma)^{\ast}
= \arg\min_{(\epsilon',\,\sigma') \in \mathcal{M}_{\theta}}
        d\!\left((\epsilon_{\mathrm{obs}},\,\sigma_{\mathrm{obs}}),\,
                (\epsilon',\,\sigma')\right)
\label{eq:constitutive_pdate}
\end{equation}

Note on Equation (\ref{eq:constitutive_pdate}}), 
The minimization is performed over all admissible stress--strain pairs, 
$(\epsilon',\sigma')$ on the learned constitutive manifold, 
$\mathcal{M}_\theta$. The observed pair,
$(\epsilon_{\mathrm{obs}},\sigma_{\mathrm{obs}})$ is projected onto,
$\mathcal{M}_\theta$ by minimizing the manifold distance $d(\cdot,\cdot)$.

where \(M_\theta\) is the learned constitutive manifold. 
This projection viewpoint applies not only to solid mechanics but equally to 
biomechanics, cellular processes, chemical kinetics, and porous--flow systems, 
all of which possess intrinsic constitutive relations that can be learned in 
this manner.

\section{Manifold Federation: Real World and World Model}
\begin{quote}
    \textit{The World Model must operate under the learning complexity characterized in,  Section~\ref{sec:learning_complexity}. The key question becomes: what strategy allows us to navigate this complexity?}
\end{quote}

Because  training data carry high-order nonlinearity, and minibatch training vs near infinite data scope, a large model quickly develops plasticity and gradually loses learnability as curvature accumulates. As the parameter count grows, the model also begins to suffer a Babuška-type paradox: the numerical error increases rather than decreases, Section~\ref{sec:Babuska_paradox}, since adding more nodes amplifies the geometric inconsistency in the stacking piecewise manifold. At the same time, real-world data has data complexity, Section~\ref{sec:data_complexity}, making it neither possible nor meaningful to aggregate all data into a single centralized model. Under these conditions, federated learning becomes the logical choice: each small elastic model learns a local piece of the manifold where nonlinearity is manageable. Together, these distributed piecewise manifolds align into a coherent global system without requiring all data to reside in one place.

\subsection{Learning Complexity}
\label{sec:learning_complexity}

Neural networks, as a learnable numerical computation framework, must operate
under the structural forces imposed by real-world data. Among these forces,
\emph{high--order nonlinearity is the critical source of complexity}: it shapes
curvature, discontinuity, and multi-scale behavior across the stacked piecewise
manifold. Combined with the near-infinite scope of the data, this nonlinearity
defines the essential difficulty of learning.

Let
\begin{equation}
S = \{s_1, s_2, \dots, s_n\}
\end{equation}
denote all factors that influence the learning process. We identify the
\emph{structural basis} as the minimal subset of forces that determine how the
manifold evolves, how fixed points form, and how elasticity is preserved or
lost:
\begin{equation}
B = \mathrm{Basis}(S)
\end{equation}

In the Deep Manifold framework, this basis specializes in
\[
B = \{
\text{data complexity},\;
\text{training dynamics},\;
\text{neural plasticity},\;
\text{stochasticity}
\}
\]

\emph{Data complexity} (Section~\ref{sec:data_complexity}) arises primarily from high--order
nonlinearity and near-infinite scope.  
\emph{Training dynamics} (Section~\ref{sec:training_dynamics}) introduces continuously shifting
boundary conditions, with \emph{minibatch fragmentation} that exposes only
microscopic, discontinuous slices of the manifold.  
\emph{Neural plasticity} (Sections~\ref{sec:data_efficient}) governs the creation and decay of
elasticity as curvature accumulates.  
\emph{Stochasticity} (Section~\ref{sec:nn_stochastic}) contributes to inequality-driven variability
that defines the allowable region of numerical convergence.

\begin{figure}[H]
    \centering
    \includegraphics[width=0.9\linewidth]{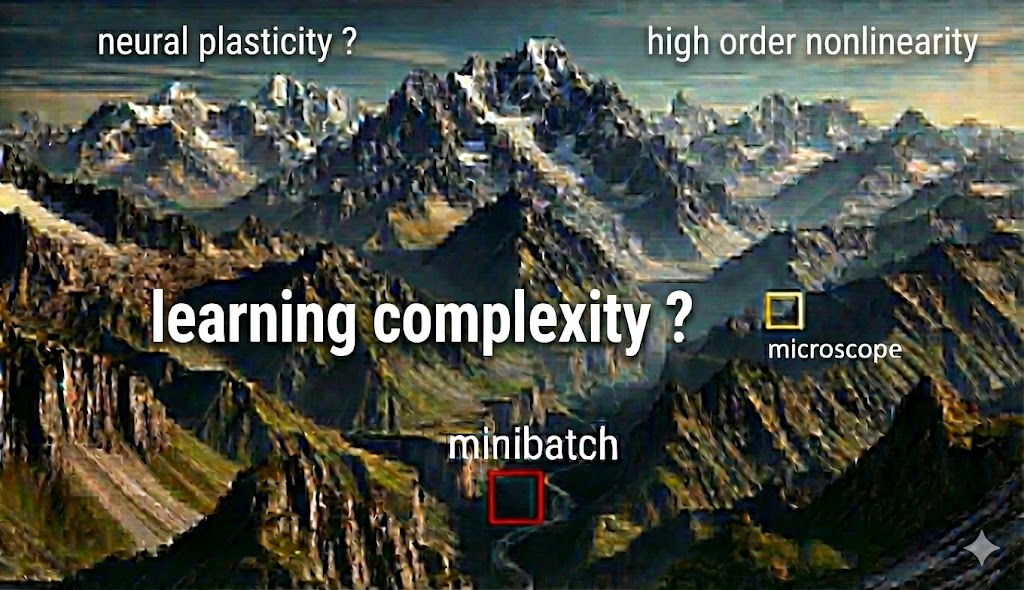}
    \caption{Learning Complexity}
    \label{fig:learning_complexity}
\end{figure}

These forces interact to produce \emph{learning complexity}, defined
abstractly as
\begin{equation}
\mathrm{Learning\ Complexity} = f(B),
\end{equation}
capturing the composite numerical--geometric difficulty of shaping a stable
manifold from partial, fragmented, stochastic, high--order nonlinear data.

\noindent
In summary:  
\begin{enumerate}
    \item \emph{Structural basis} = the fundamental forces governing manifold formation,
    \item \emph{Learning complexity} = the behavior that emerges from their interaction.
\end{enumerate}

\subsection{Model CAP Theorem}
\label{sec:cap_theorem}
In distributed systems, the CAP theorem states that a database cannot 
simultaneously guaranty Consistency, Availability, and Partition tolerance. 
Because partition tolerance is unavoidable, a system must trade between 
consistency and availability. This result is structural: it arises not from 
implementation detail but from the fundamental constraints of a distributed 
environment.

A similar structural limitation appears in large language models. Under the 
learning complexity described in Section~\ref{sec:learning_complexity}: driven by data high--order 
nonlinearity, near--infinite scope, minibatch fragmentation, neural plasticity, 
and stochasticity, a single model cannot simultaneously maximize 
\emph{Coverage}, \emph{Accuracy}, and \emph{Performance}.

\begin{itemize}
    \item \textbf{Coverage} denotes how much of the real--world manifold the 
    model attempts to represent across semantic, symbolic, temporal, modal, and 
    nonlinear axes (Section~\ref{sec:data_complexity}).
    \item \textbf{Accuracy} denotes local geometric fidelity: curvature 
    alignment, fixed--point stability, and residual correctness within each 
    manifold slice.
    \item \textbf{Performance} denotes the numerical efficiency of 
    iterated--integral inference: stability, latency, and computational 
    tractability across stacked piecewise manifolds.
\end{itemize}

Learning complexity forces an inherent constraint: a single model may enlarge 
its manifold (Coverage), improve its curvature fidelity (Accuracy), or maintain 
numerical efficiency (Performance), but not all three simultaneously. Pushing 
Coverage and Accuracy accelerates geometric stiffness and collapses Performance. 
Optimizing Accuracy and Performance restricts the manifold and reduces Coverage. 
Expanding Coverage and Performance sacrifices local curvature Finiteness, breaking 
Accuracy.

Formally, the stacked piecewise manifold offers a finite curvature capacity, while 
minibatch fragmentation limits how much global geometry is revealed per 
iteration. As curvature accumulates, plasticity decays, fixed point basins 
narrow, and the elastic region of the model diminishes. These forces impose a 
CAP-like boundary on single-model learnability:
\begin{equation}
\text{Learning Complexity}
= f(\text{Coverage},\ \text{Accuracy},\ \text{Performance}),
\label{eq:cap_lc}
\end{equation}
where increasing any two terms induces geometric and numerical pressure on the 
third.

This \emph{Model CAP Theorem} establishes a foundational constraint for 
frontier-scale systems. It lays the groundwork for the following sections, where 
scalability is achieved not by enlarging a single manifold, but by distributing 
Coverage across many small elastic models and aligning them through manifold 
federation.

\subsection{Mosaic of Small Elastic Model}
\label{sec:mosaic_small_elastic_model}
\begin{figure}[H]
    \centering
    \includegraphics[width=0.5\linewidth]{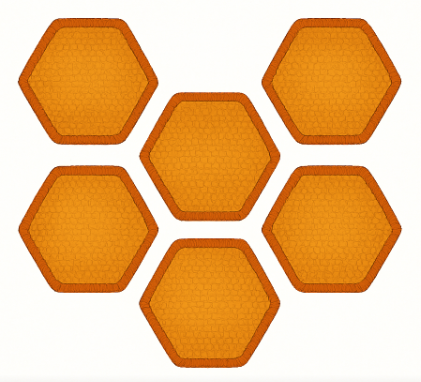}
    \caption{Mosaic of Small Elastic Model}
    \label{fig:mosaic_small_elastic_model}
\end{figure}

Given the near-infinite scope of data spanning multiple dimensions and embedded with high-order nonlinearity, the task of approximation becomes too fragmented and granular for a single frontier-scale model to capture effectively. Attempting to cover all scales, see Figure \ref{fig:mosaic_small_elastic_model}, with a global structure, risks inefficiency, instability, and even error amplification.

We pose the same question for AI: when models scale to billions of parameters (nodes), do we encounter an analogous paradox? That is, could excessive scaling introduce systemic errors, making the model less faithful rather than more? Just as adding more nodes to a numerical scheme can amplify rather than cancel error, extending model scale and context can cause errors to propagate rather than diminish. More broadly, the challenge of increasingly long-context agentic tasks reinforces this concern: errors do not simply accumulate, they can compound exponentially with task length. What looks like a small deviation at one step may cascade across iterations, leading to breakdowns in reasoning, tool use, or context handling.

The Deep Manifold answer points toward an alternative: instead of one monolithic frontier model, the future lies with millions of small models (SLMs). Each model captures a local piece of the manifold at granular scale, aligned with the nonlinearity of its data slice. Together, these SLMs form a mosaic that resists error compounding, achieving accuracy, efficiency, and adaptability that no single oversized model can sustain. This shift from frontier to distributed architecture mirrors the transition from brute-force scaling to modular, manifold-like grouping, where structure, not size, provides the path forward. Since a manifold is a very general concept (point, curve, surface, or even high-dimensional hyperbolic space), each small model can itself be understood as a manifold, and the mosaic of SLMs as an interconnected manifold system.

\subsection{Prompt Complexity}
\label{sc:prompt_complexity}
Prompting is a geometric operation performed at inference time. A prompt imposes 
boundary conditions that guide the iterated--integral process introduced in 
Section~\ref{sec:iterated_integral}, determining how the model traverses its 
stacked piecewise manifolds. Prompting carries its own complexity: the Ask 
restricts the region of convergence, the Instruction shapes the boundary of the 
iterated integral, and the model contributes only the fixed points already 
present in its manifold.

\paragraph{Model.}
The model provides the available fixed points. Inference cannot create new fixed 
points; it can only select among those formed during training. For a given prompt, 
the model may or may not contain a suitable fixed point.

\paragraph{Ask (scope).}
The Ask specifies the intended outcome and defines a region restricted by the scope of 
the representational manifold. A sharp ask identifies a compact target region; a 
diffuse ask leaves the iterated-integral trajectory underdetermined.

\paragraph{Instruction (boundary condition).}
The Instruction supplies the boundary condition for the iterated integral in 
Section~\ref{sec:iterated_integral}. Its structure determines which slices of the 
stacked manifold are traversed and whether the trajectory remains stable. A 
sufficiently rich instruction provides intermediate anchors and narrows the 
admissible integration boundary.

\begin{figure}[H]
    \centering
    \includegraphics[width=0.60\linewidth]{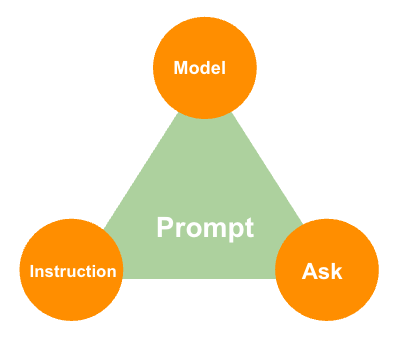}
    \caption{Prompt Complexity}
    \label{fig:prompt_complexity}
\end{figure}

A prompt therefore acts as a compatibility condition: the \emph{Model} contributes 
fixed-point candidates; the \emph{Ask} restricts the target region; the 
\emph{Instruction} defines the boundary at each layer. When these do not align, 
the inference trajectory cannot settle into a stable fixed point.

\paragraph{Relation to Multi-Model Systems}
A single model does not guaranty a suitable fixed point for every request. This 
directly motivates the structures developed in Section~7.4--7.6:

\begin{enumerate}
    \item In the \emph{Mosaic of Small Elastic Models} 
    (Section~\ref{sec:mosaic_small_elastic_model}), different models correspond 
    to different local manifold slices; an agent may need to try multiple models 
    to locate the slice containing a compatible fixed point.

    \item In the \emph{Federation Triangle} (Section~\ref{sec:federation_triangle}), 
    the agent becomes the operator that tests different instructions, evaluates 
    intermediate residuals, and switches models when necessary.

    \item In \emph{Federated Learning} (Section~\ref{sec:federated_learning}), 
    aligned small models provide multiple fixed-point candidates, enabling the 
    agent to route prompts to the manifold where stable convergence is achievable.
\end{enumerate}

Prompt complexity therefore extends naturally to agentic behavior: an agent may 
vary the instruction (boundary), vary the model (manifold), or both, until a 
convergent fixed point is found.

\paragraph{Mathematical Representation on Prompt Complexity}
Let the prompt be decomposed into an ask $a$ and an instruction $I$:
\begin{equation}
p = (a, I)
\end{equation}

Inference follows the iterated integral conditioned by the boundary of 
Section~\ref{sec:iterated_integral}:
\begin{equation}
h_{k+1}(p)
=
\int_{\partial\Omega_k(I,a)}
f_k\!\big(h_k(p), x_k\big)\,
d\mu_k(I,a)(x_k)
\label{eq:prompt_integral}
\end{equation}

The prompt induces a Prompt Operator:
\begin{equation}
\Phi_p = \Phi_I \circ \Phi_a
\label{eq:prompt_operator}
\end{equation}
and inference becomes the fixed-point iteration:
\begin{equation}
h_{k+1} = \Phi_p(h_k), 
\qquad 
h^\ast = \Phi_p(h^\ast)
\label{eq:prompt_fixed_point}
\end{equation}

When a single model does not admit such an $h^\ast$, an agent can instantiate a 
new operator $\Phi_{p,m}$ by altering the instruction $I$, switching to another 
model $m$, or both. A stable fixed point is reached only when the chosen 
model--instruction pair yields a convergent operator:
\begin{equation}
h^\ast = \Phi_{p,m}(h^\ast)
\label{eq:prompt_multimodel_fixedpoint}
\end{equation}

Prompt complexity therefore reflects not only the internal structure of a single 
prompt but also the need to select the correct boundary and the correct manifold 
among many possible models.

\subsection{Federation Triangle: Agentic AI}
\label{sec:federation_triangle}

\begin{quote}
\emph{“In daily practice, solving a problem is the search for a fixed point; 
decision and execution are simply the iterations that move us toward it.”}
\end{quote}

Neural networks do not solve all problems and cannot solve all problems. Under learning complexity (Section~\ref{sec:learning_complexity}) and the Model CAP constraint (Section~\ref{sec:cap_theorem}), a single model cannot simultaneously sustain wide coverage, local accuracy, and numerical performance across the near-infinite scope of the real world. Moreover, the world in which we operate is fundamentally symbolic (Section~\ref{sec:semantic_symbolic}): databases, spreadsheets, scientific solvers and decades of validated numerical tools already constitute extensive, stable fixed-point systems. Neural networks complement, but do not replace, these operators. As shown in Section~\ref{sec:pinn_nc}, PINNs and classical numerical computation co-exist; the same principle governs agentic systems.

This leads to the \emph{Federation Triangle}: Model, Agent, and Tool. Together, they define how a problem is actually solved in practice.

\paragraph{Model---manifold of fixed points.}
The model provides a representational manifold learned from large-scale pretraining. Its fixed points are semantic and approximate; they do not cover the entire symbolic or numerical space. The model cannot stably represent all problems, only those within its learned manifold geometry.

\paragraph{Agent---operator over fixed points.}
The Agent performs the iteration. Select the boundary condition, decompose the task, evaluate intermediate residuals, and determine when the model’s manifold is insufficient. Functionally, an agent applies an operator
\begin{equation}
x_{t+1} = \Phi_{\text{agent}}(x_t)
\end{equation}
driving the search toward a representational or actionable fixed point.

\paragraph{Tool---extension beyond the manifold.}
Tools are exact operators: retrieval engines, databases, simulations, regression, calculus, and symbolic systems. They form validated fixed-point mappings
\begin{equation}
x_{t+1} = \Phi_{\text{tool}}(x_t)
\end{equation}
whose stability and correctness far exceed the model’s semantic approximations. The tools extend the manifold into symbolic and numerical regions where the learning complexity prevents the model from operating reliably.

\begin{figure}[H]
    \centering
    \includegraphics[width=0.60\linewidth]{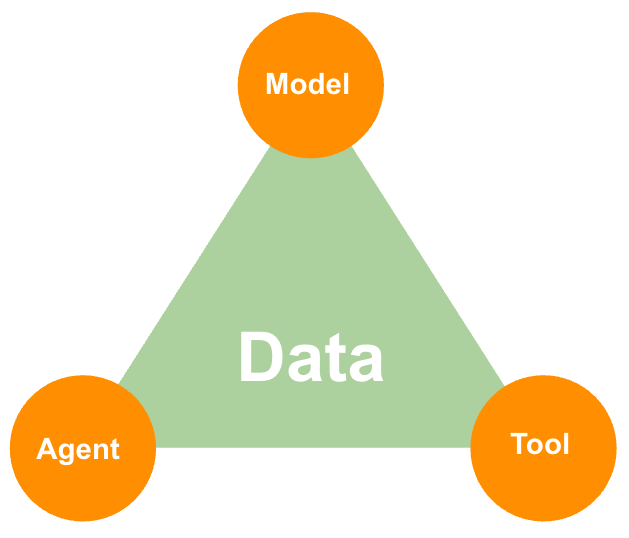}
    \caption{Federation Triangle}
    \label{fig:federation_triangle}
\end{figure}

\paragraph{Fixed-Point Iteration as Agentic Behavior}

The problem is solved when the combined operator stabilizes:
\begin{equation}
x_{t+1} = x_t
\end{equation}
Thus, the agentic loop is a composite fixed-point iteration:
\begin{equation}
x_{t+1}
  = \Phi_{\text{tool}} 
    \circ \Phi_{\text{agent}} 
    \circ \Phi_{\text{model}}(x_t)
\end{equation}
where the \emph{Model} proposes manifold trajectories, the \emph{Agent} chooses and adjusts the operator, and the \emph{Tool} guaranties decisive, symbolic, or numerical convergence.

The final solution is the fixed point of this composition:
\begin{equation}
x^{*}
= \left(
    \Phi_{\text{tool}}
    \circ 
    \Phi_{\text{agent}}
    \circ 
    \Phi_{\text{model}}
  \right)(x^{*})
\end{equation}

\paragraph{Small-Scope Model, Strong Agent, Modular Toolset}

This structure yields a practical rule:
\begin{enumerate}
    \item \emph{Model} --- broad semantic manifold and fixed-point candidates;
    \item \emph{Agent} --- iterative operator selecting the convergence path;
    \item \emph{Tool} --- precise symbolic or numerical anchor.
\end{enumerate}

Agentic AI thus emerges from \emph{fixed-point composition} across distributed operators. The model contributes semantic elasticity; tools contribute stability; and the agent orchestrates iteration. Real-world computation becomes a federated manifold system in which convergence arises from the alignment of Model, Agent, and Tool.

\subsection{Federated Learning}
\label{sec:federated_learning}
With the emergence of \textit{foundation models}, federated learning (FL) should be re-examined through their lens. Under the \textit{Deep Manifold} principle, three core insights emerge naturally:

\begin{enumerate}
    \item Foundation models reduce data burden:  
    The foundation model already encodes a global prior manifold, dramatically lowering the amount of data each client needs. Local fine-tuning becomes the adjustment of boundary conditions, not a reconstruction of representations.
    \begin{equation}
    \theta_i^\star =
    \arg\min_{\theta}
    \mathbb{E}_{x\sim p_i}\!\left[\ell(\theta;x)\right]
    + \beta\, 
    \mathbb{E}_{x\sim \mu}\!\left[
    \mathrm{KL}\!\big(p_\theta(\cdot|x)\,\|\,p_{\theta_0}(\cdot|x)\big)
    \right]
    \label{eq:foundation_prior}
    \end{equation}
    Each local model $\theta_i$ refines its manifold from the foundation prior $\theta_0$.
    
    \item Local models as local manifolds:
    Each local model forms a submanifold of the foundation manifold, which is capable of representing the high-order nonlinearity of local data. The manifold curvature itself absorbs data diversity, allowing the system to maintain global geometric continuity.
    \begin{equation}
    G_i(\theta) =
    \mathbb{E}_{x\sim p_i}\!\left[
    \nabla_\theta \log p_\theta(\cdot|x)\,
    \nabla_\theta \log p_\theta(\cdot|x)^\top
    \right],
    \quad
    \end{equation}
    \begin{equation}
   \theta_{t+1} = \operatorname{Exp}_{\theta_t}\bigl(-\eta\, G_i^{-1}(\theta_t)\nabla_\theta L_i(\theta_t)\bigr)
    \label{eq:local_manifold_geometry}
    \end{equation}

    Training unfolds as the motion of \emph{the piecewise manifolds} in the manifold defined by $G_i$. Here $G_i^{-1}$ denotes the Fisher Information Metric on slice~$i$.
providing the Riemannian geometry used by the natural-gradient update.
    \item Cross-model reward coupling:  
    Instead of sharing parameters or gradients, each local model treats others as \emph{reward models}. By minimizing \textit{KL-divergence} between its output distribution and those of other models, it achieves manifold alignment through boundary consistency---no data exchange, no privacy concern.
    \begin{equation}
    q_{-i}(\cdot|x)=\sum_{j\neq i}\alpha_{ij}\,p_{\theta_j}(\cdot|x),
    \quad
    \sum_{j\neq i}\alpha_{ij}=1
    \label{eq:reward_model_mixture}
    \end{equation}
    \begin{equation}
    \theta_i^\star
    =\arg\min_{\theta}
    \mathbb{E}_{x\sim p_i}\!\left[\ell(\theta;x)\right]
    + \lambda\,\mathbb{E}_{x\sim \mu_b}\!\left[
    \mathrm{KL}\!\big(p_\theta(\cdot|x)\,\|\,q_{-i}(\cdot|x)\big)
    \right]
    \label{eq:kl_boundary_coupling}
    \end{equation}
    KL coupling replaces gradient exchange, enforcing inter-model alignment while Federated Fixed-Point Condition

    \begin{equation}
    \forall i:\;
    \nabla_{\theta_i}\Big\{
    \mathbb{E}_{x\sim p_i}[\ell(\theta_i;x)]
    +\lambda\,\mathbb{E}_{x\sim \mu_b}\!\big[
    \mathrm{KL}(p_{\theta_i}\,\|\,q_{-i})
    \big]
    \Big\}=0
    \label{eq:fixed_point_condition}
    \end{equation}
    Convergence arises when all local manifolds reach mutual consistency under KL boundary equilibrium.
\end{enumerate}

\begin{table}[H]
\centering
\caption{Deep Manifold Federated Learning}
\begin{tabular}{ll}
\toprule
\textbf{Symbol} & \textbf{Description} \\
\midrule
$\theta_0$ & Parameters of the foundation model; global prior manifold. \\
$\theta_i$ & Parameters of local model $i$; adapted from foundation prior. \\
$\mathcal{M}_0$ & Foundation manifold—global latent geometry learned from large-scale data. \\
$\mathcal{M}_i$ & Local manifold of model $i$; curved submanifold encoding local nonlinearity. \\
$\mathcal{D}_i$ & Local dataset of client $i$; defines the boundary condition shaping $\mathcal{M}_i$. \\
$p_i(x)$ & Local data distribution. \\
$p_\theta(\cdot|x)$ & Predictive distribution of model with parameter $\theta$. \\
$G_i(\theta)$ & Fisher information metric defining manifold geometry of model $i$. \\
$q_{-i}$ & Ensemble of other models’ outputs acting as reward/boundary reference. \\
$\mathrm{KL}(\cdot\|\cdot)$ & Kullback–Leibler divergence; measures curvature or boundary misalignment. \\
$\mu_b$ & Boundary sampling distribution used for KL alignment. \\
$\lambda, \beta$ & Regularization coefficients controlling foundation prior and boundary influence. \\
$\mathcal{M}_g$ & Emergent global manifold—smooth composition of all local manifolds. \\
\bottomrule
\end{tabular}
\label{tab:symbolic_meaning}
\end{table}

\vspace{1em}
This \textbf{{Deep Manifold Federated Learning}} redefines federated learning not as distributed optimization, but as \textit{distributed manifold alignment}.  Each local manifold $\mathcal{M}_i$ optimizes under its own boundary condition, while the global manifold $\mathcal{M}_g$ emerges through the curvature-aligned fusion of local piecewise manifolds. Together, these equations describe FL as a \emph{distributed manifold alignment} process, where global learning emerges from the fixed-point convergence of locally curved geometries under consistent topological constraints. The result is a privacy-free, geometry-consistent federation in which all models co-evolve toward a shared equilibrium manifold $\mathcal{M}_g^\star$.

The global equilibrium can be expressed as
\begin{equation}
\mathcal{M}_g^\star = \operatorname*{arg\,min}_{\mathcal{M}_g}
\sum_i D_{\mathcal{M}}(\phi_i(\mathcal{M}_i^\star), \mathcal{M}_g)
\approx
\bigsqcup_i \phi_i(\mathcal{M}_i^\star)
\label{eq:global_equilibrium_manifold}
\end{equation}
representing topological continuity and curvature equilibrium among distributed learners.

\begin{figure}[H]
    \centering
    \includegraphics[width=0.7\linewidth]{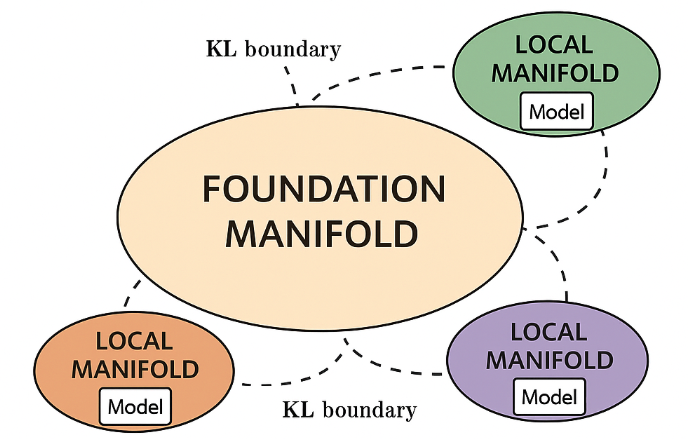}
    \caption{Deep Manifold Federated Learning}
    \label{fig:federated_learning}
\end{figure}

\textbf{Unified Framework}: Deep Manifold Federated Learning can be viewed as a \emph{grand unifying framework} bridging the extremes of modern AI architectures: the \emph{very large foundation model} and the \emph{Mosaic of Small Elastic Models}. In this view, the federated system functions as a collection of \emph{loosely coupled elastic manifolds}, each locally autonomous but globally coherent through manifold-level interactions.

Each local model acts as an \emph{elastic manifold element}, adapting to its own data domain while maintaining geometric alignment with the foundation manifold.
The global manifold emerges not from centralized parameter fusion, but from continuous synchronization of these elastic submanifolds under boundary constraints.
Federated learning thus becomes the large-scale limit of manifold coupling—where foundation models provide curvature priors, small models contribute adaptive elasticity, and the entire system converges toward a unified deep manifold equilibrium.

This framework also offers a potential \textit{solution to the fixed-point paradox in foundation models}. In large-scale training, the fixed point is inherently dynamic: the optimization trajectory deforms the manifold, and the stationary point shifts continuously as the loss landscape evolves.
By introducing \textit{frozen small elastic models} as reference manifolds within the federated structure, we create \textit{quasi-static fixed points} that anchor the evolving global manifold.
These frozen elements act as geometric stabilizers, constraining manifold drift and enabling controlled convergence of an otherwise moving fixed-point system:
\begin{equation}
\nabla_{\theta_t}\mathcal{L}(\theta_t)=0, \quad \frac{d\theta_t}{dt}\neq0
\quad\Rightarrow\quad
\nabla_{\theta_i}\mathcal{L}_i(\theta_i)=0, \;
\theta_i\in\Theta_{\text{frozen}}, \;
i\in\mathcal{S}
\label{eq:anchored_fixed_point}
\end{equation}

Here, $\mathcal{S}$ denotes the subset of small frozen models that serve as quasi-static anchors.
Deep Manifold Federated Learning therefore transforms a moving equilibrium into a \emph{structured manifold of anchored fixed points}, providing both theoretical stability and practical scalability for foundation model training.

\subsection{Inherited Limitation}
\begin{quote}
    \textit{Our perception is finite, and so is every representation built from it.}
\end{quote}
\subsubsection{Human Limitation}
As we build an AI world model, we must first acknowledge that our own understanding of the real world is extremely limited. All training data come through human perceptual channels—language, vision, logic—each a coarse, biased encoding of reality. A model trained on human data inevitably inherits this perceptual boundary: it learns not the real world itself, but the world-as-seen-by-humans. Our senses define the boundary condition of the model, constraining what can be represented and learned. Until multiple perceptual systems are unified, every AI world model will remain a partial slice embedded inside the human perceptual manifold.

\subsubsection{Real World Digital Representation}
Digital representations add a second layer of limitation. Every measurement—RGB images, X-ray, MRI, CT, sensor arrays—is a numerical projection built on our already-limited understanding. Each modality captures a different slice of the physical object and introduces information loss, distortions, and artifacts from the sensing process itself. As Daphne Koller’s pathology work showed, the representation chosen can fundamentally change the predictive power: a temporal-survival representation outperformed spatial annotations, revealing that “reality” expresses itself through multiple representational manifolds. In this sense, abundant data alone is not enough; what the model learns is constrained by the representation itself, not the underlying physical world.

Dr. Daphne Koller’s pioneering 2011 work in breast cancer pathology revealed a profound insight: prediction became more accurate when trained in patient survival outcomes (a causal, temporal representation) than in pathologist annotations (a spatial, visual representation). This finding implied that reality expresses itself through multiple representational manifolds, some geometric, some temporal, some symbolic, and that meaningful learning arises from their alignment, not from any single view.

When an image of a tumor is digitized, it becomes a representation transformation: the complex biological structure of cancer is reduced to pixel intensities so that the computer can “understand”. However, such a transformation inevitably incurs information loss, introduces interpretive contamination, and embeds artifacts from the sensing process itself. No one would claim that RGB values are the best representation of a tumor.

In a deeper sense, Dr. Koller’s framework foreshadowed today’s reinforcement learning: learning not from static labels, but from outcome-driven feedback, the reward signal that anchors representation to consequence, bridging perception and reality.

\subsubsection{Open End vs ``Blue-Collar'' AI}
Real-world data are high-order nonlinear, nearly infinite in scope, and fundamentally stochastic. Under such conditions, a truly \emph{open-ended} AI system, one capable of absorbing the full vertical and horizontal scope of reality, is mathematically and computationally prohibitive. 

This remains the ambition of many AI researchers, but the learning space is too large and manifold curvature accumulates too quickly, for any single system to remain both elastic and stable.

Across enterprise settings, from agriculture to oil/gas, from children's education to healthcare, AI can deliver tremendous value in the form of \emph{self-service automation}. These domains operate on structured symbolic systems, limited scopes, and well-defined boundaries. They demand reliability, efficiency, and integration, not open-ended generality. 

We refer to this as \emph{``Blue-Collar'' AI}: models that solve concrete operational tasks within constrained manifolds, where nonlinearity is manageable and fixed points are well formed.

Sections~\ref{sec:mosaic_small_elastic_model} and~\ref{sec:federated_learning} reveal the structural pathway forward.  
\begin{enumerate}
    \item \textit{Mosaic of Small Elastic Models } provides granularity: many small models, each capturing a local piece of a vast manifold where large models lose learnability.
    \item \textit{Federated Learning} provides coherence: distributed small manifolds aligned into a global system without centralized data, forming a multi-manifold federation.
\end{enumerate}

Together, these form the only viable route toward an \emph{open-end} platform: not a single frontier model, but an \emph{open mosaic} of small elastic models, connected through federated manifold alignment. In practice, Blue-Collar AI solves today's real problems; the open-end side emerges only when many such elastic manifolds are aligned into a coherent global structure.

\section{Closing}
\subsection{Fundamental Questions}
We believe that \emph{Deep Manifold} opens a new mathematical window into the understanding of neural networks. The merit and fidelity of this discovery rest upon two foundational points. Only if these points are established as correct can Deep Manifold grow into a meaningful and potentially powerful mathematical framework for neural networks.

\begin{enumerate}
    \item Is data inherently high-order nonlinear?
    \item Is the neural network fundamentally a numerical computation framework?
\end{enumerate}

\subsection{History Repeating Itself}
Just as in the supercomputer era from the 1970s to the 1990s, scaling once again hit its ceiling : can't scale, the infamous ‘can’t compute any further, can’t compute accurately’ bottleneck. Today, in the age of foundation models and large-scale artificial intelligence, we are facing a strikingly similar moment: capabilities are unprecedented, yet our understanding of high-order nonlinearity remains extremely limited. Without deeper insight into nonlinearity,  manifold dynamics, and fixed-point behavior, we are doomed to repeat the same mistakes

\bibliographystyle{splncs04}
\bibliography{deep-manifod-part2}
\end{document}